\documentclass[manuscript,screen,review=false,anonymous=false]{acmart}
\usepackage{multirow}
\usepackage{subcaption}
\usepackage{float}
\usepackage{enumitem}
\usepackage[table]{xcolor}
\usepackage{graphicx}
\usepackage[percent]{overpic}

\definecolor{LightGreen}{rgb}{0.9,0.98,0.93}

\AtBeginDocument{%
  }

\setcopyright{none}
\copyrightyear{}
\acmYear{}
\acmDOI{}
\acmConference[]{}  
\acmVolume{}
\acmNumber{}
\acmArticle{}
\acmPrice{}
\acmISBN{}
\acmSubmissionID{}
\acmISBN{}
\begin{document}

\title{LastAct: Trajectory-Guided Latest-Activity Localization for Real-Time Smart-Home Activity Recognition}

\author{Zishuai Liu$^*$}
\affiliation{%
  \institution{School of Computing, University of Georgia}
  \country{USA}
}
\author{Ruili Fang$^*$}
\affiliation{%
  \institution{School of Computing, University of Georgia}
  \country{USA}
}
\author{Jin Lu}
\affiliation{%
  \institution{School of Computing, University of Georgia}
  \country{USA}
}
\author{Fei Dou}
\affiliation{%
  \institution{School of Computing, University of Georgia}
  \country{USA}
}
\begin{abstract}
Human Activity Recognition (HAR) from ambient sensors enables smart-home applications such as health monitoring and assisted living. In realistic deployments, however, sensor events arrive as a continuous stream and activity boundaries are unknown. Sliding-window inference therefore produces many windows that straddle transitions and contain mixed activities, creating \textit{boundary contamination} that violates the pre-segmented instance assumption used by most benchmarks and models. Moreover, many pipelines under-use spatial context by treating sensor IDs as independent tokens. We present LastAct, a trajectory-centric framework for streaming smart-home HAR that targets \emph{the most recent activity} under mixed windows while explicitly modeling spatial structure.
% that explicitly models spatial structure and boundary ambiguity for streaming smart-home HAR. 
LastAct projects sensor events onto the home floorplan to form a layout-aligned trajectory image sequence that preserves spatial continuity. A lightweight gate identifies contaminated windows, and a boundary localizer estimates the last transition to enable boundary-guided masking that emphasizes post-boundary evidence and suppresses stale context.
%On top of this representation, a boundary-guided masking mechanism emphasizes signals from the most recent activity within each window, reducing interference from preceding activities in cross-activity windows. 
For efficiency, we reuse a precomputed layout-aligned template cache to avoid repeated rendering. Empirically, across four public smart-home datasets under near-realistic mixed-activity protocols, LastAct achieves competitive or superior performance on pure windows and yields substantial Macro-F1 gains on cross/mixed windows, demonstrating improved robustness under near-realistic sliding-window regimes.
\end{abstract}



\renewcommand{\footnotetextcopyrightpermission}[1]{
  \footnotetext{This manuscript is under review.}
}

\maketitle
\footnotetext[1]{Equal contribution.}
\thispagestyle{empty}
\pagestyle{empty}
\section{Introduction}

Human Activity Recognition (HAR) in smart homes supports a broad class of human-centric applications, including healthcare monitoring \cite{islam2020development, enshaeifar2018health, chen2023digital}, elder care \cite{riboni2016smartfaber, do2018rish}, and ambient assisted living (AAL) \cite{guerra2023ambient, zakka2024action, bari2024advancements}. Compared with camera-based or wearable approaches, ambient sensing leverages unobtrusive infrastructure (e.g., motion detectors, door contacts, and environmental signals such as temperature and pressure) to infer activities of daily living (ADLs) while preserving privacy and requiring minimal maintenance \cite{tewell2019monitoring, uddin2018ambient,alam2012review,chan2008review, mustafa2021iot}. With increasing demand for aging-in-place and remote monitoring amid demographic aging \cite{kanasi2016aging}, HAR systems must operate reliably under realistic deployment conditions characterized by continuous, asynchronous sensor streams.

\begin{figure}[htbp]
  \centering
  \includegraphics[width=0.8\linewidth]{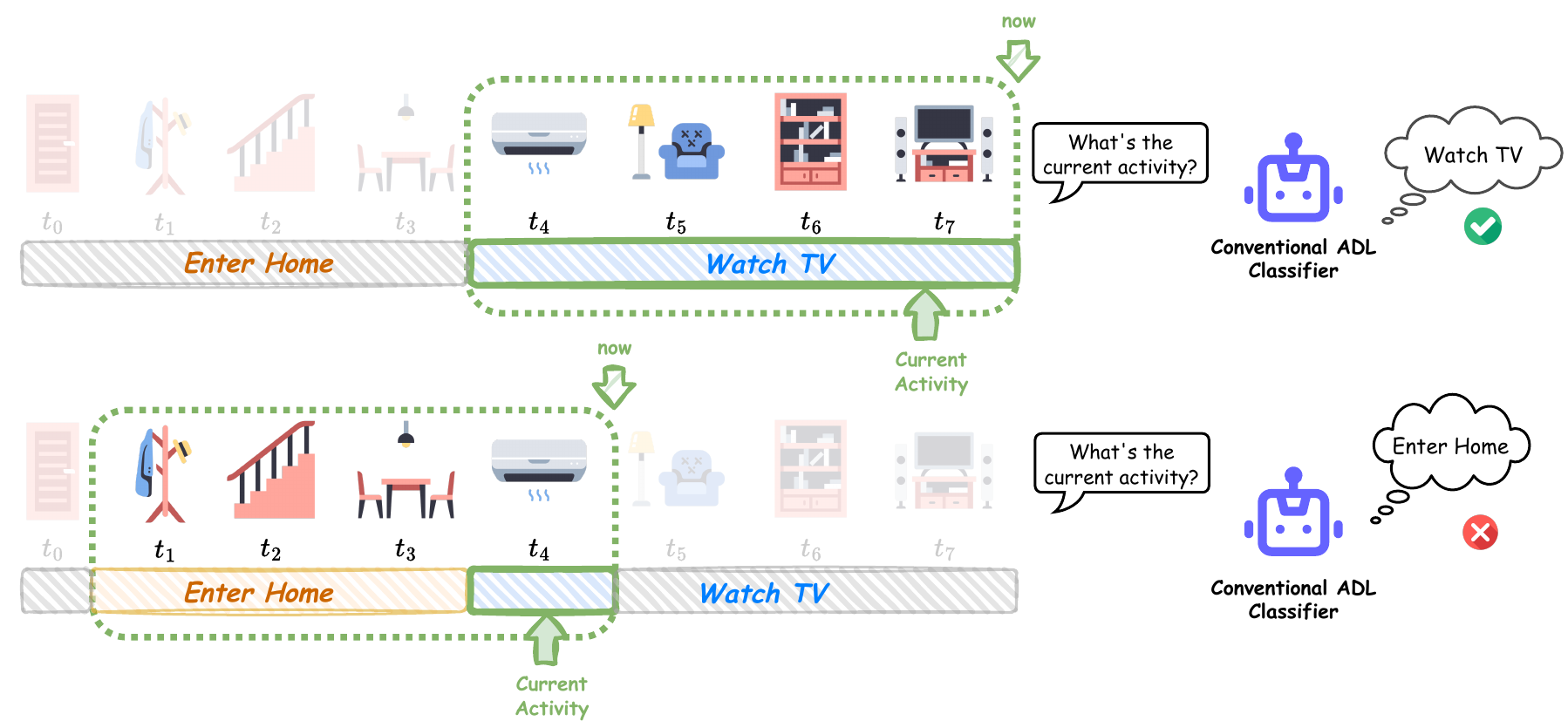}
\caption{Boundary contamination in sliding-window smart-home HAR. 
\textbf{Top:} when the inference window lies within a single activity (a pure window), predicting the activity at the latest timestamp is straightforward (e.g., \emph{Watch TV}). \textbf{Bottom:} in realistic settings, windows frequently span activity transitions (cross-activity windows), where mixed activity fragments introduce ambiguity in latest-timestep prediction, despite clear majority context.}
% \textbf{Top:} when the inference window falls within a single activity (a ``pure'' window), a conventional ADL classifier can confidently predict the ongoing activity (e.g., \emph{Watch TV}). 
% \textbf{Bottom:} in real deployments, windows often straddle transitions and contain fragments of multiple activities (a ``cross-activity'' window), creating ambiguity about the \emph{current} activity.

  \label{fig:intro_problem}
\end{figure}

Despite recent progress in deep learning for ambient-sensor HAR \cite{ghadi2022improving, bouchabou2021survey, bouchabou2021using, bouchabou2021fully, chen2023leveraging,ghods2019activity2vec,liciotti2020sequential,singh2017convolutional}, a critical practical problem remains insufficiently formalized: detecting the currently active activity as early as possible from a continuous sensor stream. Most state-of-the-art methods are developed and evaluated on pre-segmented activity instances with explicit begin/end annotations, assuming that each window contains a single activity.
% ~\cite{ghadi2022improving, bouchabou2021survey, bouchabou2021using, bouchabou2021fully, chen2023leveraging,ghods2019activity2vec,liciotti2020sequential,singh2017convolutional}. 
In real deployments, however, sliding-window inference frequently captures activity transitions, producing windows that contain fragments of multiple activities -- a phenomenon we refer to as boundary contamination (cross-activity window ambiguity), illustrated in Fig.~\ref{fig:intro_problem}. 
The practical impact of mixed-activity windows on recognition performance has been recognized in several earlier streaming HAR studies. Proposed mitigations fall into two broad categories: change-point detection methods that attempt to segment the stream at activity transitions~\cite{thakur2022online,ni2016dynamic,de2021change} and feature-level strategies that re-weight or filter events within a window to suppress stale observations~\cite{ yala2015feature, krishnan2014activity, yan2016real}. However, none of these approaches explicitly detects whether a given window is contaminated, localizes the activity boundary within it, and targets the latest activity for prediction within a unified framework. Without such a formulation, it is difficult to design targeted solutions, establish controlled benchmarks, or systematically compare methods under varying degrees of activity mixing.

% \textbf{(2) Under-utilization of spatial context.} 
What's more, human behavior is fundamentally spatial, yet many HAR pipelines treat sensor IDs as independent categorical tokens \cite{liciotti2020sequential}, ignoring sensor placement and spatial relationships. Some methods impose coarse spatial structure by manually arranging sensors along a 1D axis or using adjusted coordinates \cite{gochoo2018unobtrusive}. Others model inter-sensor dependencies via graphs or incorporate room/layout information through auxiliary embeddings \cite{thukral2025layout, chen2024towards, plotz2024using}. In the wearable domain, temporal-spatial attention mechanisms have shown promise for dynamically emphasizing informative input regions~\cite{li2023temporal}, but these operate on dense IMU signals and do not transfer directly to sparse, event-triggered ambient streams. Existing smart-home spatial representations, however, typically discretize space and fail to capture the continuous flow of movement across the environment, limiting both recognition robustness and interpretability under realistic deployments.

% \textcolor{orange}{
% \textbf{(3) Neglect of time-sensitive contextual cues.} Temporal regularities are also critical for disambiguating ADLs, but are often encoded only implicitly via event order \cite{liciotti2020sequential, gochoo2018unobtrusive, mohmed2020employing} or through coarse time bins (e.g., weekday or time-of-day indicators) \cite{chen2024towards, civitarese2024large}. Such limited encoding overlooks recurring cyclic patterns in daily routines (e.g., morning versus night behaviors) and weakens the model's ability to separate activities that are spatially similar but temporally distinct.
% }

To address these challenges, we propose a trajectory-centric framework for streaming smart-home HAR that predicts the most recent activity within each sliding window under realistic boundary contamination. LastAct integrates layout-grounded spatial cues and cyclic temporal context to build discriminative representations, and introduces boundary-aware inference to isolate the most recent activity signal from preceding context. We first project sensor events onto the floorplan to construct a \textbf{layout-aligned trajectory image-sequence representation}, and then encode \textbf{cyclic temporal context} to capture recurring routine patterns across days and hours. Building on these representations, we introduce a \textbf{guided masking/cropping} mechanism that focuses inference on the latest activity evidence within each window, reducing interference from preceding activities. Finally, for efficient deployment, we introduce a trajectory memory bank that caches and reuses trajectory representations across overlapping windows, reducing redundant computation during streaming inference.

In summary, our contributions are:
\begin{itemize}
    \item We provide a precise and actionable formalization of \textbf{boundary contamination} (cross-activity window ambiguity) as a critical obstacle for both a raw-stream sliding-window protocol and a controlled near-realistic evaluation protocol that parameterizes window purity.
    \item We propose a \textbf{layout-aligned trajectory image-sequence representation} that better preserves spatial continuity than tokenized sensor IDs and complements existing coarse layout encodings.
    % \item We incorporate \textbf{cyclic temporal encoding} to model time-of-routine cues beyond event order or coarse bins \cite{liciotti2020sequential, gochoo2018unobtrusive, mohmed2020employing, chen2024towards, civitarese2024large}.
    \item We introduce a \textbf{guided masking} strategy that localizes the latest activity signal within contaminated windows, targeting online inference settings where activity boundaries are unavailable.
    \item We design a \textbf{trajectory memory bank} that caches layout-aligned trajectory images for overlapping windows, improving online inference efficiency by avoiding redundant trajectory computation.
\end{itemize}

\section{Related Work}
Smart-home Human Activity Recognition (HAR) has shifted from classifying isolated actions to interpreting continuous, noisy sensor streams~\cite{chen2019sensor, bouchabou2021survey}, exposing limitations of the traditional Activity Recognition Chain (ARC)~\cite{bulling2014tutorial}, which rigidly separates representation, segmentation, and classification~\cite{liu2022practical}.

Existing methods are broadly knowledge-driven or data-driven. Knowledge-driven systems use ontologies and expert-defined rules~\cite{chen2019sensor}, offering transparency but scaling poorly due to manual rule engineering and sensitivity to real-world noise. Data-driven approaches learn mappings from sensor events to activity labels, capturing non-linear patterns and improving robustness in unconstrained settings. Recent work increasingly adopts end-to-end deep learning, which unifies feature learning and classification and reduces ARC-style modular inefficiencies~\cite{wang2019deep}.

However, a persistent gap remains between offline benchmarks and real-time use: most models are evaluated on pre-segmented data~\cite{ghadi2022improving, bouchabou2021survey, bouchabou2021using, bouchabou2021fully, chen2023leveraging,ghods2019activity2vec,liciotti2020sequential,singh2017convolutional}, sidestepping the overlapping and ambiguous boundaries that dominate live streams. Robust deployment therefore requires methods that explicitly handle boundary uncertainty in continuous, unlabeled environments. The remainder of this section reviews progress in spatial–temporal embeddings, modular decision logic, and the challenges of continuous real-time detection.

\subsection{Spatial and Temporal Embedding Techniques}
The robustness of smart-home activity recognition hinges on how spatial and temporal cues are represented, yet prior work often trades off computational efficiency against the fine-grained resolution required for real-time deployment.

\textit{Spatial modeling.} Early approaches ignored explicit layout, treating sensor activations as purely sequential streams \cite{liciotti2020sequential, oguntala2021passive}. Later methods encoded implicit spatial structure by arranging sensor IDs in matrices according to physical proximity \cite{gochoo2018unobtrusive, mohmed2020employing, arrotta2022dexar}, which captures coarse neighborhood relations but not full movement trajectories. Graph-based models instead build sensor graphs from inferred interaction dependencies (e.g., functional links between appliances and motion sensors) \cite{ye2023graph, plotz2024using}. More recently, TDOST \cite{thukral2025layout} and related LLM-based methods \cite{chen2024towards} embed room-level semantics, but they often oversimplify movement dynamics and discard path-level information needed to separate activities occurring in the same region.
\textit{Temporal modeling.} Most methods rely on sliding windows to model sequential dependencies \cite{liciotti2020sequential, gochoo2018unobtrusive, mohmed2020employing, lesani2021smart, arrotta2022dexar}. Some add coarse temporal cues (e.g., weekday or discretized hour bins) \cite{chen2024towards, ye2023graph, thukral2025layout}, but these do not fully capture the periodic/cyclical structure of behavior, and few combine multiple temporal granularities (hour/weekday/minute). TDOST~\cite{thukral2025layout} finds that appending coarse time-of-day labels does not improve recognition accuracy, and CARE~\cite{ zhao2026care} shows that widening temporal bins degrades performance. Both suggest that such encodings are insufficient to disambiguate spatially similar activities that differ primarily by time (e.g., morning vs.\ evening medication routines in the same kitchen area).

\textit{Unsupervised discovery and interpretability.} DISCOVER \cite{karpekov2025discover} uses self-supervised embeddings to identify recurring patterns (“activity hubs”) without labels, but the resulting clusters can be difficult to interpret clinically. VISAR \cite{karpekov2025visar} addresses this by replaying sensor sequences over 2D floor plans for expert validation and has shown that broad labels like “Kitchen Activity” can hide distinct sub-activities with different spatial–temporal signatures \cite{karpekov2025visar}. At the same time, these tools expose a persistent gap: transitions between hubs (i.e., trajectories) are often treated as noise or poorly localized outliers. Building on this insight, our framework preserves discovered hubs via a frozen spatial representation and introduces a Pure–Cross gating module so transitional periods—highlighted as clinically meaningful in DISCOVER/VISAR—are explicitly retained and localized using a boundary detector.

% Though they have not been widely adopted in ambient sensing settings, some previous work applied temporal-spatial attention mechanisms to wearable HAR, offering relevant architectural insights. TS-DyConv~\cite{li2023temporal} dynamically weights convolution kernels along temporal and spatial channel dimensions. However, wearable methods process dense, continuously sampled inertial signals under pre-segmented assumptions, and "spatial" refers to body-worn sensor axes rather than room layout. These differences in data characteristics, activity granularity, and deployment constraints limit direct transferability to event-triggered ambient sensing.
% 
%
Temporal--spatial attention has also been explored in wearable HAR. For example, TS-DyConv~\cite{li2023temporal} dynamically weights convolution kernels along temporal and spatial sensor dimensions to emphasize informative regions of dense inertial signals. However, its setting differs from ours: in wearable HAR, ``spatial'' typically refers to body-worn sensor axes or channels, whereas in smart-home HAR it corresponds to physical sensor locations within a floorplan. Moreover, TS-DyConv targets standard window-level classification and does not explicitly model cross-activity windows or latest-activity boundaries. In contrast, LastAct constructs layout-aligned trajectory sequences from sparse ambient events and performs pure/cross gating, boundary localization, and boundary-guided inference to focus prediction on the most recent activity.

\subsection{Classification Models for Smart Home Activity Recognition}
ADL recognition performance depends on both classifier architecture and learning paradigm. Early statistical models such as SVMs \cite{yala2015feature} and HMMs \cite{asghari2020online} replaced hand-written rules with learned patterns but relied heavily on handcrafted features that generalize poorly to “in-the-wild” variability. Deep learning largely superseded these approaches by automating feature extraction: LSTMs/GRUs became common for long-range temporal dependencies \cite{liciotti2020sequential, singh2017human}, and attention mechanisms further improved robustness by emphasizing salient events within windows \cite{park2018deep}.

However, high supervised accuracy on curated datasets often comes with monolithic “always-predict” behavior that ignores signal ambiguity. To reduce label dependence and improve robustness, self-supervised and contrastive learning has gained traction. 
% 
% Haresamudram et al. \cite{haresamudram2022contrastive} showed that Contrastive Predictive Coding can learn transferable activity representations from unlabeled streams across tasks and home layouts, supporting a modular design where a fixed backbone feeds specialized heads. 
% 
Haresamudram et al.~\cite{haresamudram2022contrastive} showed that Contrastive Predictive Coding can learn transferable activity representations from unlabeled wearable streams across tasks and users; while originally developed for body-worn inertial sensors, the underlying contrastive objective has since been adapted to ambient event-triggered settings (e.g., CARE~\cite{zhao2026care}, Chen et al.~\cite{chen2023leveraging}), supporting a modular design where a fixed backbone feeds specialized heads.
In parallel, hierarchical/gated inference is increasingly used to address temporal ambiguity by decoupling stable-state recognition from transition localization \cite{karpekov2025discover, thukral2025layout}, avoiding forced assignments of ambiguous windows \cite{karpekov2025visar}. This modularity enables applying boundary logic only when transitions are suspected, aligning with explainable HAR efforts \cite{das2023explainable} and addressing resolution gaps emphasized in surveys \cite{bouchabou2021survey}, while improving feasibility for real-time deployment in messy environments \cite{karpekov2025visar}.

\subsection{Real-Time and Continuous Activity Detection}

% The final challenge in activity recognition is moving models from laboratory settings into unconstrained environments where data is continuous and unsegmented. In these cases, handling "real-time" prediction requires the system to process data fast enough to be useful without waiting for an activity to finish before starting to analyze it \cite{haresamudram2022contrastive}.
A central barrier to deploying smart-home activity recognition is handling continuous, unsegmented streams in unconstrained environments. In this setting, ``real-time'' inference requires producing timely predictions without waiting for an activity instance to complete \cite{haresamudram2022contrastive}.

% Standard sliding windows are common because they are fast, but their rigid size is a major weakness \cite{bouchabou2021survey}. Since human actions vary in length, a fixed window often cuts an activity in half or lumps two different actions together. To fix this, researchers have explored Change Point Detection (CPD) to find the exact moment a sensor pattern shifts. For instance, Dhekane et al. \cite{CPDannotation} introduced a framework that uses CPD to improve the annotation and recognition of activities, helping to identify transitions that standard models might miss. However, such CPD methods remain difficult to use in live settings because they often require "look-ahead" data—waiting for future events before deciding where a boundary occurred.

Sliding-window inference remains the dominant choice due to its low latency, but fixed window sizes are structurally mismatched to variable-duration human behaviors \cite{bouchabou2021survey}. A window may fragment a single activity or merge adjacent activities, creating ambiguous inputs. To better localize transitions, prior work has explored Change Point Detection (CPD) to identify when the underlying sensor-generating process shifts. For example, Dhekane et al. \cite{CPDannotation} propose a CPD-based framework that improves activity annotation and recognition by detecting boundaries missed by standard windowing. However, many CPD methods are difficult to operationalize online because accurate boundary decisions often require ``look-ahead'' (future events), which conflicts with strict streaming constraints.

An alternative line of work addresses mixed-activity windows not by avoiding them but by adjusting the feature representation within them. Krishnan and Cook~\cite{krishnan2014activity} pioneered this direction for ambient smart-home streaming HAR, proposing three complementary strategies: exponential time-decay weighting to downweight temporally distant events, mutual-information-based weighting to suppress events from functionally unrelated sensor areas, and appending the previous window's classification probabilities as contextual features. Yala et al.~\cite{yala2015feature} extended this framework with improved feature extraction methods including density-based and TF-IDF representations, while Yan et al.~\cite{yan2016real} introduced spatio-temporal correlation-based dynamic segmentation to prevent grouping events from different functional areas into the same window.

These approaches share our motivation of suppressing stale context, but apply uniform heuristic weighting to every window indiscriminately. In other words, an event gets down-weighted just because it is old, even if it belongs to the current activity, while a recent event from the previous activity still counts, even if it is irrelevant. This may potentially lose informative signal and retain irrelevant noise simultaneously. Our framework avoids this trade-off by first detecting whether contamination is present through a learned gating mechanism, then localizing the transition boundary, and applying masking guided by the estimated boundary position — suppressing pre-boundary events while preserving all post-boundary evidence at full strength.

% Recent work has also highlighted the problem of "background noise" in continuous streams, where long periods of inactivity or "null" events confuse the classifier \cite{mazankiewicz2020incremental}. While several specialized models have been proposed to address these streaming challenges, we do not include them in our formal benchmark. Most existing real-time frameworks are designed for specific hardware configurations or are evaluated on private datasets with unique sensor layouts. This makes a direct "apples-to-apples" comparison impossible without the original source code or identical home layouts.
Continuous streams also contain substantial ``background'' periods (inactivity or null events) that can dominate the timeline and confuse classifiers \cite{mazankiewicz2020incremental}. Although specialized streaming systems have been proposed to address such challenges, we do not include many of them in our formal benchmark: existing real-time frameworks are frequently tailored to specific hardware setups or evaluated on private datasets with idiosyncratic sensor layouts, making direct, reproducible comparison infeasible without identical environments or released code.

% To solve this, there is a shift toward modular processing rather than simple window-based classification. By using a gatekeeper to separate stable "pure" activities from "cross" transitions, systems can apply different logic to each state. This prevents the model from making a "best guess" on a messy transition and instead allows it to focus on accurately localizing the boundary \cite{karpekov2025visar}. This modular approach is becoming a priority for creating reliable, clinically relevant timelines of behavior.
These constraints have motivated a shift from monolithic window labeling toward modular processing. By using a gatekeeper to distinguish stable ``pure'' windows from transitional ``cross'' windows, the system can apply different logic to each regime, avoiding forced predictions during messy transitions and instead prioritizing boundary localization when a transition is suspected \cite{karpekov2025visar}. Such modularity is increasingly important for generating reliable, clinically interpretable timelines from noisy, real-world streams.

% Overall, the limitations in current research highlight a clear need for frameworks that move beyond monolithic classification. While deep learning and layout-agnostic models have improved accuracy, they still struggle with the temporal ambiguity of real-world sensor streams. This motivates our approach to explicitly model spatial trajectories and rich temporal patterns, providing the contextual grounding necessary to distinguish stable activities from transitional boundaries in real-time.
In summary, despite progress in deep learning and layout-agnostic modeling, temporal ambiguity in continuous sensor data remains a key failure mode. This gap motivates our framework, which explicitly models spatial trajectories and richer temporal patterns to separate stable activities from transitional boundaries under real-time constraints.
\section{Methodology}
% \fei{try to add at least one figure to each subsection}}
In this section, we present a trajectory-based deep learning framework for human ADL recognition based on ambient sensor data. The proposed approach is designed to jointly capture spatial layout information and temporal dynamics from sensor activation sequences, while remaining robust to real-world challenges such as heterogeneous sensors, and noisy activity boundaries. 
% As illustrated in Figure~\ref{fig:framework_overview}, our framework consists of several modular components; in the remainder of this section, we describe each module in detail and explain how they work together end-to-end.
% \begin{revblock}
\subsection{Problem Formulation}
\label{sec:problem_formulation}

We study ADL recognition from continuous smart-home sensor streams, where events arrive asynchronously and fixed-length windows may straddle activity transitions. Let $E=\{e_1,\ldots,e_T\}$ denote the time-ordered event stream, where each event is $e_t=[s_t,v_t,\tau_t]$, with sensor identity $s_t$, sensor value or state $v_t$, and timestamp $\tau_t$. Activities are annotated as episodes with begin/end markers. Let $\ell(t)\in\mathcal{Y}$ denote the activity label of event $e_t$ when it falls inside an annotated episode.

Given a window length $w$, an observation window is $W^{(k)}=\{e_{t_k},e_{t_k+1},\ldots,e_{t_k+w-1}\}$, where $t_k$ is the starting index. Our task is \emph{latest-activity recognition}: the target label is the activity at the latest event in the window, $y^{(k)}:=\ell(t_k+w-1)$. A prediction is correct if and only if $f(W^{(k)})=y^{(k)}$. Thus, the target is not defined by majority vote; it may occupy only a minority of events in the window when the window occurs shortly after an activity transition.

To quantify boundary contamination, we define activity purity as the fraction of events in $W^{(k)}$ that belong to the latest-activity label: $\rho^{(k)}=\frac{1}{w}\sum_{j=0}^{w-1}\mathbf{1}[\ell(t_k+j)=y^{(k)}]$. A 100\%-purity window contains only the latest activity, whereas a low-purity window contains a short trailing segment of the latest activity preceded by stale context from earlier activities or transitional intervals. We do not introduce an explicit \textit{Other} class for unlabeled inter-episode events; when present inside a window, they are treated as contextual events rather than semantic activity labels.

The learning objective is therefore to estimate $f:W^{(k)}\mapsto y^{(k)}$, with $y^{(k)}\in\mathcal{Y}$, while maintaining robustness as $\rho^{(k)}$ decreases, i.e., as evidence for the current activity becomes shorter relative to stale preceding context. 

\subsection{Episode-Aligned and Near-Realistic Window Construction}
\label{sec:window_construction}
\begin{figure}[htbp]
  \centering
  \includegraphics[width=0.9\linewidth]{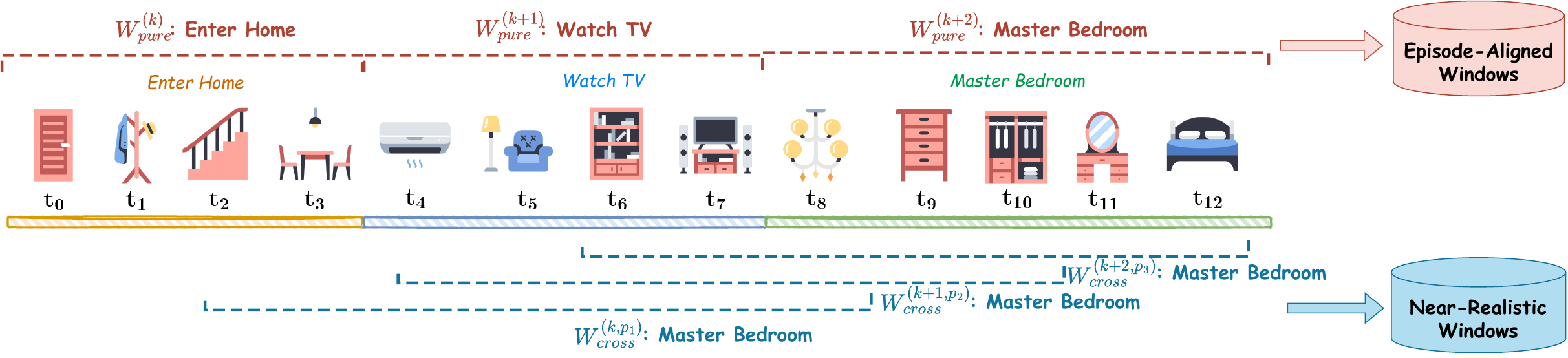}
  \caption{
  Schematic illustration of our window construction on a toy sequence of three activities (\emph{Enter Home}, \emph{Watch TV}, \emph{Master Bedroom}). Pure windows $W_{\text{pure}}^{(k)}$ align with single activity episodes, while cross-activity windows $W_{\text{cross}}^{(k,p)}$ are anchored at the target episode and include additional context from surrounding activities, with the purity level $p$ controlling the proportion of in-episode events.
  }
  \label{fig:window_construction}
\end{figure}

We construct fixed-length observation windows to support activity of daily living (ADL) recognition under near-reality conditions. Each dataset provides activity annotations with explicit \texttt{begin} and \texttt{end} markers, which define contiguous activity episodes in the sensor event stream. Figure~\ref{fig:window_construction} schematically illustrates our window construction on a toy sequence of three activities (\emph{Enter Home}, \emph{Watch TV}, \emph{Master Bedroom}). Episode-aligned \emph{pure windows} $W_{\text{pure}}^{(k)}$ (top) contain only events from a single annotated episode, whereas \emph{cross-activity windows} $W_{\text{cross}}^{(k,p)}$ (bottom) are anchored at the target episode but also include events from surrounding activities, with $p$ controlling how many events come from the target episode.

\paragraph{Episode-Aligned (Pure) Windows.}
For each annotated activity episode starting at time index $t_s$, we instantiate an episode-aligned (pure) window using the first $w$ sensor events:
\begin{equation}
W^{(k)}_{\text{pure}} = \{ e_{t_s}, e_{t_s+1}, \ldots, e_{t_s+w-1} \}.
\end{equation}
Windows are fixed-length: episodes shorter than $w$ events are zero-padded, while longer episodes are truncated. This construction yields one window per activity episode and provides clean supervision in which all non-padding events originate from a single, well-defined target activity.

\paragraph{Near-Realistic Mixed-Activity (Cross-Activity) Windows.}
While episode-aligned windows provide clean supervision, real smart-home streams rarely exhibit perfectly segmented activities. Sensor activations around activity boundaries naturally mix events from adjacent activities, resulting in ambiguous observations. To evaluate robustness under such near-realistic conditions, we construct mixed-activity windows with controlled activity purity.

Specifically, for each activity episode, we use its \texttt{begin} boundary as an anchor point in the original continuous event stream. Given a desired purity level $p \in (0,1]$, we first retrieve the earliest $p \cdot w$ events following the anchor, which belong to the target activity episode. The remaining $(1-p)\cdot w$ events are then taken from the immediate preceding context in the raw stream. All events are preserved in their original temporal order.

Let $w$ denote the window length and $p \in (0,1]$ the desired purity level. For an activity episode beginning at index $t_s$, we allocate
\begin{equation}
m_p = \big\lfloor p \cdot w \big\rfloor, \qquad 
m_c = w - m_p
\end{equation}
events to the target episode and its surrounding context, respectively. The mixed-activity window anchored at this episode is then defined as
\begin{equation}
W_{\text{cross}}^{(k,p)}
= \{ e_{t_s - m_c}, \ldots, e_{t_s - 1}, e_{t_s}, \ldots, e_{t_s + m_p - 1} \},
\label{eq:cross_window}
\end{equation}
where indices are clipped to the valid range of the stream, and zero-padding is applied if fewer than $w$ events are available. By construction, approximately a fraction $m_p / w \approx p$ of events in $W_{\text{cross}}^{(k,p)}$ originate from the target episode, while the remaining events come from the immediately preceding context in the raw stream.

This construction yields fixed-length windows in which a controlled fraction of events originates from the target activity, while the remaining events capture realistic contextual or transitional behavior preceding the episode. A purity level of $p=1.0$ corresponds to the episode-aligned case (up to padding when necessary), whereas lower purity levels increasingly reflect boundary ambiguity and cross-activity contamination.

Importantly, all mixed-activity windows are constructed directly from the original sensor streams rather than through synthesis.
% , without relying on any explicit \textit{Other} activity labeling. 
This preserves the original temporal correlations and sensor dynamics, and enables a continuous and principled evaluation of model robustness as activity boundaries become increasingly ambiguous.

\paragraph{Terminology.}
In the remainder of this paper, we refer to windows with $p=1.0$ as \textit{pure windows}, since all non-padding events come from a single activity episode. Windows with $p<1.0$ are called \textit{cross-activity windows}, as they cross at least one activity boundary and contain events from multiple episodes. Cross-activity windows are further grouped into different purity ranges (e.g., 10--30\%, 40--60\%, 70--90\%) to analyze recognition performance under varying levels of activity mixing.

\begin{figure}[htbp]
  \centering
  \begin{overpic}[width=0.9\textwidth]{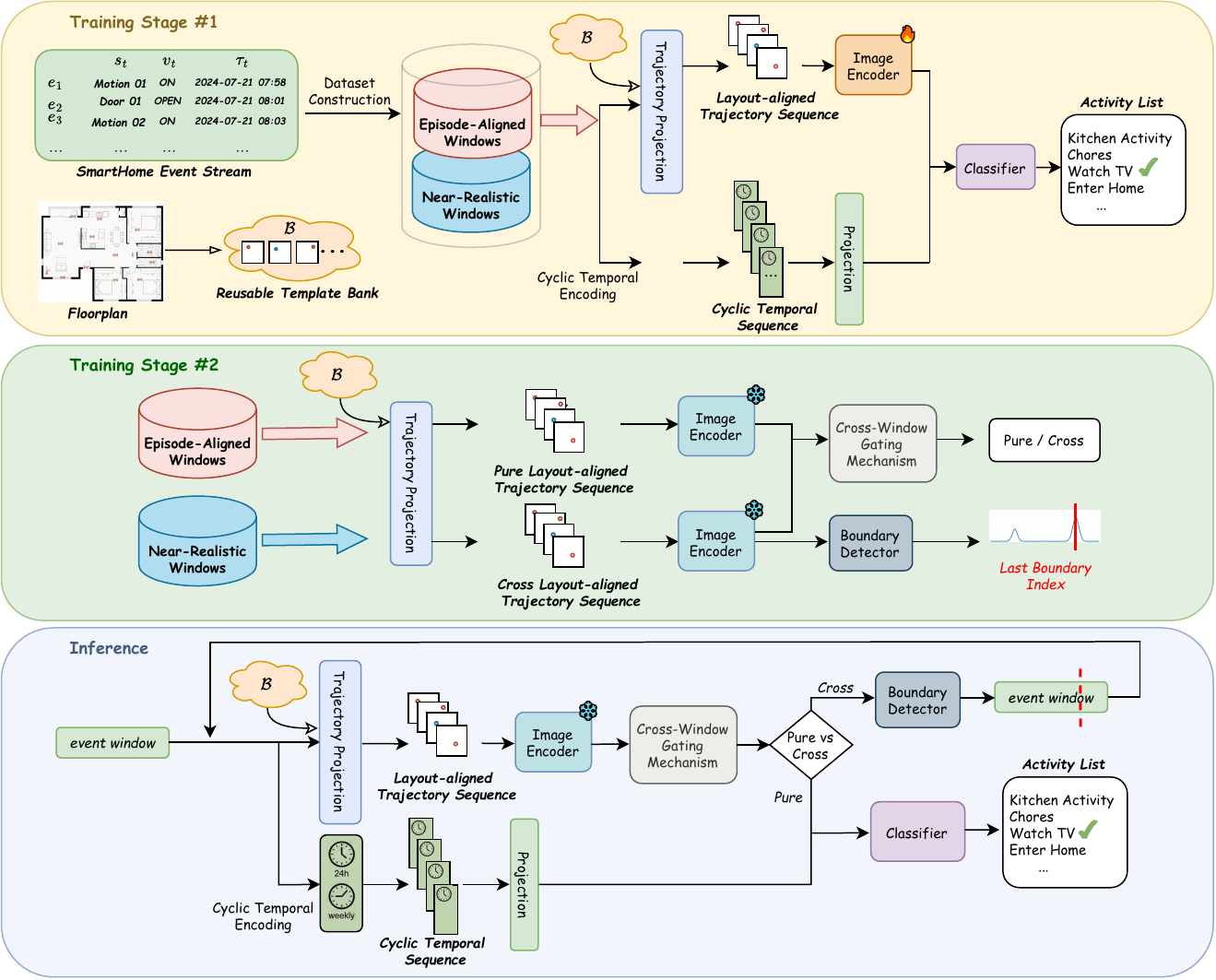}
  % coordinates are percentages: (0,0) bottom-left, (100,100) top-right
  \put(2,78.1){(a)}
  \put(2,50){(b)}
  \put(2,26.5){(c)}
\end{overpic}
  % \caption{
  % Framework overview of our trajectory-based ADL recognition pipeline. The architecture employs a multi-stage optimization strategy to resolve temporal ambiguity in unsegmented sensor streams. Stage 1:\zs{add Description}
\caption{\textbf{Framework overview.}
(a) \textbf{Backbone training (Sec.~\ref{sec:trajectory_backbone}):} we construct episode-aligned and near-realistic windows from the smart-home event stream and floorplan, project events into a layout-aligned trajectory image-sequence (with reusable template caching), optionally fuse cyclic temporal encodings, and train an image encoder and classifier for multi-label ADL prediction.
(b) \textbf{Gate and boundary training (Secs.~\ref{sec:gating}-\ref{sec:boundary_detection})} using pure (episode-aligned) and cross (near-realistic) windows, we train a cross-window gating mechanism to detect pure vs.\ cross windows, and a boundary detector to estimate the last boundary index in cross windows.
(c) \textbf{Inference (Sec.~\ref{sec:prediction}):} given a streaming event window, the gate first determines pure vs.\ cross; if cross, the boundary detector localizes the most recent activity segment for time-focused prediction, enabling robust recognition under boundary contamination while reusing the same classifier.}

  \label{fig:framework_overview}
\end{figure}

\subsection{Trajectory-Based Representation and Activity Backbone}
\label{sec:trajectory_backbone}

\begin{figure}[htbp]
  \centering
  \includegraphics[width=0.9\linewidth]{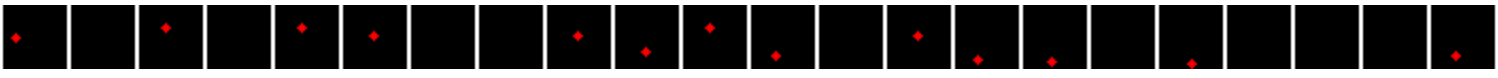}
  \caption{
  \textbf{Example layout-aligned trajectory image sequence from a Milan “Chores” window.} Each panel corresponds to one timestep in the event window. Active sensor events are projected onto the floorplan-aligned canvas and rendered as small markers at their corresponding spatial locations. The resulting image sequence preserves the temporal evolution of sparse ambient-sensor activations while maintaining their spatial layout.
  }
  \label{fig:trajectory_example}
\end{figure}

To explicitly model resident movement patterns within the home environment, we propose a trajectory-based representation that transforms temporal event sequences into spatial visualizations aligned with the physical floor plan. This approach enables the model to leverage geometric relationships between sensors and observe activity patterns as they unfold across physical space. The resulting spatial sequences are then processed by an episode-aligned activity encoder designed to learn discriminative features from temporally coherent activity segments. An overview of this workflow is illustrated in Figure~\ref{fig:framework_overview}(a).

In this work, we use the term \emph{trajectory} to denote the temporally ordered spatial trace induced by sensor activations within an event window. Formally, after mapping each sensor $s_t$ to a floorplan-aligned coordinate $(u_{s_t}, v_{s_t})$, an event window $W^{(k)} = \{e_{t_k}, \ldots, e_{t_k+w-1}\}$ induces a trajectory
\[
\mathcal{T}^{(k)}
=
\bigl(
(u_{s_{t_k}}, v_{s_{t_k}}),
\ldots,
(u_{s_{t_k+w-1}}, v_{s_{t_k+w-1}})
\bigr),
\]
optionally augmented with sensor state, modality, and timestamp information. The trajectory image sequence described below is a rendered representation of this ordered spatial trace, where each timestep is encoded as a floorplan-aligned image frame.

% \begin{figure}[htbp]
%   \centering
%   \includegraphics[width=0.95\linewidth]{figures/methodology/activity_backbone_overview.png}
%   \caption{
%   Overview of the trajectory-based activity backbone.
% Given a floorplan and an episode-aligned event window $(s_t, v_t, \tau_t)$, events are projected onto the floor plan to form trajectory frames and are augmented with cyclic temporal encodings (hour, weekday). An image encoder and temporal projection produce per-timestep embeddings, which are concatenated and fed into a sequence encoder; the pooled window representation is finally passed to a classifier to predict the activity label.
%   }
%   \label{fig:activity_backbone_overview}
% \end{figure}

\paragraph{Layout-aligned sensor cache and trajectory construction.}
We start from the original floorplan image used by practitioners to annotate sensor locations. Each sensor $s$ is placed on this image at pixel coordinates $(\tilde{x}_s, \tilde{y}_s)$, reflecting where the device is installed in the home (e.g., bedroom doorway, kitchen counter). Let $\Omega$ denote this original image coordinate system of size $\tilde{H} \times \tilde{W}$. To obtain a compact and efficient trajectory canvas, we choose a target resolution $R \times R$ (e.g., $R \in \{32, 64\}$) and define a simple resizing map
\[
\pi: \{1,\ldots,\tilde{W}\} \times \{1,\ldots,\tilde{H}\} \rightarrow \{0,\ldots,R-1\}^2,
\]
which rescales $(\tilde{x}_s, \tilde{y}_s)$ to integer pixel coordinates $(u_s, v_s) = \pi(\tilde{x}_s, \tilde{y}_s)$ on the trajectory canvas. This step keeps the relative layout structure but reduces the spatial resolution to accelerate training and inference. Figure~\ref{fig:trajectory_example} illustrates an example layout-aligned trajectory image sequence generated from a Milan “Chores” window. Each panel corresponds to one timestep in the window, illustrating how sparse sensor activations are represented as a temporally ordered sequence on the resized floorplan-aligned canvas.

As a one-time preprocessing step, we construct a \emph{layout cache} that fixes how each sensor appears on a canvas of size $R \times R$. For each sensor identity $s$ we instantiate a blank (all-zero, i.e., black) 3-channel $R \times R$ canvas and render a small disk centered at $(u_s, v_s)$ using a modality-specific color. This yields a bank of reusable templates
\[
\mathcal{B} = \{ \mathbf{I}_s^{(b)} \in \mathbb{R}^{3 \times R \times R} \mid s \in \mathcal{S},\, b \in \mathcal{B}_s \},
\]
where $b$ indexes a coarse activation bin. For binary sensors such as motion, door, or light switches, we only materialize the ``on'' template (the ``off'' state corresponds to the absence of any footprint on an otherwise blank canvas). Temperature sensors are treated as active whenever a valid reading is present. For datasets with richer numeric modalities and only coarse spatial metadata (e.g., room labels)---such as the Orange deployment---we assign each sensor a stable pseudo-location by sampling $(\tilde{x}_s, \tilde{y}_s)$ uniformly within its annotated room region on the floorplan before applying $\pi$, and discretize its readings into a small number of value bins (we use $B=8$ in practice), building one template $\mathbf{I}_s^{(b)}$ per bin with progressively varying intensities. Because this discretization may trade off value resolution against template sparsity, we assess its sensitivity on Orange in Appendix~\ref{app:bin_sensitivity}. In all cases, the cache encodes both geometry (where the sensor sits on the floorplan) and appearance (which color and footprint it uses) once and for all; at inference time, trajectory construction reduces to template lookup and compositing instead of per-event drawing. 
% \begin{revblock}
We validate the sensitivity of these rendering choices (color encoding and disk shape) in Appendix~\ref{app:color-disk_sensitivity}
% \end{revblock}

Given an event window
\[
W^{(k)} = \{ e_{t_k}, \ldots, e_{t_k+w-1} \}, \quad e_t = [s_t, v_t, \tau_t],
\]
we build a spatial trajectory sequence by reusing this cache. For each timestep $j \in \{1,\ldots,w\}$, we collect all events active at that timestep and, for each sensor $s_t$, discretize $v_t$ to a bin $b_t \in \mathcal{B}_{s_t}$ and select the corresponding template $\mathbf{I}_{s_t}^{(b_t)} \in \mathcal{B}$ to obtain an RGB trajectory frame $\mathbf{X}_j \in \mathbb{R}^{3 \times R \times R}$. Repeating this process across the window yields a layout-aligned trajectory sequence
\[
\mathbf{X}^{(k)} = (\mathbf{X}_1,\ldots,\mathbf{X}_w) \in \mathbb{R}^{w \times 3 \times R \times R},
\]
which makes resident movement patterns directly visible as trails on the floorplan and preserves geometric relations between sensors. We further assess the sensitivity of this rendering design in Supplementary Section X by comparing the default colored circular markers with black-and-white rendering and alternative marker shapes, including triangles and rectangles; the results show only minor performance differences, indicating that the model is not strongly dependent on the specific color or marker shape. Compared to aggregating all events in $W^{(k)}$ into a single static occupancy or density image, representing each window as a sequence of trajectory frames explicitly preserves the temporal ordering of spatial footprints. This is particularly important under our near-reality setting, where many windows straddle activity boundaries: early frames may predominantly reflect the previous activity, while later frames are dominated by the upcoming one. By feeding $\mathbf{X}^{(k)}$ into a sequence encoder rather than a purely static image classifier, our backbone can exploit how trajectories evolve over time (e.g., room-to-room transitions, entry/exit patterns) instead of averaging them into a single blurred snapshot.
Because the heavy rendering work has been amortized into the cache, this representation is consistent across windows and computationally efficient even for long sensor streams, and it only requires coarse layout metadata (room membership and sensor type) rather than exact centimeter-level coordinates.

\paragraph{Cyclic temporal encoding.}
In addition to spatial trajectories, temporal context provides important cues for disambiguating activities that follow regular daily routines. Each event $e_t = [s_t, v_t, \tau_t]$ is associated with a timestamp $\tau_t$ in the format \texttt{YYYYMMDDHHMMSS}. We factorize $\tau_t$ into periodic components spanning fine-to-coarse resolutions, $\mathcal{C}_{\text{full}}=\{m_t,h_t,d_t\}$, where $m_t\in[0,59]$ (minute-of-hour), $h_t\in[0,23]$ (hour-of-day), and $d_t\in[0,6]$ (day of week; Monday--Sunday).

To respect their periodic nature, each component $x \in \mathcal{C}_{\text{full}}$ is mapped to a cyclic code via sine and cosine:
\begin{equation}
\theta_x(t) = 2\pi \cdot \frac{x_t}{R_x}, \qquad
\mathrm{feat}_x(t) = [\sin(\theta_x(t)), \cos(\theta_x(t))],
\end{equation}
where $R_x$ denotes the range of component $x$ (60 for minutes, 24 for hours, 7 for weekdays). The cyclic codes from all active components are then concatenated and passed through a small learnable projection to obtain a timestep-wise temporal embedding for event $e_t$:
\begin{equation}
\mathbf{e}^{\text{time}}_t
= f_{\text{time}}\Big( \big[ \mathrm{feat}_x(t) \mid x \in \mathcal{C} \big] \Big) \in \mathbb{R}^d,
\end{equation}
where $f_{\text{time}}$ is implemented as a projection layer in our experiments.

In our main configuration, we focus on the cyclic structure most relevant to smart-home routines by selecting
$
\mathcal{C} = \{h_t, d_t\} \subseteq \mathcal{C}_{\text{full}}.
$
Typical smart-home datasets span at most a few weeks or months, so absolute calendar fields (year, month, day) carry little additional signal, whereas \emph{weekday} and \emph{hour} align naturally with daily and weekly activity patterns (e.g., weekday mornings vs.\ weekend evenings). Minute-level encodings primarily reflect fine-grained, often noisy timing differences (e.g., whether an activity starts at 07{:}01 vs.\ 07{:}03), and are therefore used only in ablations rather than in our default setting. The resulting continuous cyclic encoding removes discontinuities at period boundaries (e.g., between 23{:}59 and 00{:}00) and lets the model capture daily and weekly regularities, while keeping the temporal channel compact and easy to control.

\paragraph{Activity backbone on pure windows.}
Building on the spatial trajectory sequence $\mathbf{X}^{(k)}$ and its aligned temporal descriptors, we first learn an episode-aligned activity backbone using pure windows only. For each window, an image encoder maps every trajectory frame to a spatial feature vector. When temporal encoding is enabled, the corresponding temporal descriptors are concatenated with these spatial features to form per-timestep behavior embeddings. 

The resulting sequence of embeddings is then processed by a causal sequence encoder, which summarizes the window into a single trajectory representation via a mask-aware pooling layer that ignores padding positions and emphasizes informative timesteps. A shallow prediction head finally maps this window-level representation to an activity label distribution.

The backbone is trained exclusively on pure windows ($p = 1.0$), i.e., windows fully contained within single annotated episodes. This episode-aligned training regime allows the encoder stack to focus on learning clean, activity-specific trajectory patterns, without being confounded by heterogeneous transition intervals or neighboring-activity events. In later sections, we reuse this backbone within a trajectory-guided inference pipeline that augments it with pure/cross gating and boundary-aware alignment for mixed-activity windows.

\subsection{Trajectory-Guided Near-Realistic Activity Recognition Framework}
\label{sec:methodology_framework}

Traditional recognition models often rely on the assumption of activity purity, yet in realistic settings, sensor data is a continuous stream where activities seamlessly transition into one another. To bridge the gap between discretized processing and the fluid nature of human behavior, we propose the Trajectory-Guided Near-Realistic Activity Recognition Framework. This approach treats the output of our pretrained representation backbone as a deep feature trajectory. We hypothesize that activity boundaries correspond to latent turbulence in the trajectory, reflected as changes in feature variance and feature-vector norm. By localizing these shifts, the framework decouples the recognition task into a three-stage pipeline: (1) gating the data stream based on its statistical stability, (2) pinpointing the precise boundary of the final activity, and (3) performing inference on the isolated post-boundary segment.

\subsubsection{Cross-Window Gating Mechanism}
\label{sec:gating}
To process efficient real-time streams, we first employ a lightweight \textit{Gating Mechanism} to distinguish between pure and cross windows as illustrated in Fig.~\ref{fig:framework_overview}(b). Instead of re-processing the entire raw sequence, we analyze the statistical properties of the spatial feature trajectory $H_{spatial}$ extracted from the backbone's CNN encoder. We hypothesize that cross windows exhibit higher feature variance and sharper temporal jumps compared to pure windows. 

We form a compact gate descriptor $v_{\text{gate}}\in\mathbb{R}^{2D+1}$ by time-pooling three summary statistics over $T$: (i) a masked mean capturing the overall activity context, (ii) a masked standard deviation capturing within-window variability (typically higher in cross windows), and (iii) a top-$K$ temporal-difference score computed from $\|\Delta H_t\|_2=\|h_t-h_{t-1}\|_2$ and pooled as the mean of the $K$ largest magnitudes (with $K=5$) to emphasize abrupt transitions while suppressing minor fluctuations.

This descriptor is fed into a Multilayer Perceptron (MLP) binary classifier trained with Cross-Entropy loss. To address the potential imbalance between pure and cross window occurrences in real-world streams, we weight the loss function using the inverse class frequency.

Segments identified as "pure" are processed immediately by the standard activity classifier, while those exhibiting high turbulence are passed to the boundary detection stage.
\subsubsection{Feature-Augmented Boundary Detection}
\label{sec:boundary_detection}
When turbulence is detected, the objective shifts to localizing the exact timestep  
$\hat\tau$ where the preceding activity ends and the target activity begins.
We employ a detector that fuses high-level semantic context with sharp, localized sensor signals encoded by the backbone.

The detector takes the frozen trajectory features $H$ from the backbone and projects them to a model dimension $d_{model}$. These are augmented with learnable positional encodings and processed by a Transformer Encoder with $L$ layers. The self-attention mechanism allows the model to compare feature states across the entire window effectively identifying the point where the semantic context shifts from one activity to another. A final linear projection layer followed by a Sigmoid activation outputs a sequence of boundary probabilities $P = [p_1, \dots, p_T]$, where each $p_t \in [0, 1]$ represents the likelihood of a transition occurring at timestep $t$.

Since our goal is to identify the \textit{last} significant boundary in the window (determining the current active class), we employ a composite loss function $\mathcal{L}_{bound}$:
\begin{equation}
    \mathcal{L}_{bound} = \mathcal{L}_{BCE} + \lambda_{card} \mathcal{L}_{card} + \lambda_{last} \mathcal{L}_{KL} + \lambda_{peak} \mathcal{L}_{peak}
\end{equation}
where:
\begin{itemize}
    \item $\mathcal{L}_{BCE}$ is the weighted Binary Cross Entropy loss calculated per timestep against the ground truth boundary labels.
    \item $\mathcal{L}_{card}$ is a cardinality constraint (Smooth L1 Loss) ensuring the predicted number of boundaries matches the ground truth, reducing false positives.
    \item $\mathcal{L}_{KL}$ is a specialized \textit{Last-Boundary} term. We model the ground-truth target as a Gaussian distribution centered at the last true boundary index $t_{last}$. We minimize the KL divergence between the predicted normalized probability distribution and this target Gaussian distribution. This explicitly guides the model to focus attention on the most recent transition, which is crucial for real-time prediction.
    \item $\mathcal{L}_{peak}$ is a peakiness regularization term that minimizes the entropy of the predicted boundary probability distribution, encouraging the model to produce sharp, distinct peaks rather than diffuse uncertainty.

\end{itemize}

\subsubsection{Boundary-Aware Activity Prediction}
\label{sec:prediction}
% The final stage performs boundary-aware inference by isolating the target activity from preceding context in cross windows. Once the last boundary $\hat{\tau}$ is identified via Non-Maximum Suppression (NMS) over the boundary score sequence, we invoke a boundary-guided attention mechanism. Rather than physically discarding pre-boundary timesteps, the pre-trained backbone applies attention pooling with weights explicitly biased toward the post-boundary interval $t \in [\hat{\tau}, T]$, yielding a dynamically ``cropped'' representation in latent space.

In the inference stage, the system adapts the classification strategy based on temporal composition of the input window. Each incoming window is first processed by the cross-window gating module to determine its state as either a pure window or a cross window. 

For pure windows, the representation is passed directly to the primary classification head. For cross windows, the system leverages the estimated last boundary index ($\hat{\tau}_{\text{last}}$) to isolate the signal of interest. To achieve this without losing context or inducing artifacts through hard truncation, we employ a  \textbf{Soft-Attention Cropping} strategy. We construct a guided attention mask $M_{\text{guided}}$, such that:
\[
M_{\text{guided}}(t) =
\begin{cases}
\epsilon, & t<\hat{\tau}_{\text{last}} \\
1, & t \ge \hat{\tau}_{\text{last}}
\end{cases}
\]
where $\epsilon$ represents a near-zero weight. The backbone's attention pooling layer aggregates features under this guided mask, effectively filtering historical signal while preserving the temporal sequence of the current behavior. This mask-driven aggregation allows the classifier to generate a prediction based solely on the most recent status, ensuring the system responsive to real-time behavior shifts rather than being anchored by stale sensor data.

% \subsection{Trajectory-Guided Near-Realistic Activity Recognition Framework}
% \label{sec:trajectory_framework}

% Building on the trajectory-based representation and episode-aligned backbone in Section~\ref{sec:trajectory_backbone}, we now describe how the model is extended to operate under near-reality conditions, where windows may partially overlap multiple activity episodes. The framework introduces two additional components on top of the backbone: a pure/cross gate that routes windows according to their purity, and a boundary detector that localizes recent transitions inside mixed windows. Together, these modules enable trajectory-guided, boundary-aware inference without ever introducing an explicit \textit{Other} label.
`

\section{Experimental Setup}
\subsection{Datasets}
\begin{figure}[h!]
    \centering
    \begin{tabular}{cccc}
        \begin{subfigure}[t]{0.20\linewidth}
            \centering
            \includegraphics[width=\linewidth]{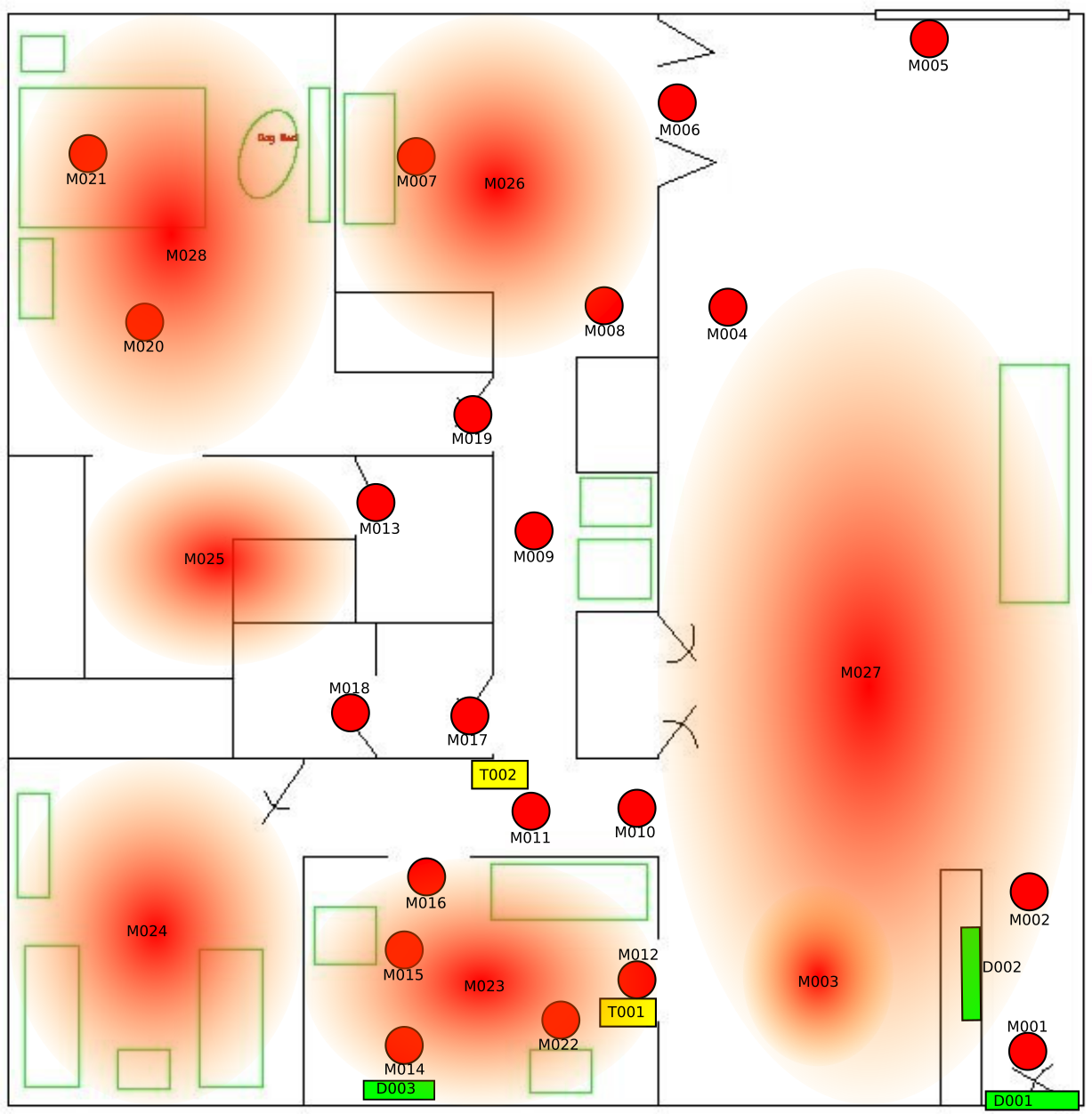}
            \caption{Milan}
            \label{fig:laymilan}
        \end{subfigure}
        \hfill
        \begin{subfigure}[t]{0.30\linewidth}
            \centering
            \includegraphics[width=\linewidth]{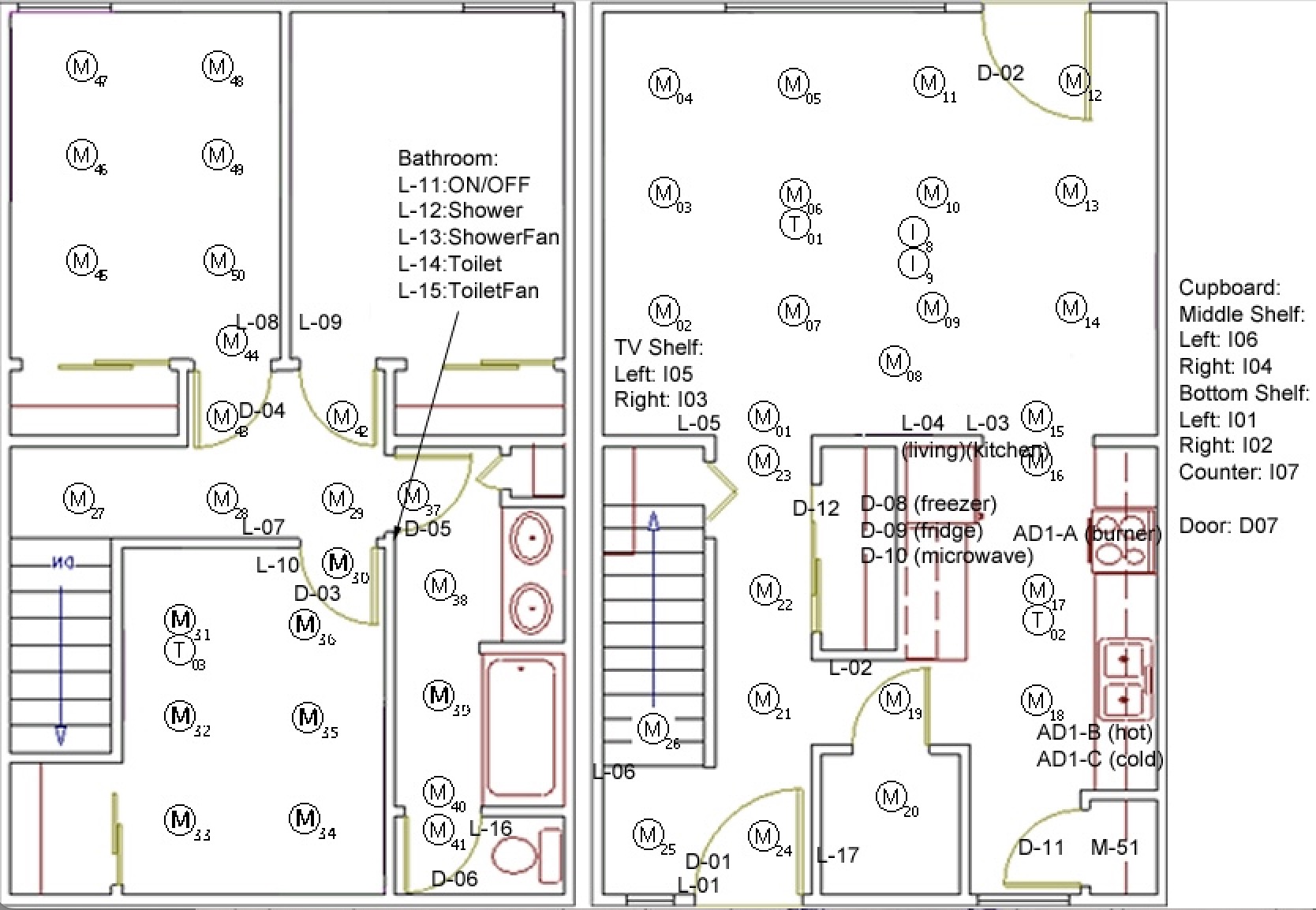}
            \caption{Kyoto7}
            \label{fig:laykyoto}
        \end{subfigure} 
        \hfill
        \begin{subfigure}[t]{0.26\linewidth}
            \centering
            \includegraphics[width=\linewidth]{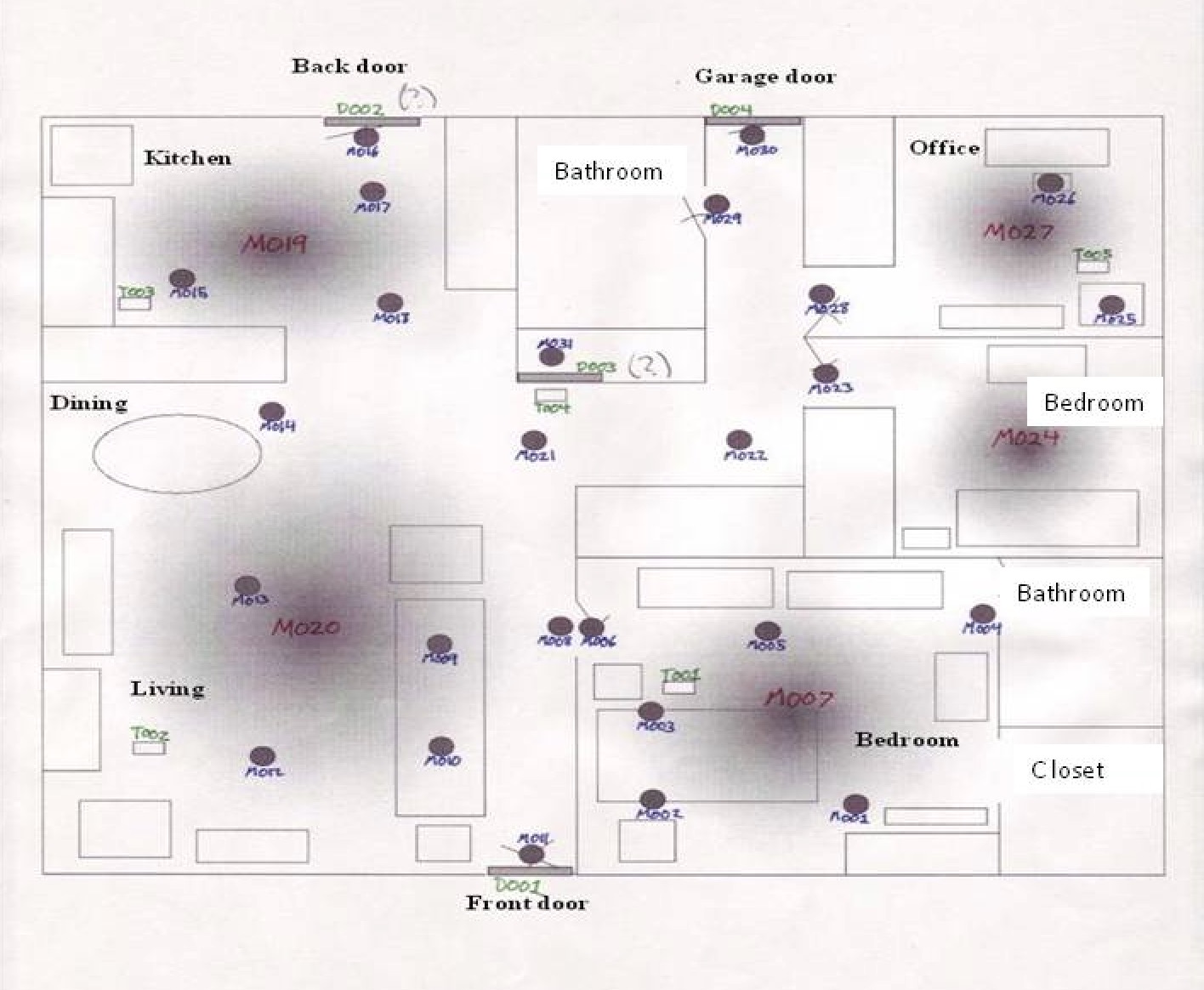}
            \caption{Aruba}
            \label{fig:layaruba}
        \end{subfigure} 
        \hfill
        \begin{subfigure}[t]{0.22\linewidth}
            \centering
            \includegraphics[width=\linewidth]{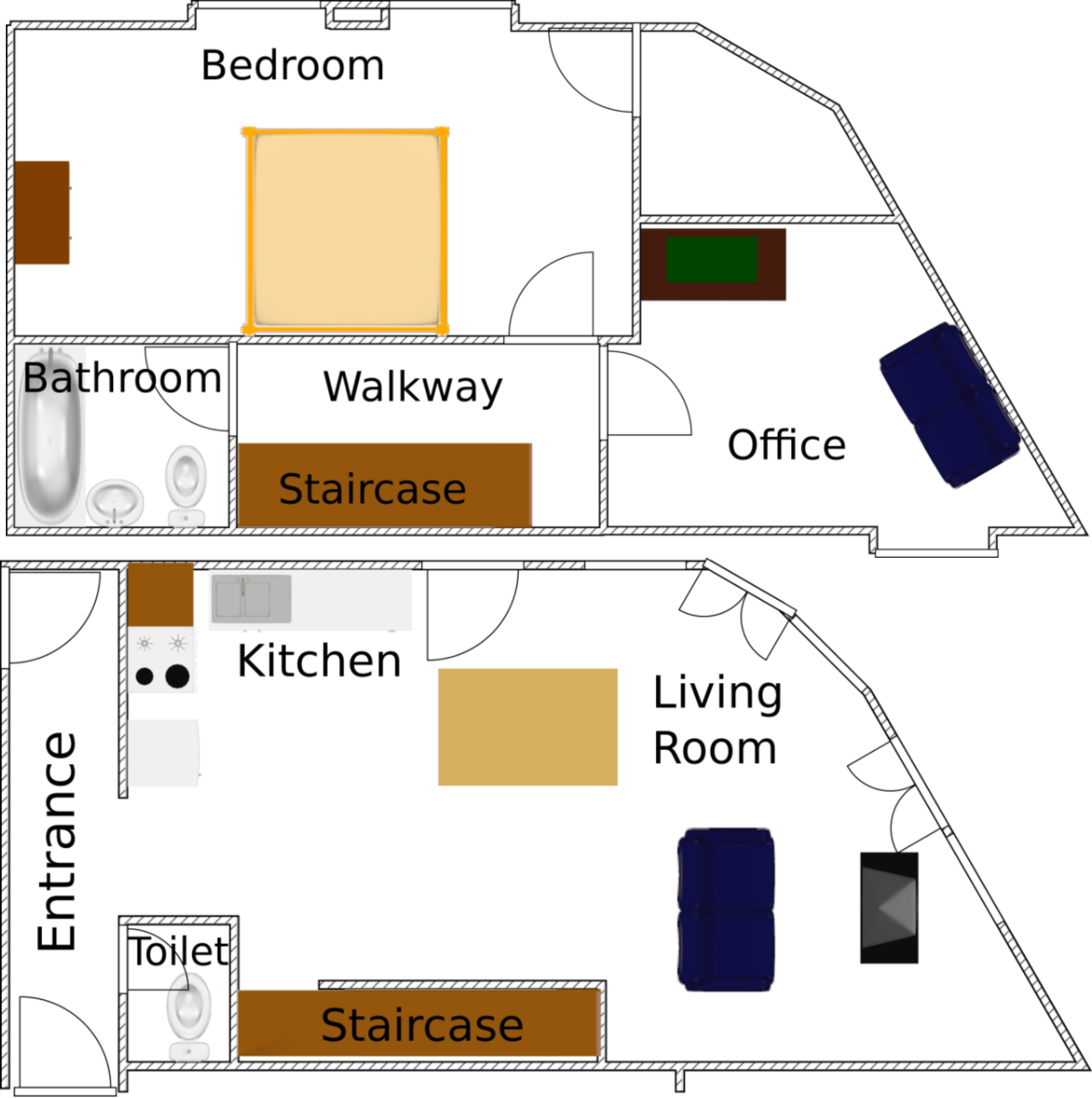}
            \caption{Orange}
            \label{fig:layorange}
        \end{subfigure} 
    \end{tabular}
    \caption{Floorplan layouts and sensor distributions for four smart homes: Milan, Kyoto7, Aruba, and Orange.}
    \label{fig:floormap}
\end{figure}

To evaluate the effectiveness of the proposed approach, we conduct experiments on four widely used smart home datasets from the CASAS~\cite{cook2012casas} project and related deployments: Milan, Kyoto7, Aruba, and Orange~\cite{cumin2017dataset}. These datasets span diverse residential settings, sensing infrastructures, and activity complexities, enabling a comprehensive evaluation under near-realistic ADL recognition scenarios. Figure~\ref{fig:floormap} illustrates the corresponding floor plan layouts and sensor distributions.

% The Milan dataset records single-resident activity in a smart home equipped with motion ("M"), door ("D"), and temperature ("T") sensors, as illustrated in Fig.~\ref{fig:laymilan}. Fifteen distinct activities are annotated, ranging from frequent activities such as Kitchen\_Activity (554 instances) and Guest\_Bathroom (330 instances) to rare events like Meditate (17 instances). Milan provides a rich temporal distribution for studying personalized activity recognition.
The Milan dataset records long-term activity data from a single-resident smart home equipped with motion, door, and temperature sensors (Fig.~\ref{fig:laymilan}). It contains 15 annotated activities with substantial frequency variation, from common routines (e.g., \textit{Kitchen\_Activity}, \textit{Guest\_Bathroom}) to rare classes (e.g., \textit{Meditate}).

Kyoto7 captures multi-resident activity in a two-person apartment (R1, R2; Fig.~\ref{fig:laykyoto}). Beyond motion/door/temperature sensors, it includes appliance sensors (e.g., burner, water flow) and item sensors (``I''). Activities are labeled at both resident level (e.g., R1\_Work, R2\_Sleep) and shared level (e.g., Clean, Study), with \textit{Meal\_Preparation} being the most frequent, enabling evaluation under interleaved and resident-specific behaviors.

Aruba represents single-resident behavior in a home with regular visits from family members (Fig.~\ref{fig:layaruba}). It includes 11 activities and is highly imbalanced (e.g., \textit{Relax}: 2910; \textit{Meal\_Preparation}: 1606), while extremely rare classes (e.g., \textit{Resperate}: 6) pose severe class sparsity.

Orange introduces heterogeneous ambient sensing beyond the CASAS datasets, including energy and water meters, environmental sensors (temperature/humidity), appliance monitors, switches, and presence detectors. Sensors are identified by descriptive strings and linked to room-level semantics rather than explicit coordinates; it also contains global sensors (e.g., whole-home meters) that are not tied to a specific location.
\begin{figure}[htbp]
    \centering
    \includegraphics[width=0.8\textwidth]{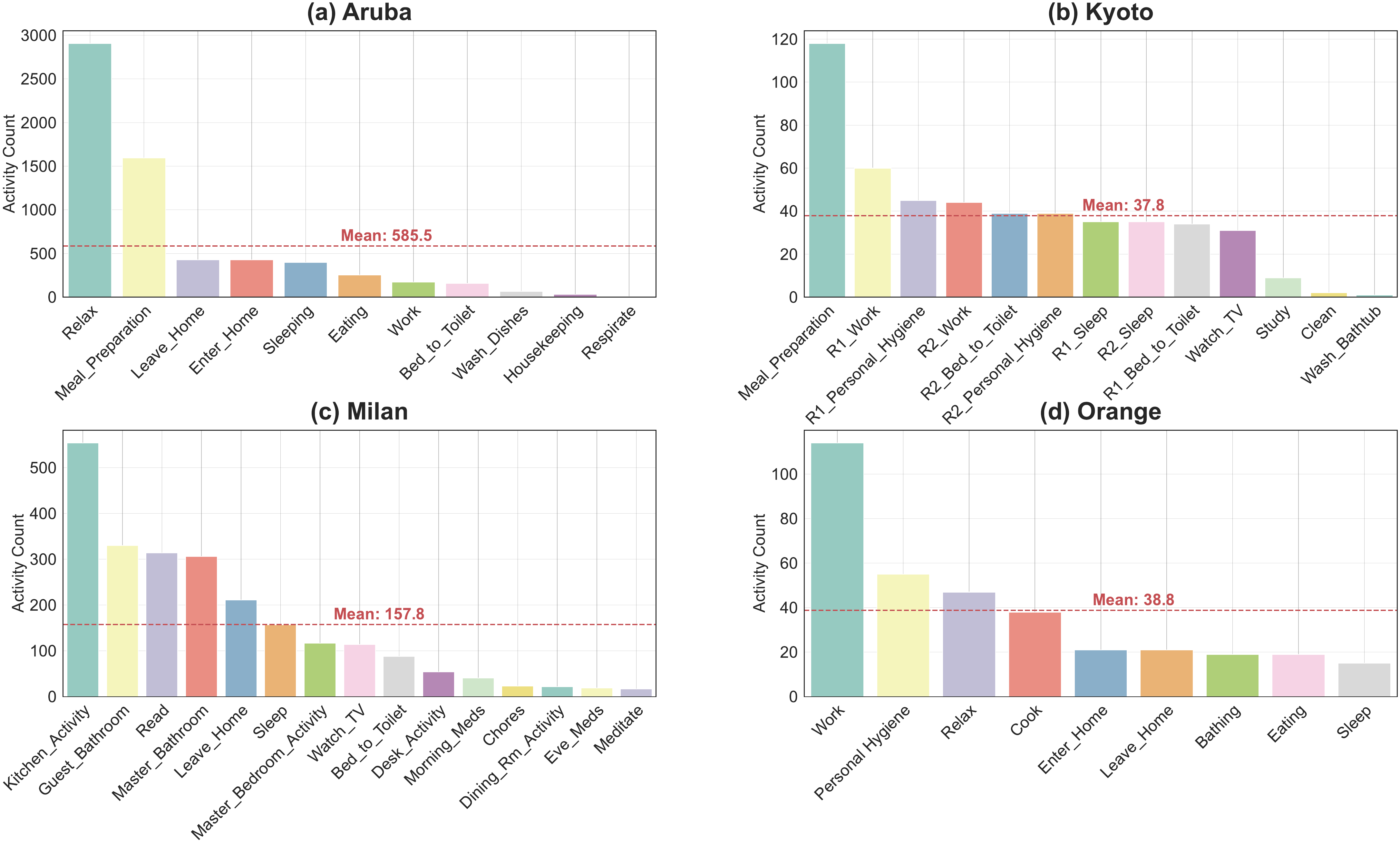}
    \caption{Activity distribution histograms across the smart home datasets. (a) Aruba, (b) Kyoto7, (c) Milan, and (d) Orange. 
The histograms show substantial class imbalance, with a few frequent activities dominating each dataset and several rare activities having limited samples. This imbalance motivates the use of Macro-F1 as the primary recognition metric and supports the need for robustness analysis beyond aggregate accuracy.}

    \label{fig:activity_histograms}
\end{figure}

None of the datasets provide machine-readable sensor coordinates. For CASAS (Milan, Kyoto7, Aruba), approximate sensor positions can be derived from the published floorplans, whereas Orange provides only room-level descriptions. Since our method requires spatial grounding, we exclude sensors that cannot be reliably associated with a specific room/location (notably global sensors) across \emph{all} methods and datasets for a controlled comparison. Additionally, Orange exhibits extreme class sparsity; following prior work (including TDOST), we apply the same label mapping to merge semantically related activities to ensure sufficient samples per class. This mapping is applied only to Orange.

Overall, these datasets span single-resident (Milan, Aruba, Orange) and multi-resident (Kyoto7) scenarios with diverse sensor modalities and pronounced class imbalance. Figure~\ref{fig:activity_histograms} summarizes the activity distributions, which motivates Macro-F1 as the primary metric and highlights the importance of robustness under rare and ambiguous conditions.

\subsection{Window Configuration}

\begin{figure*}[htbp]
    \centering
    \includegraphics[width=0.85\textwidth]{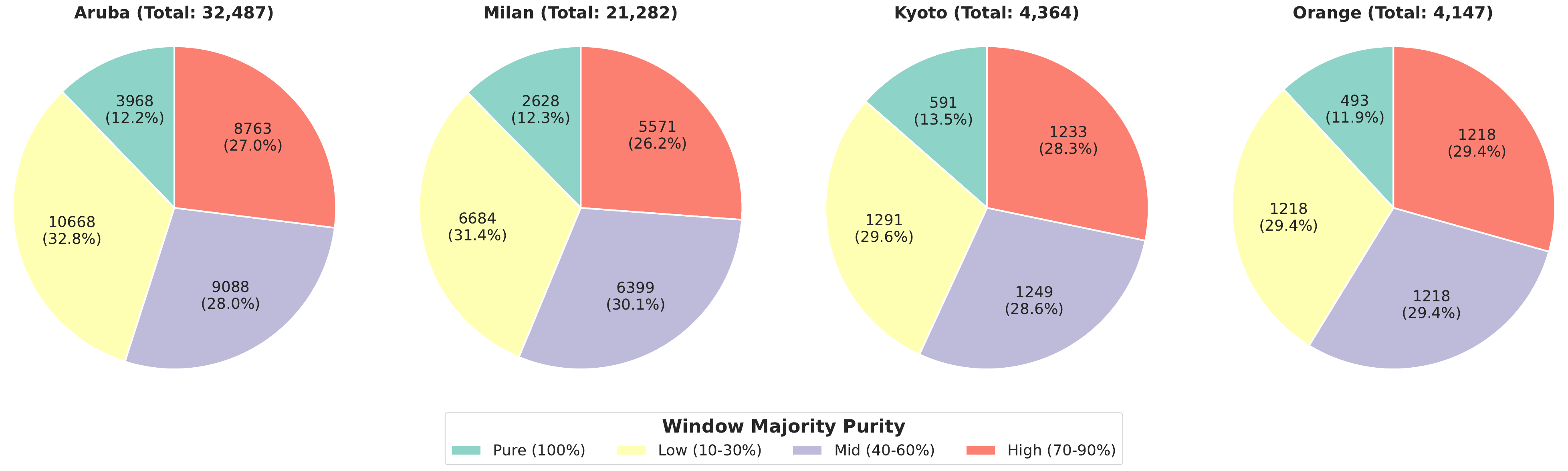}
    \caption{Distribution of activity window purity across four smart-home datasets ($w=100$) under the controlled mixed-window evaluation protocol. 
             Inner labels denote the absolute window count and the relative percentage of the total 
             evaluation set. The larger number of cross-activity windows relative to episode-aligned pure windows illustrates the prevalence of boundary-contaminated evaluation cases around activity transitions.
             % The consistent dominance of cross-activity segments (Low, Mid, and High purity) relative to episode-aligned (Pure) segments across all environments demonstrates the ``Segmentation Trap'' inherent in continuous HAR streams.
             }
    \label{fig:purity_distribution}
\end{figure*}

Following the near-reality window construction described in Section~\ref{sec:window_construction}, we evaluate fixed-length observation windows with varying configurations. Because these windows are constructed from episode boundaries rather than swept over the full stream, no fixed sliding-window stride is used; overlap can occur only among windows generated from the same activity episode at different target post-boundary proportions. Window lengths are set to $w \in \{20, 40, 60, 80, 100\}$ to examine the effect of temporal context on recognition performance. To evaluate robustness under boundary ambiguity, we test mixed-activity windows at purity levels from 10\% to 90\% (step 10\%), using episode-aligned windows (100\% purity) as the reference. As shown in \ref{fig:purity_distribution}, episode-aligned ``Pure'' windows are consistently outnumbered by the large volume of transitional ``Cross'' windows.
Across datasets, mixed-purity windows (10–90\%) dominate the stream, comprising around 86–88\% of all windows. For example, Aruba contains 10,668 low-purity (10–30\%) windows but only 3,968 pure windows, and the same skew holds across layouts from smaller Kyoto/Orange homes to larger Milan/Aruba settings. These findings reveal a "Segmentation Trap": conventional HAR models are trained and evaluated primarily on pure segments, whereas real-time streams are largely cross segments. This imbalance motivates LastAct, which uses Boundary-Guided Inference to explicitly handle frequent transitions and avoid the degradation typical of sliding-window baselines.
% Add this to your preamble if not already there
% \usepackage{graphicx}

\subsection{Evaluation Protocol}

\subsubsection{Stratified Cross-Validation}
\label{sec:day_based_protocol}
We adopt a 3-fold stratified cross-validation protocol independently for each dataset. Episode-aligned windows are split into three disjoint folds stratified by activity label. In each run, one fold is held out for testing, while the remaining two are split into training and validation sets at an 80/20 ratio. These splits are fixed across all methods and purity levels to ensure fair, comparable evaluation.

\subsubsection{Chronological Split}
To better approximate deployment, our day-based protocol constructs splits at the granularity of calendar days, then derives both pure and cross-activity windows within each split. We sort each dataset chronologically and assign the first 60\% of days to training, the next 20\% to validation, and the final 20\% to testing, preventing future information leakage and mirroring realistic use where models trained on historical data are applied to subsequent periods. For each split, Macro-F1 is computed over the set of activity labels present in the corresponding test split. Labels that occur only in training but not in the test set are not included in the macro average, since they have no test support. We verified that no test split contains activity labels absent from its corresponding training split, so all day-based evaluations remain closed-set classification tasks.

Compared to stratified splits, this chronological setting poses a distinct generalization challenge by exposing models to temporal non-stationarities (e.g., routine drift, schedule changes, sensor variations, and shifting event statistics), providing a complementary measure of long-term reliability for smart-home HAR.

% \begin{revblock}
\subsubsection{Raw-Stream Sliding-Window Protocol}
\label{sec:raw_stream_protocol}
To complement the controlled mixed-window protocols, we introduce a raw-stream sliding-window evaluation. Test windows are extracted directly from the chronological event stream, without anchoring them at annotated activity boundaries. Given a test stream \(E=\{e_1,\ldots,e_T\}\), we construct fixed-length windows of size \(w=100\):
\[
W_i=\{e_i,e_{i+1},\ldots,e_{i+w-1}\},
\]
where \(i\) is advanced by a fixed stride. Unless otherwise stated, we use stride~1 as the primary setting, corresponding to dense online inference in which the model updates its prediction after each incoming event. To examine whether the results are sensitive to conventional HAR overlap choices, we additionally evaluate stride~50 and stride~100 in Appendix~\ref{app:raw_stream_stride}, corresponding to 50\% overlap and non-overlapping windows, respectively.

Each window is labeled by the activity associated with the most recent annotated event in the window. Thus, the task is latest-activity recognition rather than majority-window classification. Ground-truth activity boundaries are not used to determine test-window locations; they are used only after extraction to assign evaluation labels and compute post-hoc purity statistics.

The raw-stream protocol follows the same day-based chronological split as Section~\ref{sec:day_based_protocol}. For this protocol, all methods are trained on pure sliding windows from the training days with \(w=100\) and step size~50, and evaluated on the same raw-stream test windows.
% \end{revblock}

\subsection{Baselines}
To evaluate the overall effectiveness of the proposed method, we begin by comparing our method against several state-of-the-art ADL modeling algorithms.
\begin{itemize}[leftmargin=*]
\item[$\bullet$] \textbf{DeepCASAS} \cite{liciotti2020sequential} maps sensor event indices into a continuous embedding space and processes them via an LSTM-based architecture to capture temporal dependencies. We utilize the \texttt{CascadeEnsembleLSTM} variant, following the original implementation's architectural configuration and hyperparameters for activity classification.

\item[$\bullet$] \textbf{DCNN} \cite{gochoo2018unobtrusive} transforms sensor sequences into binary images where axes represent chronological order and sensor IDs to capture spatial co-activation patterns. These images are processed through a deep convolutional neural network with three convolutional and three fully connected layers as specified in the original design.

\item[$\bullet$] \textbf{TCN} \cite{bai2018empirical} utilizes stacked 1D dilated convolutions and residual connections to efficiently model both short- and long-range dependencies in embedded sensor data. The architecture employs exponentially increasing dilation factors and a final fully connected layer for classification, adhering to standard configurations.
 
\item[$\bullet$] \textbf{CPC}\cite{haresamudram2024towards, haresamudram2022contrastive, oord2019representationlearningcontrastivepredictive} employs a two-stage protocol involving a convolutional encoder and a GRU trained via self-supervised InfoNCE loss. The pre-trained model is subsequently fine-tuned with a classification head on labeled data for activity recognition, following the implementation described in TDOST~\cite{thukral2025layout}.
\item[$\bullet$] \textbf{TDOST}\cite{thukral2025layout} converts sensor events into natural language narratives that encode spatial layout and semantic information before embedding them with a pre-trained text encoder. A BiLSTM then models the temporal dependencies across these narrative representations to perform activity classification using the TDOST-basic variant.
\end{itemize}

\begin{table}[htbp]
\centering
\caption{Comparison of baseline methods and LastAct in terms of temporal modeling, spatial representation, and segmentation strategies.}
\resizebox{\textwidth}{!}{
\begin{tabular}{l|p{3.5cm}|p{3.2cm}|p{3.5cm}|p{8.5cm}}
\toprule
\textbf{Method} & \textbf{Temporal Info} & \textbf{Spatial Info} & \textbf{Segmentation Strategy} & \textbf{Details / Examples} \\
\midrule
DeepCasas 
& Sequential order only 
& None 
& Activity-first, then fixed-length window 
& Uses recurrent layers to model long-range temporal dependencies from event sequences. \\

DCNN 
& Sequential order only 
& Artificial spatial 
& Activity-first, then fixed-length window 
& Converts event sequences into 2D grids where x = event order and y = sensor ID; does not use physical floor layouts. \\

TCN 
& Sequential order only 
& None 
& Sliding window on event stream 
& Employs dilated 1D convolutions to expand the temporal receptive field for long-term context modeling. \\

TDOST 
& Timestamp, weekday, duration (symbolic) 
& Textual semantic location (e.g., ``kitchen'', ``bathroom'') 
& Activity-first, select first 100 events per activity 
& Generates natural-language-like descriptions such as: ``In the morning of Thursday, there was no detected motion in the bathroom.'' \\

\textbf{LastAct (Ours)} 
& Timestamp + weekday/hour embeddings 
& Real spatial (floorplan-based) 
& Window with controlled purity spectrum 
& Constructs sensor activation trajectories grounded in physical floor layouts and jointly models spatial and temporal dynamics with learned embeddings. \\
\bottomrule
\end{tabular}
}
\label{tab:baseline_comparison}
\end{table}

Table~\ref{tab:baseline_comparison} summarizes the key modeling and segmentation strategies across baselines and our method. The examined baselines adopt distinct modeling choices for smart-home activity recognition, primarily differing in how they encode temporal information, spatial context, and window construction. DeepCASAS and TCN operate directly on sequential sensor event logs and capture temporal dependencies using event order only (i.e., without learned embeddings for timestamps or durations): DeepCASAS relies on recurrent networks to model long-range dependencies, whereas TCN uses dilated temporal convolutions to expand the receptive field while enabling parallel computation. DCNN converts event streams into image-like inputs but uses an artificial spatial layout (event order × sensor ID) that facilitates convolution yet is not grounded in the physical floorplan or sensor coordinates; it also primarily encodes binary sensor states rather than richer or continuous signals. TDOST transforms event streams into natural-language-like text, encoding temporal variables (timestamp, weekday, duration) symbolically and injecting coarse location semantics (e.g., kitchen, bathroom) alongside state changes, prioritizing interpretability over learned continuous spatiotemporal embeddings. We additionally include a CPC-based baseline following TDOST, where contrastive predictive coding learns continuous latent representations from event sequences via self-supervision before supervised fine-tuning; unlike TDOST, CPC does not rely on explicit symbolic temporal or spatial descriptors. Across segmentation strategies, most prior work is activity-first or applies sliding windows only within pre-segmented activity episodes, which largely prevents activity transitions from appearing within individual windows. In contrast, our method operates on continuous event streams, grounds sensor activations on real floorplans, learns continuous temporal embeddings (hour-of-day and day-of-week), and explicitly controls window purity by regulating the proportion of the dominant (last) activity to systematically evaluate robustness under varying degrees of boundary mixing. 

Implementation details can be found in Appendix~\ref{app:implementation}.

\section{Experiments}
We evaluate the efficiency and robustness of LastAct through a series of experiments designed to mirror the complexities of real-world smart-home deployments. Our evaluation is organized around the following five research questions:

\begin{itemize}
\item \textbf{Q1: Overall recognition under mixed (pure + cross) windows.}
How well does LastAct perform when evaluated on realistic sliding windows that contain both pure and cross-activity segments, compared to strong sequence baselines?

\item \textbf{Q2: Streaming evaluation without ground-truth boundaries.} How does LastAct perform under a deployment-realistic protocol that applies fixed-stride sliding windows directly to the raw event stream, independent of ground-truth episode boundaries?

\item \textbf{Q3: Sensitivity to activity purity and contamination.}
How does recognition performance vary across activity-purity buckets (e.g., 10--30\%, 40--60\%, 70--90\%, 100\%), and to what extent can LastAct maintain accuracy when the target activity occupies only a small fraction of the window?

\item \textbf{Q4: Backbone strength under standard protocols.} 
When isolated from the streaming framework, how strong is the trajectory-based activity recognition backbone under the conventional pure-train / pure-test protocol used in prior work?

\item \textbf{Q5: Boundary detection quality.} 
How accurately can LastAct localize the last activity boundary in a window, and how closely do detected boundaries match ground-truth ones in both quantitative metrics and qualitative examples?

\item \textbf{Q6: Generalization to temporal and spatial shifts.} 
To what extent does the method retain performance when evaluated under distribution shifts, such as chronological day-based splits and perturbations to sensor locations in the floor plan?

\item \textbf{Q7: Contribution of architectural and loss components.} 
Which components of the framework—trajectory representation capacity and boundary-regularization losses—are most critical to the observed gains, and how sensitive is performance to their design choices?
\end{itemize}

% Finally, to address \textbf{Q4}, we assess the system’s stability and generalization via a suite of robustness analyses. Specifically, we vary the window size and inject synthetic sensor noise to emulate real-world sensor instability and environmental interference. In addition, we use a chronological train–test split to capture realistic distribution shifts. Unlike a randomized split, this setup allows us to explicitly quantify robustness to long-term behavioral drift. Collectively, these stress tests characterize the method’s reliability under common perturbations and verify if the performance still remains stable across diverse deployment scenarios.

\subsection{Overall Performance on Mixed Pure–Cross Windows (Q1)}
\label{sec: overall perf}
% We benchmark the method using a sliding window protocol across multiple smart home datasets and compare it against state-of-the-art segment-based baselines. More specifically, we systematically vary the window sizes and activity purity to assess the recognition performance on practical deployment conditions. By reporting both Accuracy and Macro-F1, we evaluate robustness in continuous, unsegmented sensor streams representing real-world ambient assisted living deployments.

% % \paragraph{Experiment setup.}
% We use a 3-fold stratified split to ensure the model generalizes to . We evaluate five window sizes, $W\in{20,40,60,80,100}$, to quantify the effect of temporal granularity. Performance is reported using Accuracy and Macro-F1, with Macro-F1 as the primary metric given the pronounced class imbalance in real-world activity logs. We benchmark against several state-of-the-art HAR baselines, including DeepCASAS~\cite{} for sequential modeling, TDOST~\cite{} for text-based representations, and deep architectures such as DCNN~\cite{}, TCN~\cite{}, and CPC~\cite{}.
In this experiment, all baselines are trained only on pure windows and evaluated on a mixed test set (pure + cross). We enforce the same pure-window constraint on our activity backbone, while training the auxiliary gate and boundary detector on transition windows with purity labels and boundary annotations for boundary-aware inference.
\begin{table*}[htbp]
\centering
\caption{Performance on mixed (pure + cross) windows at window sizes $w=20, 40, 60, 80, 100$ (Accuracy / Macro-F1 \%) across datasets. Best and second-best per row (within each window size) are highlighted in \textbf{bold} and \underline{underline}, respectively.}
\label{tab:window_results}
\resizebox{0.85\textwidth}{!}{
\begin{tabular}{cc|cccccc}
\hline
\textbf{Dataset} & \textbf{Window Size} 
& DeepCASAS 
& DCNN 
& TCN 
& CPC 
& TDOST 
& \textbf{Ours} \\
\hline
\multirow{5}*{\textbf{Milan}} & 20  & 42.98 / 32.74 & \underline{47.15 / 33.68} & 39.90 / 27.92 & 39.61 / 29.01 & 40.24 / 29.48 & \textbf{63.40 / 48.20} \\
 & 40  & 41.73 / 31.96 & \underline{41.92 / 30.85} & 38.95 / 30.45 & 36.43 / 28.62 & 36.29 / 27.66 & \textbf{70.01 / 52.03} \\
 & 60  & 42.22 / 30.62 & \underline{45.18 / 32.36} & 42.21 / 32.61 & 37.74 / 28.51 & 38.98 / 30.07 & \textbf{68.25 / 51.24} \\
 & 80  & \underline{45.50 / 32.40} & 43.68 / 31.12 & 42.56 / 33.11 & 39.09 / 30.01 & 37.22 / 30.53 & \textbf{69.94 / 52.15} \\
 & 100  & 42.68 / 32.07 & 43.17 / 31.03 & \underline{44.19 / 34.58} & 41.28 / 31.04 & 38.49 / 30.66 & \textbf{70.09 / 51.50} \\ \hline
\multirow{5}*{\textbf{Aruba}} & 20  & \underline{56.37 / 38.33} & 50.86 / 33.04 & 51.05 / 33.78 & 52.98 / 34.51 & 49.63 / 31.84 & \textbf{79.51 / 57.62} \\
 & 40  & \underline{56.77 / 39.84} & 52.65 / 36.49 & 55.94 / 38.81 & 52.11 / 36.15 & 43.91 / 30.38 & \textbf{82.41 / 59.83} \\
 & 60  & \underline{63.03 / 42.01} & 54.27 / 41.66 & 60.54 / 44.46 & 47.99 / 39.83 & 46.37 / 33.76 & \textbf{85.43 / 61.26} \\
 & 80  & 53.68 / 39.43 & 53.79 / 43.94 & \underline{63.19 / 47.74} & 51.15 / 42.78 & 35.86 / 30.68 & \textbf{83.83 / 60.38} \\
 & 100  & 56.05 / 45.13 & 55.26 / 46.25 & \underline{61.75 / 49.75} & 48.72 / 45.78 & 38.27 / 33.96 & \textbf{83.34 / 62.65} \\ \hline
\multirow{5}*{\textbf{Kyoto}} & 20  & \underline{38.48 / 31.34} & 36.16 / 29.36 & 30.30 / 22.64 & 27.79 / 23.44 & 32.19 / 25.38 & \textbf{56.42 / 51.05} \\
 & 40  & \underline{41.99 / 35.35} & 36.12 / 29.90 & 32.99 / 26.46 & 30.78 / 26.93 & 34.33 / 27.18 & \textbf{61.20 / 56.26} \\
 & 60  & \underline{47.16 / 39.04} & 35.57 / 28.12 & 34.86 / 29.15 & 28.29 / 24.36 & 32.05 / 26.39 & \textbf{63.51 / 58.87} \\
 & 80  & \underline{48.52 / 38.92} & 35.79 / 27.73 & 35.12 / 30.08 & 30.15 / 26.06 & 35.30 / 28.37 & \textbf{62.11 / 53.32} \\
 & 100  & \underline{51.27 / 38.80} & 36.33 / 27.80 & 33.98 / 29.08 & 35.21 / 29.25 & 36.87 / 29.57 & \textbf{62.73 / 54.85} \\ \hline
\multirow{5}*{\textbf{Orange}} & 20  & 48.81 / 43.61 & 47.39 / 38.36 & 34.98 / 18.84 & 43.99 / 32.95 & \textbf{59.16} / \underline{52.82} & \underline{57.67} / \textbf{55.15} \\
 & 40  & 38.12 / 24.13 & 47.56 / 41.48 & 41.04 / 18.51 & 38.98 / 35.29 & \textbf{59.90 / 55.71} & \underline{57.18 / 55.01} \\
 & 60  & 32.94 / 22.02 & 38.43 / 32.68 & 36.99 / 18.58 & 33.56 / 30.04 & \underline{51.86 / 51.89} & \textbf{59.55 / 56.67} \\
 & 80  & 31.66 / 19.39 & 37.58 / 33.53 & 33.90 / 18.21 & 35.46 / 29.18 & \underline{53.90 / 54.16} & \textbf{59.57 / 55.85} \\
 & 100  & 31.67 / 23.67 & 36.11 / 30.61 & 32.66 / 17.14 & 33.50 / 24.39 & \underline{51.73 / 50.68} & \textbf{61.70 / 61.65} \\ \hline
\end{tabular}
}
\end{table*}

Table~\ref{tab:window_results} summarizes results on all mixed windows for each dataset and window size. Across Milan, Aruba, and Kyoto, our method consistently outperforms all baselines by large margins, often exceeding them by 15--25 points in Macro-F1 for moderate and large $w$. These three datasets expose one of the core challenges in ambient HAR: the raw event identifiers are low-semantic codes (e.g., \texttt{M01}) and the original streams do not explicitly provide spatial relationships between sensors. In this regime, architectures that operate directly on symbolic IDs or short local fragments (DeepCASAS, DCNN, TCN, CPC, and even TDOST when its textual templates are derived from generic sensor labels) struggle to extract discriminative patterns from mixed windows. In contrast, our trajectory-based backbone explicitly reconstructs layout-aligned sensor trajectories and feeds them to the recognition model, recovering the underlying spatial structure that is missing from the event IDs. The large
gaps on Milan/Aruba/Kyoto indicate that spatial trajectories are particularly valuable when the original sensor naming carries little semantic signal.

The behavior on Orange is complementary. Here, sensor identifiers and metadata are substantially more informative (e.g., \texttt{kitchen\_oven\_power}, \texttt{bathroom\_sink\_water}), and TDOST can exploit these rich textual cues to obtain strong performance, especially at smaller window sizes. At $w=40$, TDOST slightly outperforms our model in Macro-F1, suggesting that fine-grained, semantically labeled sensors combined with sentence-level descriptions are highly effective for very local context. However, as $w$ increases and windows accumulate more cross-activity events, TDOST becomes more sensitive to boundary contamination because it encodes the entire window as a single sentence sequence without an explicit mechanism to isolate the most relevant region. Our method, in contrast, leverages layout-aligned trajectories together with boundary-aware inference to focus on the latest activity segment. As a result, its Macro-F1 steadily improves with $w$ and surpasses TDOST by a clear margin at $w\ge 60$, turning additional temporal context from a liability into a benefit.

Across multiple datasets and window sizes, DCNN is the strongest non-sequential baseline and often ranks second to our method, consistent with its sensor–time image encoding. Nevertheless, our trajectory-based model achieves higher Macro-F1, suggesting that layout-aligned trajectories and boundary-aware inference capture spatial structure more effectively than generic image-style window encodings.

Overall, this experiment shows that our method is robust to realistic mixed-activity conditions across a wide range of window sizes. On datasets with weakly informative sensor IDs, trajectory reconstruction closes the semantic gap left by generic identifiers, while on the Orange dataset, where sensor semantics are strong and text-based modeling is highly competitive, our boundary-aware trajectory approach remains competitive at short windows and becomes clearly superior as longer horizons are considered. These results support our central claim that combining layout-aligned trajectories with boundary-focused inference is a principled way to handle boundary contamination in ADL.

% \begin{revblock}
\subsection{Raw-Stream Sliding-Window Evaluation (Q2)}
\label{sec:raw_stream_eval}
The controlled mixed-window protocol in Section~\ref{sec: overall perf} isolates the effect of boundary contamination by explicitly constructing windows at predefined purity levels. We next evaluate whether the same conclusion holds when windows are extracted directly from the continuous event stream, following the raw-stream protocol defined in Section~\ref{sec:raw_stream_protocol}. This experiment removes the use of ground-truth activity boundaries for test-window construction and therefore provides a complementary evaluation of latest-activity recognition under deployment-like sliding-window inference.

In the main raw-stream setting, we use window size \(w=100\) and stride~1, so the model produces a prediction after every incoming sensor event. Each window is labeled by the latest annotated activity in the window, rather than by the majority activity. Therefore, the task remains consistent with our problem formulation: recognizing the current activity at the newest event in the observation window. Stride~50 and stride~100 results are reported in Appendix~\ref{app:raw_stream_stride} as overlap-sensitivity analyses, corresponding to 50\% overlap and non-overlapping windows, respectively.

The empirical composition of the raw-stream test windows is reported in Appendix~\ref{app:raw_stream_purity_distribution}. These purity bins are computed only after window extraction for analysis and are not used to construct the windows or provided to the model. The distribution differs substantially across datasets. Milan and Aruba contain a large fraction of cross-activity windows, with pure windows accounting for only 44.4\% and 43.9\% of test windows, respectively. Kyoto contains a smaller but still substantial proportion of cross windows, while Orange is dominated by pure windows because many activities are long relative to \(w=100\). This variation highlights why raw-stream evaluation is complementary to the controlled purity protocol: instead of enforcing a balanced contamination spectrum, it preserves the natural mixture of pure and cross windows induced by each home.

% \begin{table}[t]
% \centering
% \caption{Empirical distribution of raw-stream test windows across purity bins. Counts are pooled over the day-based chronological test split at stride~1. \textit{Pure} denotes windows entirely contained in a single annotated episode; remaining cross windows are stratified by purity (the fraction of in-target events within the window). The aggregate Macro-F1 in Table~\ref{tab:aggregate_macrof1} is implicitly weighted by these frequencies.}
% \label{tab:purity_bins}
% \resizebox{0.85\textwidth}{!}{%
% \begin{tabular}{lcccccccr}
% \toprule
% Dataset & Pure & 0--20 & 21--40 & 41--60 & 61--80 & 81--99 & Total \\
% \midrule
% Aruba  & 64,711 (43.9\%)  & 15,403 (10.4\%) & 15,380 (10.4\%) & 17,445 (11.8\%) & 20,621 (14.0\%) & 13,892 (9.4\%) & 147,452 \\
% Milan  & 37,284 (44.4\%)  & 11,737 (14.0\%) & 9,113 (10.8\%)  & 8,751 (10.4\%)  & 8,807 (10.5\%)  & 8,352 (9.9\%)  & 84,044  \\
% Kyoto  & 18,644 (66.1\%)  & 2,029 (7.2\%)   & 1,941 (6.9\%)   & 2,235 (7.9\%)   & 1,807 (6.4\%)   & 1,559 (5.5\%)  & 28,215  \\
% Orange & 121,465 (93.8\%) & 1,694 (1.3\%)   & 1,493 (1.2\%)   & 1,331 (1.0\%)   & 1,116 (0.9\%)   & 2,393 (1.8\%)  & 129,492 \\
% \bottomrule
% \end{tabular}%
% }
% \label{tab:raw-stream_purity_distribution}
% \end{table}

\begin{table}[t]
\centering
\caption{Aggregate Macro-F1 (\%) on the raw-stream test set. All methods are trained on pure windows with window size $w{=}100$ and step size $50$. Test windows are extracted at stride~1 from the day-based chronological test split, with no reference to ground-truth episode boundaries during construction. Macro-F1 is computed once on the pooled set of all raw-stream test windows, implicitly weighted by the empirical purity-bin frequency on each home (Table~\ref{tab:raw_stream_stride_distribution}). Best per dataset in \textbf{bold}; mean~$\pm$~std over three runs.}
\label{tab:aggregate_macrof1}
\resizebox{0.45\textwidth}{!}{%
\begin{tabular}{lcccc}
\toprule
Method & Aruba & Milan & Kyoto & Orange \\
\midrule
DeepCASAS & 38.6 $\pm$ 0.5 & 36.7 $\pm$ 0.3 & 41.0 $\pm$ 1.8 & 40.9 $\pm$ 0.9 \\
DCNN      & 36.4 $\pm$ 0.7 & 35.6 $\pm$ 1.0 & 34.2 $\pm$ 0.6 & 44.7 $\pm$ 1.9 \\
TCN       & 37.0 $\pm$ 0.9 & 34.4 $\pm$ 0.9 & 36.6 $\pm$ 2.0 & 26.3 $\pm$ 1.4 \\
CPC       & 38.4 $\pm$ 0.5 & 32.7 $\pm$ 0.7 & 35.5 $\pm$ 1.5 & 53.1 $\pm$ 2.5 \\
TDOST     & 36.5 $\pm$ 0.1 & 34.9 $\pm$ 0.5 & 42.5 $\pm$ 1.2 & 63.1 $\pm$ 0.6 \\
\midrule
Ours      & \textbf{44.6 $\pm$ 1.9} & \textbf{40.1 $\pm$ 0.5} & \textbf{48.6 $\pm$ 8.5} & \textbf{72.9 $\pm$ 7.8} \\
\bottomrule
\end{tabular}%
}
\label{tab:raw_stream_aggregate}
\end{table}

Table~\ref{tab:raw_stream_aggregate} reports aggregate activity-level Macro-F1 on the stride-1 raw-stream test set. Macro-F1 is computed once after pooling all raw-stream test windows, regardless of purity, and then macro-averaging over activity classes. Thus, the metric preserves the empirical mixture of pure and cross windows while reducing dominance by frequent activities.

LastAct achieves the highest aggregate Macro-F1 on all four datasets. On Aruba, LastAct improves over the strongest baseline from 38.6\% to 44.6\%; on Milan, from 36.7\% to 40.1\%; on Kyoto, from 42.5\% to 48.6\%; and on Orange, from 63.1\% to 72.9\%. These gains are smaller than those observed under the controlled low-purity protocol because the raw-stream test set contains many high-purity or fully pure windows, especially on Orange. Nevertheless, the consistent improvement across all datasets shows that the proposed boundary-aware inference remains beneficial when test windows are obtained directly from the event stream rather than instantiated around annotated activity transitions.

To further interpret these aggregate results, we provide a post-hoc purity-stratified analysis in Appendix~\ref{app:raw_stream_purity_performance}. The trend is consistent with the controlled mixed-window evaluation: LastAct provides the largest gains in low- and mid-purity windows, where the latest activity occupies only a small portion of the observation window and stale pre-boundary context is most misleading. As window purity increases, the gap narrows and some baselines become competitive, particularly in fully pure windows where the task reduces to conventional single-activity recognition. This behavior is expected because the boundary detector is not intended as an end task by itself; rather, it is an auxiliary mechanism for identifying the most recent activity segment when the window contains evidence from multiple activities.

Overall, the raw-stream evaluation strengthens the central claim of this work. The controlled mixed-window protocol shows how recognition performance changes as boundary contamination is varied systematically, while the raw-stream protocol shows that LastAct remains effective when test windows are extracted directly from the continuous stream. Together, the two protocols demonstrate that trajectory-guided boundary-aware inference improves latest-activity recognition both under controlled contamination stress tests and under deployment-like sliding-window evaluation where activity boundaries are unavailable during window construction.

% \end{revblock}

\subsection{Recognition Performance across Activity-Purity Buckets (Q3)}
\label{sec:purity_ana}
To better interpret the aggregate results reported in Section~\ref{sec: overall perf}, further analyze the case of window size $w=100$ by stratifying test windows into four activity-purity buckets: 10--30\%, 40--60\%, 70--90\%, and 100\%. 
% These buckets correspond to qualitatively different recognition regimes. The 10--30\% range reflects the most challenging setting, where only a small fraction of the window belongs to the target activity and most events are competing context from other activities. The 40--60\% range captures moderate mixtures, in which the target activity and other activities contribute comparable amounts of evidence. The 70--90\% range represents near-pure windows that still contain preceding or trailing events as noise, while the 100\% bucket corresponds to idealized pure segments. 
This stratification allows us to disentangle model behavior under severely truncated, moderately mixed, near-pure, and fully pure conditions at a fixed temporal granularity.

In this analysis we compare our method against three representative baselines. TDOST is a state-of-the-art text-based method, leveraging sentence-level descriptions of sensor events. DCNN instantiates an alternative image-style encoding of windows and is often the best non-text baseline in our overall results. DeepCASAS serves as a classic baseline for smart-home ADL recognition. The complete purity-stratified results for all baselines in Appendix~\ref{app:purity_full_results}.

\begin{figure*}[htbp]
    \centering
    \includegraphics[width=0.90\textwidth]{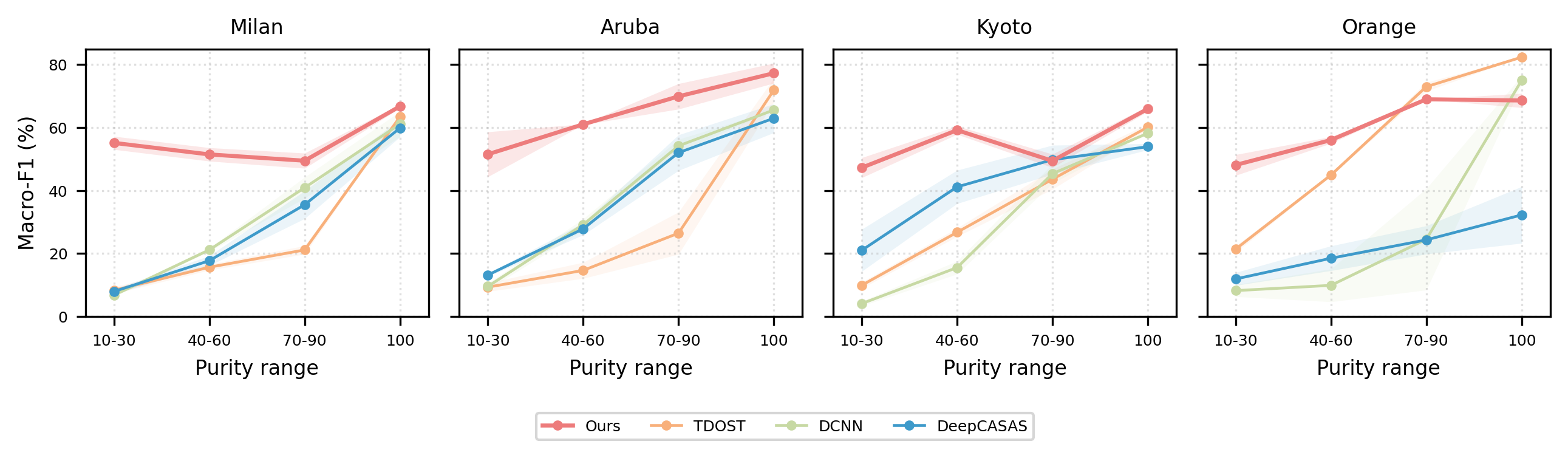}
    \caption{Recognition performance under different activity-purity ranges at window size $w=100$. Results are reported as Macro-F1 (mean $\pm$ standard deviation) over 3-fold cross-validation. Lower-purity windows contain stronger boundary contamination, where only a small fraction of events belongs to the latest target activity. The comparison shows that LastAct provides the largest gains in heavily mixed windows by localizing the latest activity segment and suppressing stale pre-boundary context, while remaining competitive as windows become near-pure or fully pure.}   
    % Cross-dataset comparison under different activity purity ranges. Results are reported as Macro-F1 (mean $\pm$ std) over 3-fold cross-validation.
    
    \label{fig:purity_range}
\end{figure*}
Figure~\ref{fig:purity_range} compares our method with baselines across different window purity bins, highlighting robustness under varying contamination levels. In heavily contaminated settings (10–30\%), where target activity signals are minimal, most baselines collapse to below 20\% Macro-F1 on Milan, Aruba, and Orange, while our method consistently achieves 47\%–55\%, showing substantial gains. This advantage persists in the 40–60\% range, with improvements of 11\%–28\%, demonstrating the effectiveness of boundary-aware inference in handling mixed-activity windows. As purity increases, the performance gap narrows. In the 70–90\% range, our method remains competitive but is surpassed by strong baselines on some datasets (e.g., TDOST on Orange). In fully clean windows (100\%), methods such as CPC and TDOST outperform ours, as the task shifts from boundary-sensitive recognition to modeling complete activity patterns.

Dataset-specific trends further support this observation: on Milan and Aruba, where sensor semantics are limited, baselines struggle under contamination, whereas on Orange, richer semantic signals enable models like TDOST to excel in high-purity settings. Overall, our method delivers significant improvements in low- and mid-purity regimes while maintaining competitive performance when windows are clean. 
\subsection{Trajectory-based Activity Backbone under Standard Pure-Window Protocol (Q4)}
After establishing the benefits of our framework under near-realistic mixed-activity conditions, we assess the intrinsic strength of the activity recognition backbone, whose trajectory encoder is shared by the pure/cross gate and boundary detector. We therefore isolate the backbone and evaluate it in the conventional pure-window setting used in prior work, training and testing only on single-activity windows. This experiment measures the intrinsic recognition capacity of the spatial trajectory encoder relative to the baselines.

\begin{table*}[htbp]
\centering
\caption{Activity recognition on \emph{pure} windows under the standard pure-train / pure-test protocol.
Entries are Macro-F1 mean $\pm$ std (\%) across runs. Best per row is in bold, second-best is underlined.}
\label{tab:pure_window_results_f1only}
\resizebox{0.85\textwidth}{!}{%
\begin{tabular}{cc|cccccc}
\hline
\textbf{Dataset} & \textbf{Window Size}
& DeepCASAS 
& DCNN 
& TCN 
& CPC 
& TDOST 
& \textbf{Ours (Stage~1)} \\
\hline

\multirow{5}{*}{\textbf{Milan}}
& 20  & 67.89$\pm$4.87 & 66.99$\pm$4.86 & \underline{69.86$\pm$2.45} & 69.20$\pm$2.12 & 69.29$\pm$2.03 & \textbf{71.67$\pm$1.44} \\
& 40  & 68.86$\pm$3.62 & 70.98$\pm$3.23 & 72.04$\pm$3.14 & \underline{72.75$\pm$1.96} & \textbf{74.16$\pm$2.46} & 72.67$\pm$2.89 \\
& 60  & 63.95$\pm$6.42 & 69.69$\pm$3.98 & 74.29$\pm$3.78 & 73.91$\pm$2.94 & \textbf{75.88$\pm$1.64} & \underline{74.36$\pm$3.89} \\
& 80  & 61.24$\pm$10.13 & 66.50$\pm$4.26 & 72.38$\pm$3.32 & \underline{72.93$\pm$1.88} & \textbf{73.04$\pm$1.85} & 72.76$\pm$2.73 \\
& 100 & 61.76$\pm$9.18 & 67.51$\pm$3.69 & 73.20$\pm$4.34 & 73.25$\pm$1.02 & \textbf{73.65$\pm$3.80} & 71.52$\pm$0.09 \\
\hline

\multirow{5}{*}{\textbf{Aruba}}
& 20  & 75.47$\pm$2.85 & 74.78$\pm$2.89 & 74.52$\pm$0.84 & 74.49$\pm$1.04 & \underline{75.52$\pm$0.87} & \textbf{76.24$\pm$1.94} \\
& 40  & 72.22$\pm$3.54 & 72.30$\pm$0.49 & \textbf{77.29$\pm$1.95} & \underline{76.90$\pm$3.69} & 76.06$\pm$0.67 & 76.27$\pm$2.42 \\
& 60  & 66.14$\pm$9.93 & 71.12$\pm$1.34 & \textbf{76.45$\pm$2.65} & 75.00$\pm$1.01 & 75.73$\pm$0.94 & \underline{75.98$\pm$1.15} \\
& 80  & 65.40$\pm$8.71 & 70.40$\pm$0.29 & 75.33$\pm$1.27 & \textbf{76.99$\pm$4.12} & 75.92$\pm$2.33 & \underline{76.43$\pm$1.18} \\
& 100 & 60.31$\pm$14.49 & 68.37$\pm$4.27 & \textbf{76.55$\pm$2.37} & 74.31$\pm$1.04 & 75.27$\pm$1.25 & \underline{75.96$\pm$1.18} \\
\hline

\multirow{5}{*}{\textbf{Kyoto}}
& 20  & \underline{59.96$\pm$3.06} & 51.64$\pm$4.77 & 59.64$\pm$6.19 & 54.70$\pm$4.21 & 53.13$\pm$1.26 & \textbf{64.14$\pm$5.38} \\
& 40  & 51.79$\pm$3.28 & 55.86$\pm$3.23 & 57.40$\pm$6.08 & \underline{57.78$\pm$4.69} & 53.46$\pm$2.09 & \textbf{70.92$\pm$3.01} \\
& 60  & 49.57$\pm$6.68 & 58.29$\pm$3.50 & 61.60$\pm$6.97 & \underline{63.15$\pm$1.35} & 52.17$\pm$7.52 & \textbf{69.01$\pm$2.32} \\
& 80  & 50.52$\pm$8.19 & 55.40$\pm$6.41 & 57.63$\pm$8.10 & \underline{62.53$\pm$6.12} & 59.11$\pm$2.89 & \textbf{65.96$\pm$5.26} \\
& 100 & 46.36$\pm$9.65 & 54.23$\pm$5.64 & \underline{58.94$\pm$9.13} & 56.87$\pm$4.85 & 57.34$\pm$1.28 & \textbf{68.22$\pm$2.85} \\
\hline

\multirow{5}{*}{\textbf{Orange}}
& 20  & 49.51$\pm$7.22 & 58.66$\pm$7.21 & 23.34$\pm$1.19 & 54.74$\pm$6.58 & \textbf{69.89$\pm$2.09} & \underline{63.79$\pm$3.36} \\
& 40  & 29.08$\pm$7.56 & 69.29$\pm$2.01 & 28.46$\pm$2.32 & 59.86$\pm$9.14 & \textbf{73.97$\pm$3.82} & \underline{71.00$\pm$2.50} \\
& 60  & 24.94$\pm$5.61 & 67.01$\pm$3.63 & 29.95$\pm$7.63 & 63.73$\pm$2.21 & \textbf{80.25$\pm$1.08} & \underline{68.88$\pm$0.83} \\
& 80  & 24.54$\pm$4.95 & 69.81$\pm$4.17 & 25.67$\pm$0.45 & 60.88$\pm$5.72 & \textbf{80.01$\pm$4.19} & \underline{74.28$\pm$3.43} \\
& 100 & 24.50$\pm$3.83 & 69.37$\pm$6.56 & 28.38$\pm$8.90 & 58.90$\pm$7.81 & \textbf{79.00$\pm$3.01} & \underline{73.99$\pm$0.29} \\
\hline

\end{tabular}}
\end{table*}

Table~\ref{tab:pure_window_results_f1only} reports Macro-F1 (mean~$\pm$~std) across window sizes $w\in\{20,40,60,80,100\}$ and datasets. Overall, our trajectory-based backbone is consistently competitive, frequently ranking best or second-best against baseline methods. On \textbf{Milan} and \textbf{Aruba}, where all models operate on the same basic modalities (motion, door, temperature sensors), our backbone either matches or slightly exceeds the strongest baseline in most configurations, with gaps typically within $\approx$1--2 F1 points and comparable variability. On \textbf{Kyoto}, which uses the same sensor types but with a considerably denser deployment (more nodes and finer spatial coverage), the advantage of trajectory encoding becomes more pronounced: our model achieves the best F1 at all window sizes, outperforming the strongest baseline by up to roughly 10--13 F1 points, indicating that richer spatial structure is well captured by our representation.

The \textbf{Orange} dataset poses a different challenge. Although it also contains many sensors, it is heavily imbalanced and spans fewer days, making pure-window recognition intrinsically difficult. Moreover, TDOST can exploit descriptive sensor names (e.g., \emph{kitchen door sensor}) rather than opaque IDs (e.g., \emph{M01}), effectively injecting semantic prior knowledge about rooms and objects. As a result, TDOST attains the highest Macro-F1 on most window sizes. Nonetheless, our backbone remains competitive on Orange: it is consistently second-best or even best (e.g., at $w=80$), and the performance gap to TDOST is generally modest despite not using any textual information.

Overall, the pure-window study confirms that our trajectory-based recognizer is a strong standalone classifier under the conventional setting. The additional improvements in near-realistic mixed-activity experiments therefore stem primarily from the \emph{framework-level} components—pure/cross gating and boundary-aware inference—rather than merely substituting a stronger backbone.

\subsection{Boundary Detection Performance (Q5)}

\paragraph{Quantitative last-boundary localization.}
% Table~\ref{tab:last_boundary_metrics_wide} summarizes the performance of the boundary detector for different window sizes $W$ on all four datasets.
% We report the accuracy of localizing the last boundary within $\pm 2$ index positions inside each window (Last-bound Acc), together with the mean absolute error MAE $=\mathbb{E}[|p_\text{last}-g_\text{last}|]$ and the signed bias $\text{mean}\Delta=\mathbb{E}[p_\text{last}-g_\text{last}]$ in index units (position offsets within the $W$-length window).
Table~\ref{tab:last_boundary_metrics_wide} summarizes the performance of the boundary detector for different window sizes $w$ on all four datasets. For each window, let $g_{\text{last}} \in \{1,\dots,w\}$ denote the index of the last ground-truth boundary inside the window and $p_{\text{last}}$ the index of the last predicted boundary (after thresholding and non-maximum suppression). \emph{Last-bound Acc} is the fraction of windows for which $|p_{\text{last}} - g_{\text{last}}| \le 2$. We also report the mean absolute error $\mathrm{MAE} = \mathbb{E}[\,|p_{\text{last}} - g_{\text{last}}|\,]$ and the signed offset $\text{Mean}\,\Delta = \mathbb{E}[\,p_{\text{last}} - g_{\text{last}}\,]$ in index units (positions within the $w$-length window), where values of Mean~$\Delta$ close to zero indicate little systematic early/late bias.

Across all window sizes, the Last-bound accuracy remains relatively stable but exhibits dataset-dependent behavior. It ranges from $78$–$83\%$ on \emph{Aruba} and \emph{Kyoto}, while on \emph{Milan} it gradually decreases from $76.3\%$ to $69.6\%$ as window size increases. On the more challenging \emph{Orange} dataset, the detector achieves a lower accuracy between $39$–$56\%$. The corresponding MAE increases with window length, ranging from $1.31$ to $6.76$ on \emph{Aruba} and reaching up to $9.68$ on \emph{Milan} at $w{=}100$. In addition, the signed mean $\Delta$ on \emph{Aruba} remains consistently negative (down to $-3.44$), indicating a slight tendency to predict boundaries earlier than ground truth in this environment.

\begin{table}[htbp]
\centering
\caption{Boundary detector evaluation across window sizes $w$. 
Last-bound Acc is the accuracy of localizing the last boundary within $\pm 2$ index positions inside each window. 
Last-bound MAE and mean$\Delta$ are measured in index units (position offsets within the $w$-length window; positive $\Delta$ = late predictions).}
\label{tab:last_boundary_metrics_wide}
\setlength{\tabcolsep}{2pt}
\renewcommand{\arraystretch}{0.9}
\resizebox{\linewidth}{!}{%
\begin{tabular}{c|ccc|ccc|ccc|ccc}
\toprule
\multirow{2}{*}{$w$} 
& \multicolumn{3}{c|}{Aruba} 
& \multicolumn{3}{c|}{Kyoto} 
& \multicolumn{3}{c|}{Milan} 
& \multicolumn{3}{c}{Orange} \\
\cmidrule(lr){2-4} \cmidrule(lr){5-7} \cmidrule(lr){8-10} \cmidrule(lr){11-13}
& Acc$\uparrow$ & MAE$\downarrow$ & mean$\Delta$ 
& Acc$\uparrow$ & MAE$\downarrow$ & mean$\Delta$ 
& Acc$\uparrow$ & MAE$\downarrow$ & mean$\Delta$ 
& Acc$\uparrow$ & MAE$\downarrow$ & mean$\Delta$ \\
\midrule
20  & 81.17$\pm$0.95 & 1.31$\pm$0.09 & -0.66$\pm$0.07 
    & 81.55$\pm$2.15 & 1.56$\pm$0.15 & 0.49$\pm$0.20 
    & 76.32$\pm$1.64 & 2.06$\pm$0.13 & 0.55$\pm$0.17 
    & 51.72$\pm$4.48 & 4.00$\pm$0.34 & 0.41$\pm$0.06 \\
40  & 82.16$\pm$1.68 & 2.05$\pm$0.20 & -0.87$\pm$0.21 
    & 79.09$\pm$1.96 & 2.39$\pm$0.33 & -0.00$\pm$0.39 
    & 74.82$\pm$1.55 & 3.51$\pm$0.21 & 0.83$\pm$0.55 
    & 39.59$\pm$1.31 & 6.35$\pm$0.29 & 0.29$\pm$1.01 \\
60  & 83.62$\pm$0.23 & 2.96$\pm$0.07 & -1.25$\pm$0.50 
    & 78.88$\pm$1.62 & 3.81$\pm$0.19 & 0.23$\pm$0.67 
    & 70.84$\pm$1.22 & 5.95$\pm$0.06 & 0.74$\pm$0.45 
    & 49.60$\pm$3.87 & 8.29$\pm$0.77 & 1.25$\pm$1.44 \\
80  & 82.55$\pm$1.21 & 4.56$\pm$0.24 & -1.63$\pm$0.25 
    & 78.62$\pm$1.15 & 4.86$\pm$0.19 & 0.05$\pm$0.81 
    & 70.94$\pm$1.53 & 7.70$\pm$0.48 & 1.05$\pm$0.12 
    & 49.91$\pm$6.63 & 11.31$\pm$2.01 & 3.60$\pm$2.05 \\
100 & 78.82$\pm$0.42 & 6.76$\pm$0.13 & -3.44$\pm$0.37 
    & 80.00$\pm$3.64 & 6.07$\pm$1.30 & 0.56$\pm$1.17 
    & 69.63$\pm$0.92 & 9.68$\pm$0.41 & 1.82$\pm$0.58 
    & 56.62$\pm$2.73 & 11.49$\pm$0.54 & 1.10$\pm$3.12 \\
\bottomrule
\end{tabular}
}
\end{table}

\paragraph{Detected vs. ground-truth boundaries.}
% To assess whether such localization accuracy is sufficient for downstream recognition, we compare cross-window activity recognition using detected boundaries vs.\ using ground-truth(GT) boundaries.
% Figure~\ref{fig:boundary_oracle_f1_per_dataset} plots Macro-F1 on cross-activity windows as a function of $w$ for the two settings.

\begin{figure}[htbp]
    \centering
    \includegraphics[width=0.9\linewidth]{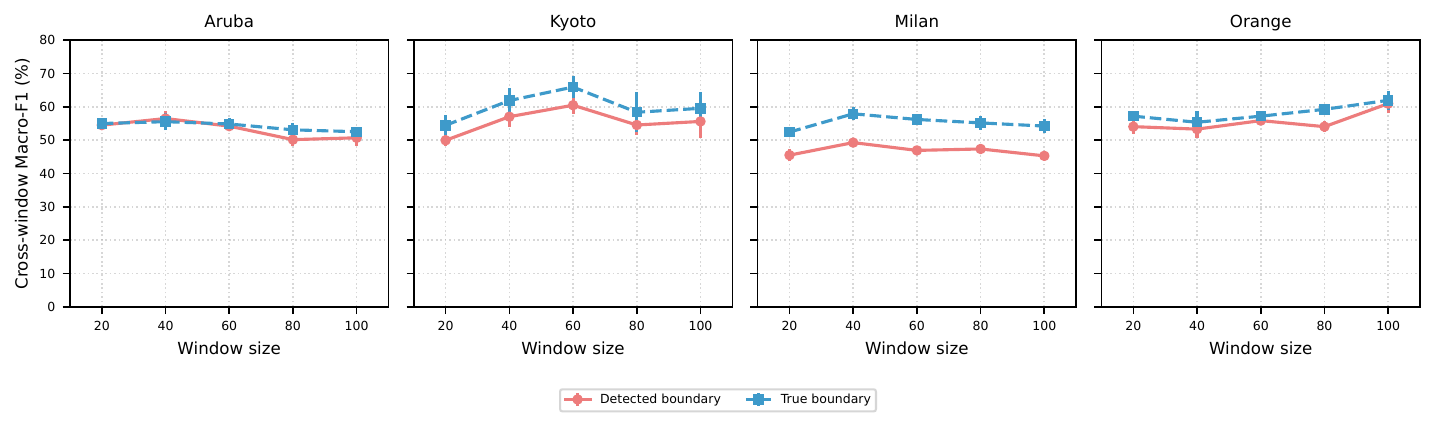}
    \caption{%
    Cross-window Macro-F1 with detected vs. true boundaries.
    }
    \label{fig:boundary_oracle_f1_per_dataset}
\end{figure}

To evaluate whether this localization accuracy is sufficient for downstream recognition, we compare cross-window activity recognition using detected boundaries against ground-truth (GT) boundaries. As shown in Figure~\ref{fig:boundary_oracle_f1_per_dataset}, the performance gap varies substantially across datasets. While Aruba, Kyoto, and Orange exhibit relatively small discrepancies, a pronounced degradation of approximately $7$–$10$ Macro-F1 points is observed on Milan.

To better understand this phenomenon, we analyze the activity length distributions (Figure~\ref{fig:activity_length_distribution} in Appendix~A.2). The \emph{Milan} dataset exhibits a significantly higher degree of activity fragmentation, with 37.4\% of activities containing fewer than 30 sensor events and a median duration of only 55, whereas \emph{Kyoto} contains substantially longer and more temporally stable activities (median = 270). In contrast, \emph{Orange} also contains relatively long activity segments, despite its lower boundary detection accuracy.

This suggests that the impact of boundary localization errors on downstream recognition is not solely determined by detection accuracy, but is strongly modulated by the intrinsic temporal granularity of the dataset. In short-duration activities, even small boundary shifts correspond to a large proportional change in temporal coverage, increasing the likelihood that a window spans multiple activity instances. This leads to higher label impurity and ambiguity in the aggregated representation.

Consequently, datasets with more fragmented activities, such as Milan, exhibit a larger gap between GT-based and detected-boundary performance. Conversely, in datasets with longer activity durations, such as Orange, even relatively lower boundary detection accuracy does not significantly degrade downstream recognition performance, as the model is less sensitive to temporal boundary perturbations.

Notably, even under the most challenging settings, our method consistently outperforms all baselines on Milan, demonstrating robustness to boundary noise and high-frequency activity transitions. Moreover, on the Orange dataset, despite relatively low boundary detection accuracy, our method still achieves competitive performance close to the GT upper bound, further highlighting its stability under imperfect boundary localization.

% Across all datasets and window sizes, the curves for detected and GT boundaries are almost overlapping, with only a small gap of about $2$--$3$ Macro-F1 points on average.
% In particular, the detector-based curves closely track the GT upper bound even on longer windows ($w{=}80,100$), where boundary localization is intrinsically harder.
% This suggests that our boundary detector recovers boundary positions that are already ``good enough'' for cross-window recognition; the remaining error stems primarily from the intrinsic difficulty of classifying heavily truncated or highly mixed activity segments, rather than from boundary mis-localization.

\paragraph{Qualitative boundary examples.}

\begin{figure}[htbp]
\centering
\setlength{\tabcolsep}{3pt} % 控制两列之间的水平间距
\renewcommand{\arraystretch}{1.0} % 控制行间距

\begin{tabular}{cc}
% ---------------- Row 1 ----------------
\subcaptionbox{Early boundary prediction ($\Delta=-1$)\label{fig:bd_orange_early}}{
  \includegraphics[width=0.48\textwidth]{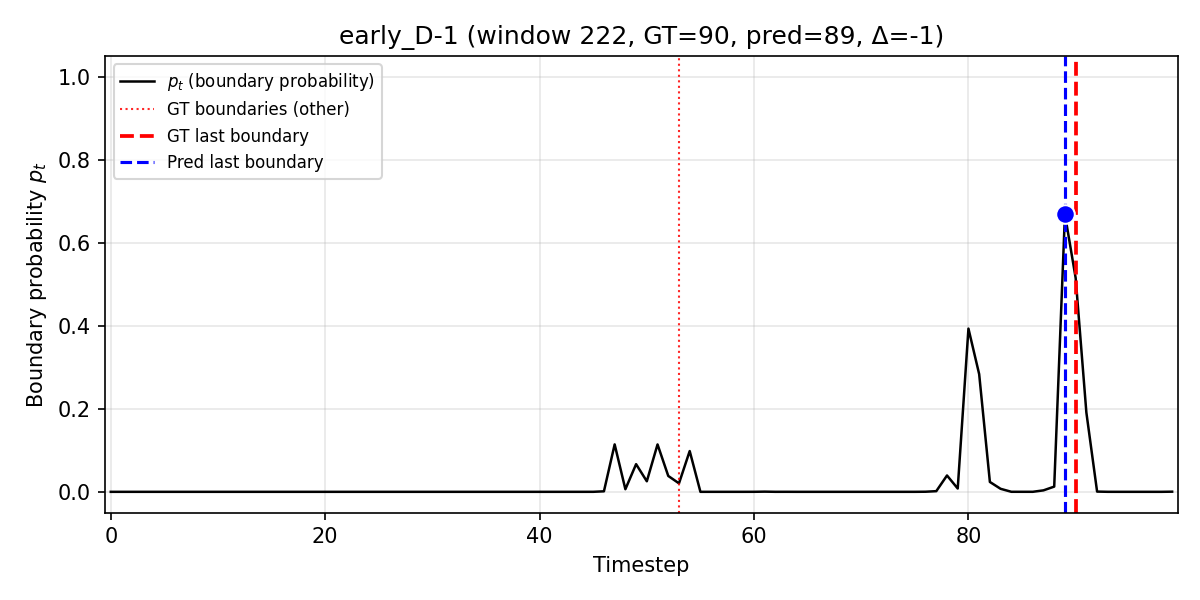}
}
&
\subcaptionbox{Hard boundary prediction ($\Delta=0$)\label{fig:bd_orange_hard}}{
  \includegraphics[width=0.48\textwidth]{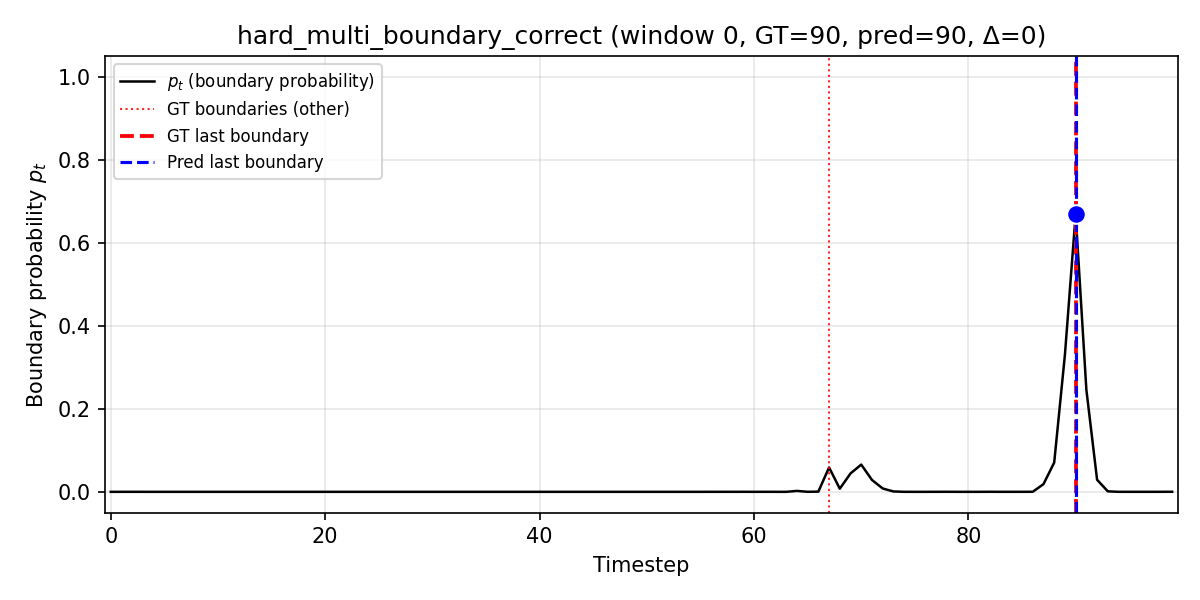}
}
\\[4pt]

% ---------------- Row 2 ----------------
\subcaptionbox{Late boundary prediction ($\Delta=+1$)\label{fig:bd_orange_late}}{
  \includegraphics[width=0.48\textwidth]{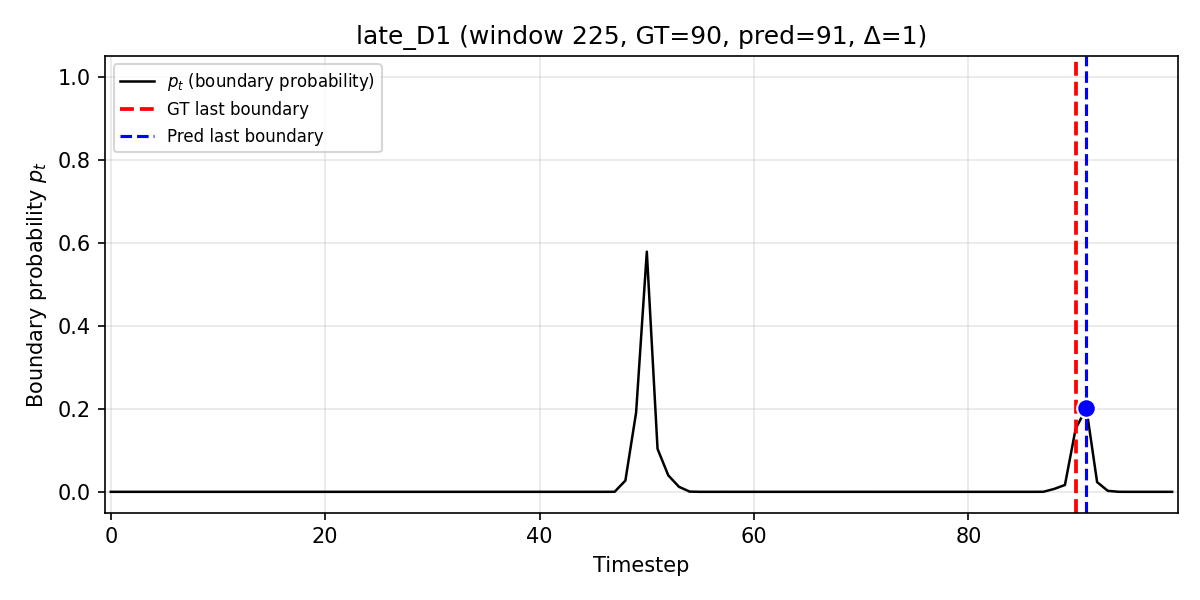}
}
&
\subcaptionbox{Perfect boundary detection ($\Delta=0$)\label{fig:bd_orange_perf}}{
  \includegraphics[width=0.48\textwidth]{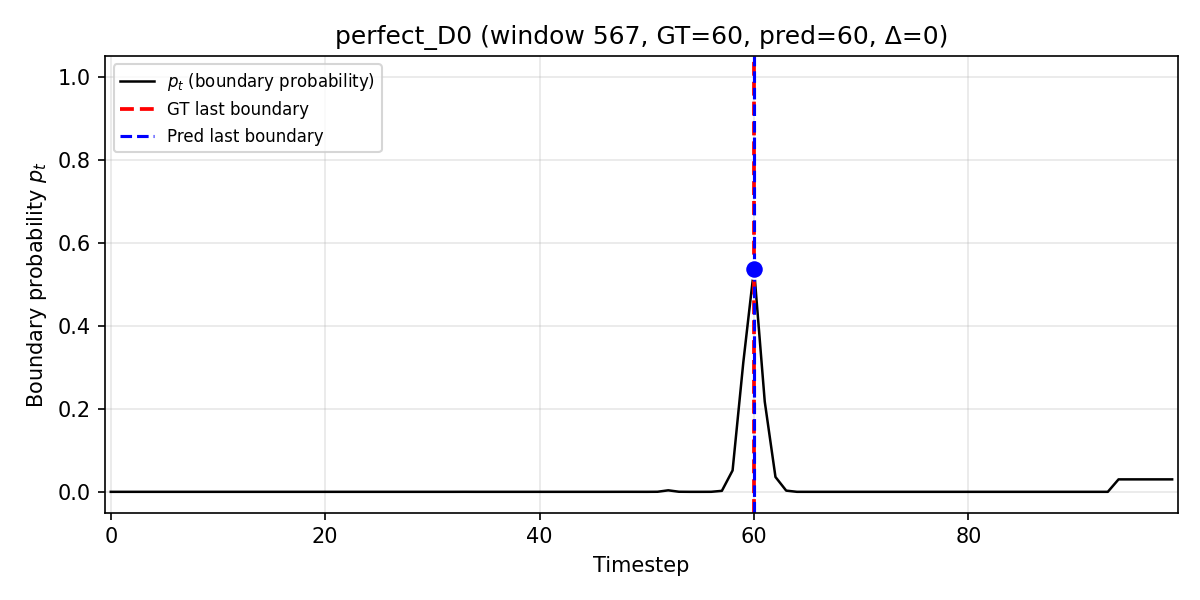}
}
\\
\end{tabular}

\caption{Qualitative boundary probability examples on the Orange dataset. 
The four panels illustrate early, hard, late, and perfect boundary cases.}
\label{fig:boundary_examples_orange_2x2}
\end{figure}

Figure~\ref{fig:boundary_examples_orange_2x2} provides qualitative boundary probability examples on the Orange dataset.
The four panels illustrate (a) early predictions ($\Delta=-1$), (b) hard but acceptable cases where the detector selects the correct last boundary among multiple candidates ($\Delta=0$), (c) slightly late predictions ($\Delta=+1$), and (d) perfect detections ($\Delta=0$) with a sharp, unimodal peak aligned to the ground-truth boundary.
These examples corroborate the quantitative findings: in most windows the detector produces a clear peak near the true last boundary, while early/late cases typically deviate by only one or two index positions.
% Figure~\ref{fig:boundary_examples_orange_2x2} visualizes the predicted boundary probability profiles on the Orange dataset for four representative windows: an early case ($\Delta$ = -1), a hard multi-boundary case with a correct last boundary ($\Delta$ = 0), a late case ($\Delta$ = +1), and a perfect detection ($\Delta$ = 0).
% In all examples, the detector produces a sharp peak in the vicinity of the ground-truth last boundary.
% Even in the early and late cases, the maximum is only one index away from the ground truth, consistent with the small MAE values reported in Table~\ref{tab:last_boundary_metrics_wide}.
% The hard example further illustrates that, when multiple candidate peaks appear within a window, the model still tends to assign the highest mass to the true last boundary.
% These qualitative patterns corroborate our quantitative metrics: boundary errors are typically narrow local shifts around the correct change point rather than large mis-localizations.

Together with Table~\ref{tab:last_boundary_metrics_wide} and Figure~\ref{fig:boundary_oracle_f1_per_dataset}, this shows that our boundary detector is accurate, unbiased, and provides near-GT boundary quality for cross-window ADL recognition.

\subsection{Generalization (Q6)}
% \subsubsection{Robustness to Window Length Variation}\label{sec: window size variation}
% \textcolor{orange}{In depth analysis of table~\ref{tab:window_results}}
% window size in (20,40,60,80,100)
\subsubsection{Generalization to Future Days}
Table~\ref{tab:future_results_f1} reports macro-F1 under the strict day-based split, broken down by purity ranges. Overall, the chronological split is substantially more challenging than the stratified protocol for all baselines:
when models are trained only on past days and evaluated on future days, their performance on mixed windows (10--30 and 40--60 purity) often collapses to single-digit or low double-digit F1. In contrast, our boundary-aware framework remains remarkably robust. On Milan and Aruba, for example, our model attains 61.11\% and 68.57\% F1 in the 10--30 range, compared to at most 8--9\% for the best baseline.
Similarly, on Kyoto and Orange at 10--30 purity, our method improves over the strongest baseline by more than 20 F1 points (47.20\% vs.\ 21.50\%, and 53.59\% vs.\ 25.15\%), despite the temporal non-stationarities introduced by the future-day split.
This pattern persists in the 40--60 range, where our approach consistently yields 50--70\% F1 across datasets, whereas baselines remain below 25--40\%.

\begin{table}[htbp]
\centering
\caption{Performance under day-based chronological splits (training on past days, testing on future days).}
\label{tab:future_results_f1}
\resizebox{0.85\textwidth}{!}{
\begin{tabular}{cc|cccccc}
\hline
\textbf{Dataset} & \textbf{Purity Range}
& DeepCASAS
& DCNN
& TCN
& CPC
& TDOST
& \textbf{Ours} \\
\hline

\multirow{4}{*}{\textbf{Milan}}
& 10--30  & 6.44 $\pm$ 0.00 & 5.03 $\pm$ 0.00 & 5.85 $\pm$ 0.00 & 4.88 $\pm$ 0.00 & \underline{8.41 $\pm$ 0.00} & \textbf{61.11 $\pm$ 1.73} \\
& 40--60  & 13.12 $\pm$ 0.00 & 16.37 $\pm$ 0.00 & \underline{20.16 $\pm$ 1.00} & 12.05 $\pm$ 1.00 & 14.72 $\pm$ 0.00 & \textbf{59.01 $\pm$ 1.41} \\
& 70--90  & 33.06 $\pm$ 3.46 & 36.55 $\pm$ 1.73 & \underline{41.84 $\pm$ 5.19} & 29.86 $\pm$ 0.01 & 15.55 $\pm$ 1.00 & \textbf{54.93 $\pm$ 0.00} \\
& 100     & 59.83 $\pm$ 2.82
           & 61.08 $\pm$ 4.00 
           & \underline{65.20 $\pm$ 4.24}
           & \textbf{68.85 $\pm$ 3.46}
           & 63.39 $\pm$ 0.00
           & 62.29 $\pm$ 0.66 \\
\hline

\multirow{4}{*}{\textbf{Aruba}}
& 10--30  & \underline{6.54 $\pm$ 1.41}  & 5.81 $\pm$ 1.00 & 6.09 $\pm$ 0.00 & 5.77 $\pm$ 0.00 & 5.47 $\pm$ 0.00 & \textbf{68.57 $\pm$ 3.16} \\
& 40--60  & 16.79 $\pm$ 4.24 & 13.87 $\pm$ 1.41 & \underline{21.22 $\pm$ 1.40} & 9.68 $\pm$ 1.41 & 9.52 $\pm$ 2.00 & \textbf{54.85 $\pm$ 4.12} \\
& 70--90  & 43.68 $\pm$ 6.07 & 49.02 $\pm$ 1.00 & \underline{50.26 $\pm$ 1.73} & 44.02 $\pm$ 1.73 & 17.21 $\pm$ 5.09 & \textbf{59.13 $\pm$ 5.29} \\
& 100     & 62.95 $\pm$ 4.58
           & 65.58 $\pm$ 0.00
           & \underline{77.98 $\pm$ 7.00}
           & \textbf{83.36 $\pm$ 5.09}
           & 71.91 $\pm$ 3.74
           & 70.25 $\pm$ 3.53 \\
\hline

\multirow{4}{*}{\textbf{Kyoto}}
& 10--30  & \underline{21.50 $\pm$ 6.55}  & 4.23 $\pm$ 1.00 & 5.33 $\pm$ 2.23 & 7.25 $\pm$ 1.41 & 10.56 $\pm$ 1.00 & \textbf{47.20 $\pm$ 3.74} \\
& 40--60  & \underline{41.88 $\pm$ 4.58} & 14.23 $\pm$ 2.00 & 24.69 $\pm$ 4.79 & 12.41 $\pm$ 3.16 & 27.47 $\pm$ 1.73 & \textbf{61.11 $\pm$ 1.73} \\
& 70--90  & \underline{49.17 $\pm$ 4.16} & 45.91 $\pm$ 2.64 & 46.90 $\pm$ 2.23 & 42.60 $\pm$ 1.00 & 44.02 $\pm$ 3.74 & \textbf{51.23 $\pm$ 1.73} \\
& 100     & 53.91 $\pm$ 1.00
           & 58.16 $\pm$ 0.00 
           & 60.31 $\pm$ 1.00
           & \underline{72.95 $\pm$ 1.73}
           & 60.12 $\pm$ 1.00
           & \textbf{66.93 $\pm$ 2.50} \\
\hline

\multirow{4}{*}{\textbf{Orange}}
& 10--30  & 15.80 $\pm$ 1.00 & 13.53 $\pm$ 4.47 & 14.77 $\pm$ 0.00 & 18.55 $\pm$ 2.00 & \underline{25.15 $\pm$ 1.00} & \textbf{53.59 $\pm$ 1.41} \\
& 40--60  & 31.89 $\pm$ 2.44 & 11.80 $\pm$ 3.31 & 20.71 $\pm$ 1.00 & 22.45 $\pm$ 4.69 & \underline{42.34$\pm$ 2.00} & \textbf{67.75 $\pm$ 2.44} \\
& 70--90  & 43.27 $\pm$ 5.38 & 31.61 $\pm$ 10.34 & 32.08 $\pm$ 1.73 & 41.66 $\pm$ 7.00 & \textbf{73.85 $\pm$ 1.73} &  \underline{69.49 $\pm$ 2.82} \\
& 100     & 54.50 $\pm$ 3.46
           & 54.55 $\pm$ 3.67
           & 37.97 $\pm$ 1.41
           & 57.93 $\pm$ 7.74
           & \textbf{88.44 $\pm$ 0.00}
           & \underline{72.80 $\pm$ 3.00} \\
\hline

\end{tabular}
}
\end{table}

As purity increases, all methods benefit from having cleaner windows, but our advantages remain most pronounced for mixed segments.
In the 70--90 range, our model continues to outperform prior work on Milan, Aruba, and Kyoto by 5--15 F1 points, and remains competitive with the strongest text-based baseline (TDOST) on Orange.
At 100\% purity, the task effectively reduces to classical single-activity recognition on fully segmented windows.
In this regime, strong sequence or text-based baselines such as CPC and TDOST achieve the highest scores on some datasets (e.g., 83.36\% on Aruba and 88.44\% on Orange).
Our model nonetheless remains competitive: it matches or exceeds most sensor-based baselines and only trails the best-performing method on a subset of datasets, while maintaining absolute F1 in the 62--73\% range.

Crucially, this modest gap appears only in the easiest regime, where all methods already achieve relatively high accuracy and the incremental value of further gains is limited.
In contrast, under low- and medium-purity windows (10--30 and 40--60) with future-day evaluation, our boundary-aware architecture improves macro-F1 by 20--60 points over the strongest baselines.
These results indicate that our approach achieves a better overall trade-off across window purities: it preserves strong performance on idealized pure windows, but, more importantly, dramatically improves robustness on realistic cross-activity windows, which constitute the dominant and most challenging regime in long-term deployments.

\subsubsection{Robustness to Spatial Uncertainty and Layout Mismatch.}

\begin{table}[t]
\centering
\caption{Test-time layout-mismatch results under direct perturbation of the resized \(32 \times 32\) trajectory canvas. The model is trained on the original unperturbed layout, while Gaussian noise is added only to test-time sensor locations on the resized trajectory canvas. Macro-F1 is reported as mean \(\pm\) standard deviation on pure and cross-activity windows.}
\label{tab:test_time_canvas_mismatch}
\resizebox{0.8\textwidth}{!}{%
\begin{tabular}{l|cc|cc|cc|cc}
\toprule
\multirow{2}{*}{$\sigma^2$} &
\multicolumn{2}{c|}{\textbf{Milan}} &
\multicolumn{2}{c|}{\textbf{Aruba}} &
\multicolumn{2}{c|}{\textbf{Kyoto}} &
\multicolumn{2}{c}{\textbf{Orange}} \\
\cmidrule(lr){2-3}\cmidrule(lr){4-5}\cmidrule(lr){6-7}\cmidrule(lr){8-9}
 & Pure F1 & Cross F1 & Pure F1 & Cross F1 & Pure F1 & Cross F1 & Pure F1 & Cross F1 \\
\midrule
None
& \(68.88 \pm 1.01\) & \(52.11 \pm 0.78\)
& \(72.64 \pm 0.30\) & \(55.75 \pm 1.36\)
& \(59.20 \pm 3.23\) & \(55.78 \pm 4.40\)
& \(63.83 \pm 0.64\) & \(61.19 \pm 3.22\) \\

0.1
& \(68.95 \pm 0.99\) & \(51.96 \pm 0.51\)
& \(72.71 \pm 0.18\) & \(55.59 \pm 1.15\)
& \(58.06 \pm 3.00\) & \(57.74 \pm 3.88\)
& \(62.92 \pm 3.14\) & \(59.64 \pm 4.48\) \\

1
& \(68.53 \pm 1.09\) & \(49.10 \pm 1.57\)
& \(73.55 \pm 1.56\) & \(52.18 \pm 2.17\)
& \(58.05 \pm 1.14\) & \(57.54 \pm 4.94\)
& \(53.64 \pm 2.98\) & \(54.66 \pm 5.91\) \\

3
& \(68.18 \pm 0.75\) & \(44.69 \pm 1.21\)
& \(72.42 \pm 1.07\) & \(49.45 \pm 2.74\)
& \(55.09 \pm 0.78\) & \(52.09 \pm 3.43\)
& \(51.59 \pm 2.03\) & \(46.91 \pm 5.13\) \\
\bottomrule
\end{tabular}%
}
\end{table}

% \begin{revblock}
LastAct requires floor-plan metadata to construct layout-aligned trajectory sequences, but it does not operate directly on raw physical coordinates. Since homes have different floor-plan sizes and image resolutions, all sensor locations are projected onto a fixed \(R \times R\) trajectory canvas. As a result, minor perturbations in the original floor-plan coordinate system may be absorbed by resizing and discretization. We therefore evaluate a stricter test-time layout-mismatch setting by perturbing sensor locations directly on the resized \(32 \times 32\) canvas, which changes the rendered trajectory geometry observed by the trained model.

Specifically, the model is trained using the original unperturbed layout, while i.i.d. zero-mean Gaussian noise is added only during testing to the rendered sensor coordinates on the \(32 \times 32\) canvas, with variance
\(
\sigma^2 \in \{0.05, 0.1, 0.5, 1, 1.5, 2, 2.5, 3\}.
\)
The model is not retrained or fine-tuned under the perturbed layouts. Table~\ref{tab:test_time_canvas_mismatch} reports the original-layout performance and the performance under different perturbation levels \((\sigma^2=3)\). 

The results show that LastAct is not hypersensitive to small perturbations of rendered sensor locations, but performance decreases when the spatial mismatch becomes large enough to distort the trajectory canvas. Under the strongest perturbation, cross-window Macro-F1 drops by \(7.42\), \(6.30\), \(3.69\), and \(14.28\) points on Milan, Aruba, Kyoto, and Orange, respectively. Pure-window performance remains nearly unchanged on Milan and Aruba, but decreases more noticeably on Kyoto and Orange, particularly on Orange, where the pure-window drop reaches \(12.24\) points.

This trend is consistent with the role of spatial information in LastAct. Small coordinate changes may preserve the same local sensor neighborhoods and room-level trajectory structure after projection, whereas larger canvas-level perturbations alter the spatial representation that the model learned during training. Therefore, the degradation under strong perturbation should not be interpreted as a failure of the trajectory representation; rather, it confirms that the model uses spatial layout information. In practical deployments, LastAct does not require perfectly calibrated pixel-level coordinates, but it does assume that the projected layout preserves stable functional spatial relationships. If sensors are moved to different rooms or functional areas, or if mobile/wearable devices are introduced without stable room-level anchoring, the layout cache should be updated and the model should be recalibrated or fine-tuned.

\subsection{Ablation Studies (Q7)}
\subsubsection{Ablation on trajectory representation capacity}
In this section, we ablate the spatial hyperparameters used to construct trajectory image sequences and study how they affect both recognition accuracy and training efficiency.
We vary the trajectory image resolution and the dot radius used to render sensor activations, and evaluate their impact on Test Macro-F1 and per-epoch training time.
\begin{figure}[t]
    \centering
    \includegraphics[width=0.55\linewidth]{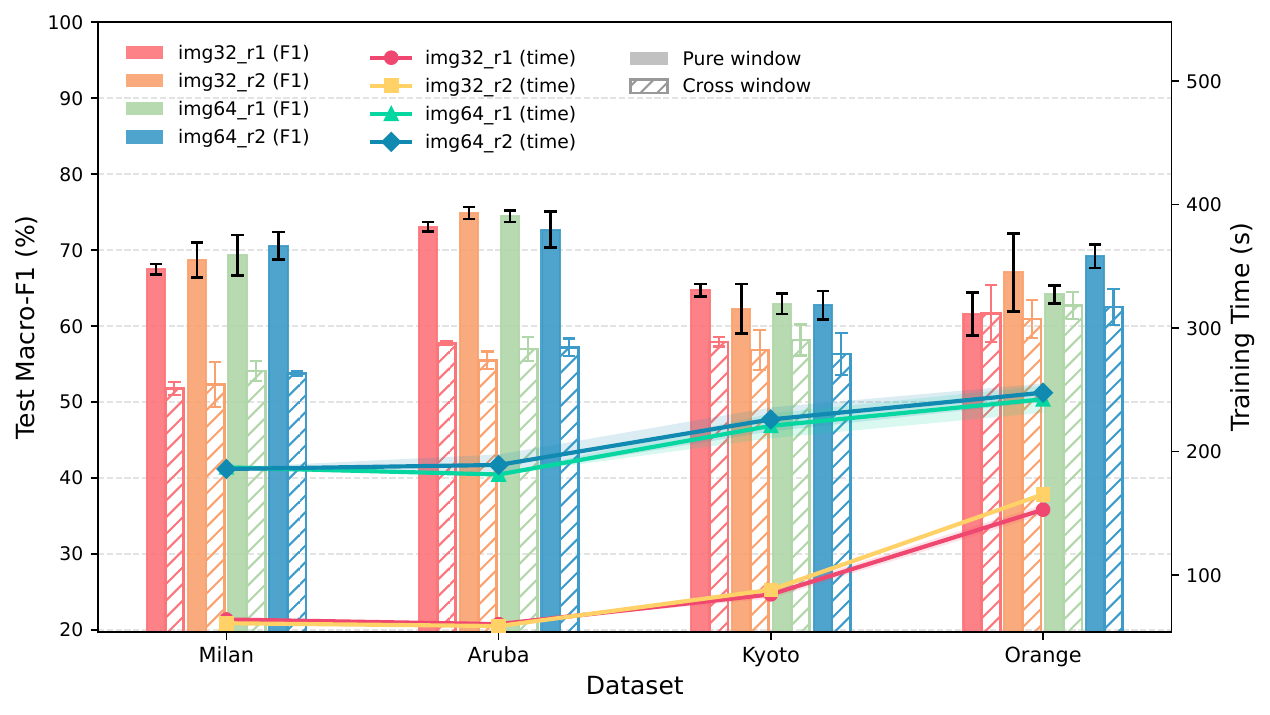}
    \caption{%
    Test Macro-F1 score and training time across four datasets under different spatial parameter settings for trajectory image construction.
    Bars (left y-axis) report Test Macro-F1 scores (\%) for four configurations: \texttt{img32\_r1}, \texttt{img32\_r2}, \texttt{img64\_r1}, and \texttt{img64\_r2}, where \texttt{img32} and \texttt{img64} denote trajectory image sizes of $32 \times 32$ and $64 \times 64$, respectively, and \texttt{r1} and \texttt{r2} indicate dot radii of 1 and 2 pixels. 
    Lines (right y-axis) show the corresponding training time in seconds, with shaded regions indicating $\pm$ one standard deviation.
    }
    \label{fig:performance_time_tradeoff}
\end{figure}
% Fig.~\ref{fig:performance_time_tradeoff} illustrates the trade-off between recognition performance and training efficiency across four datasets (Milan, Aruba, Kyoto, and Orange) under different image resolution and dot radius settings. Overall, changes in image resolution and dot radius lead to comparable but not strictly monotonic performance differences, indicating that neither parameter alone consistently dominates recognition accuracy across datasets.

Fig.~\ref{fig:performance_time_tradeoff} summarizes the trade-off between recognition performance and training efficiency across four datasets under different image resolutions and dot radii. Overall, varying these spatial parameters leads to only modest changes in Test Macro-F1, indicating that the trajectory-based recognizer is fairly robust to the exact rendering configuration.

Increasing resolution from $32 \times 32$ to $64 \times 64$ does not yield uniform accuracy gains: some datasets (e.g., Milan, Kyoto, Orange) benefit under specific settings, whereas Aruba shows almost no improvement and occasionally a slight drop. The effect of dot radius is similarly dataset- and resolution-dependent. With $32 \times 32$ images, moving from \texttt{r1} to \texttt{r2} often produces comparable or slightly worse F1, suggesting that thicker points can blur fine-grained trajectories on a small canvas. At $64 \times 64$, however, larger radii sometimes help, indicating that local rendering granularity and global image size must be tuned jointly rather than in isolation.

In contrast, training time is clearly more sensitive to these choices: both higher resolution and larger dot radius incur noticeable computational overhead. Taken together, the results reveal a mild performance–efficiency trade-off. Configurations such as \texttt{img32\_r2} offer a good balance for most datasets, while higher-cost settings (e.g., \texttt{img64\_r2}) may be reserved for scenarios where additional computation is acceptable.

\subsubsection{Ablation on Boundary Regularization Losses}
Table~\ref{tab:ablation_loss_components_datasetwise} disentangles the impact of the boundary-detector loss components. 
\textit{BCE} uses only token-wise labels, while \textit{Card} regularizes the number of predicted boundaries, \textit{Last} emphasizes the final boundary within each window, and \textit{Peak} sharpens boundary probabilities.

\begin{table}[htbp]
\centering
\caption{Ablation on boundary-detector loss components and downstream performance.
We report last-boundary localization (Acc/MAE) and downstream recognition (Cross-activity window Macro-F1).
Acc and F1 are reported in \%.}
\label{tab:ablation_loss_components_datasetwise}
\setlength{\tabcolsep}{2pt}
\renewcommand{\arraystretch}{1.05}
\resizebox{0.9\linewidth}{!}{%
\begin{tabular}{l|ccc|ccc|ccc|ccc}
\toprule
\multirow{2}{*}{\textbf{Config}} 
& \multicolumn{3}{c|}{\textbf{Milan}} 
& \multicolumn{3}{c|}{\textbf{Aruba}} 
& \multicolumn{3}{c|}{\textbf{Kyoto}} 
& \multicolumn{3}{c}{\textbf{Orange}} \\
\cmidrule(lr){2-4}\cmidrule(lr){5-7}\cmidrule(lr){8-10}\cmidrule(lr){11-13}
& Last Acc $\uparrow$ & Last MAE $\downarrow$ & Macro F1 $\uparrow$
& Last Acc $\uparrow$ & Last MAE $\downarrow$ & Macro F1 $\uparrow$
& Last Acc $\uparrow$ & Last MAE $\downarrow$ & Macro F1 $\uparrow$
& Last Acc $\uparrow$ & Last MAE $\downarrow$ & Macro F1 $\uparrow$ \\
\midrule
BCE      
& 86.07$\pm$0.43   & 4.22$\pm$0.42   & 50.82$\pm$0.78
& 96.54$\pm$0.27   & 1.19$\pm$0.10   & 54.26$\pm$1.25
& 77.34$\pm$3.55   & 5.54$\pm$0.77   & 53.96$\pm$2.77
& 64.18$\pm$5.14   & 8.26$\pm$0.81   & 56.98$\pm$3.51 \\
w/o Card 
& 88.04$\pm$0.90   & 3.78$\pm$0.41    & \underline{52.48$\pm$0.68}
& \textbf{97.36$\pm$0.40}           & \textbf{0.83$\pm$0.10}           & 55.54$\pm$1.61
& \underline{85.40$\pm$0.19}           & 3.54$\pm$0.28            & 55.16$\pm$2.79
& \textbf{77.76$\pm$1.57}              & 4.96$\pm$0.34      & \underline{60.83$\pm$3.17} \\
w/o Last 
& 87.16$\pm$0.36          & 3.81$\pm$0.25         & 51.99$\pm$1.11
& 97.14$\pm$0.18          & 0.98$\pm$0.05          & \underline{55.56$\pm$1.04}
& 81.33$\pm$1.65          & 4.44$\pm$0.26           & \textbf{56.34$\pm$2.54}
& 51.94$\pm$1.92          & 9.39$\pm$1.21          & 51.94$\pm$1.92 \\
w/o Peak 
& \textbf{88.59$\pm$0.08}      & \textbf{3.46$\pm$0.13}              & 51.91$\pm$0.62
& \underline{97.30$\pm$0.39}   & \underline{0.84$\pm$0.11}        & \textbf{56.86$\pm$1.71}
& \textbf{86.19$\pm$2.28}      & \underline{3.28$\pm$0.63}         & \underline{55.94$\pm$2.55}
& 60.21$\pm$2.49              & \underline{4.66$\pm$0.02}          & 60.21$\pm$2.49 \\
Full     
& \underline{88.38$\pm$0.21}     & \underline{3.48$\pm$0.16}    & \textbf{52.55$\pm$1.25}
& 97.16$\pm$0.30         & 0.86$\pm$0.09           & 55.35$\pm$1.68
& 85.28$\pm$1.41        & \textbf{3.03$\pm$0.03}         & 55.62$\pm$2.61
& \underline{77.04$\pm$2.36}       & \textbf{4.43$\pm$0.60}      & \textbf{61.23$\pm$2.65} \\
\bottomrule
\end{tabular}
}
\end{table}

% Table~\ref{tab:ablation_loss_components_datasetwise} disentangles the contribution of each loss component used to train the boundary detector.
% The BCE baseline optimizes only token-wise boundary labels, while the additional terms encourage globally consistent structure:
% cardinality regularization (\textit{Card}) aligns the number of predicted boundaries with the empirical distribution,
% the \textit{Last} term emphasizes the final boundary within each window, and the \textit{Peak} term penalizes diffused probability mass and promotes sharp peaks.
Adding structural terms on top of plain BCE consistently improves both last-boundary localization and downstream recognition. 
For instance, on Milan the full objective reduces MAE from 4.22 to 3.48 index positions and raises cross-window Macro-F1 from 50.82\% to 52.55\%; on Kyoto and Orange, MAE drops by roughly 40–45\% (5.54$\rightarrow$3.03, 8.26$\rightarrow$4.43) with gains of 1–3 F1 points over BCE. 
This confirms that pure token-wise supervision is insufficient for reliable boundary detection under mixed-activity conditions.

The individual components are complementary. 
Removing \textit{Card} typically hurts MAE and F1 on the more complex, multi-boundary datasets (Kyoto, Orange), suggesting that matching the boundary count distribution stabilizes training when several change points can occur per window. 
Dropping \textit{Last} severely degrades Orange (Last Acc 51.94\% vs.\ 77.04\%, Macro-F1 51.94\% vs.\ 61.23\%), indicating that explicitly supervising the final boundary is crucial when multiple candidates are present. 
The \textit{Peak} term trades slightly higher localization error for improved F1 on some datasets, whereas the full objective achieves a balanced profile: it is best or second-best in Macro-F1 on all datasets while keeping MAE among the lowest. 
Overall, shaping boundary structure via cardinality, last-boundary emphasis, and peak sharpness is key to obtaining the high-quality boundaries that underpin our cross-window ADL gains.

\section{Conclusion}
% We presented LastAct, a trajectory-centric framework for streaming smart-home HAR that targets boundary contamination in near-realistic window inference. Our method constructs layout-aligned trajectory image-sequence representations from ambient sensor streams, applies pure/cross gating with boundary localization to focus prediction on the most recent activity evidence, and uses a trajectory memory bank to reuse computations across overlapping windows for efficient deployment. Through comprehensive studies, we demonstrated the effectiveness of our method on real-word datasets.
We presented LastAct, a trajectory-centric framework for near-realistic sliding-window smart-home HAR that leverages layout-aligned trajectories and maintains robustness under boundary uncertainty. To support principled evaluation in this setting, we formulate near-realistic sliding-window HAR under mixed-activity windows and construct simulation datasets that reproduce cross-activity contamination in continuous smart-home streams.  Builiding on this formulation, LastAct converts ambient sensor events into layout-aligned trajectory image sequences, applies pure/cross gating with boundary localization to focus prediction on the most recent activity signal, and leverages a trajectory memory bank caches and reuses trajectory images to improve online inference efficiency. Extensive experiments on multiple real-world smart-home datasets show that our framework consistently improves over state-of-the-art baselines under both stratified and day-based mixed-activity protocols, while retaining competitive performance in the classical pure-window setting.
% Future work will study robustness to imperfect or missing layout metadata, reduce error propagation from the gate/boundary modules via uncertainty-aware routing, and evaluate end-to-end efficiency (latency, memory, energy) on edge or gateway hardware under long-term deployment.

% While our mixed-activity evaluation moves beyond conventional pure-window protocols, it still represents a controlled approximation of real deployments. We explicitly construct window datasets with fixed lengths and coarsely discretized purity ranges, which allows us to stress-test models under a much higher proportion of cross-activity windows than in prior work. However, pushing this idea to finer-grained purity bands or denser windowing would quickly lead to a combinatorial explosion in the number of windows, with cross-activity windows vastly outnumbering pure ones and making training and evaluation prohibitively expensive. 

% \begin{revblock}
% \textbf{Deployment limitations.}
LastAct assumes region-consistent spatial metadata for ambient sensors, such as room-level assignments or an approximate home layout, but does not require fine-grained coordinate measurements. This assumption may limit deployment in commercial smart-home systems where floorplans are unavailable, sensor metadata are incomplete or sensors are frequently moved. Approximate pseudo-floorplans can be constructed from installation metadata, user-provided room labels, or inferred sensor co-activation patterns, as in our treatment of Orange, but such approximations rely on stable sensor placement. 

The current evaluation is also within-home and does not establish cross-home
generalization. Performance may degrade when transferring to homes with different
layouts, sensor configurations, routines, inhabitant behaviors, or contextual conditions.
Because collecting annotated target-home data is expensive and intrusive, LastAct
currently assumes access to some labeled target-home data for supervised training or
adaptation. Future work should reduce this dependency through automatic layout
inference, topology-level spatial abstractions, cross-home pretraining, domain
adaptation, semi-supervised learning, and few-shot calibration.

% \textbf{Extension to parallel activities.}
% Although LastAct focuses on single-label latest-activity recognition, its layout-aligned trajectory representation is also relevant to parallel-activity settings, where concurrent activities may produce spatially separated sensor patterns on the floor plan. To preliminarily assess this possibility, we conducted an isolated-vs-parallel activity detection experiment on the multi-resident Kyoto7 dataset. Using a lightweight binary classifier built on the pretrained trajectory encoder, the model achieved $0.7182 \pm 0.0135$ Macro-F1 and $0.8512 \pm 0.0223$ recall for parallel episodes across three runs. These results suggest that trajectory encoding can capture useful signatures of concurrent activities. However, this experiment is only a proof of feasibility: it detects whether activity overlap is present, but does not yet solve full multi-label parallel-activity recognition or resident-specific activity attribution. We provide the dataset construction, protocol, and detailed metrics in Appendix~\ref{app:parallel_feasibility}.

% \textbf{Extension to parallel activities.}
LastAct focuses on single-label latest-activity recognition and does not address the
broader problem of parallel or overlapping activity recognition, where multiple
activities may be active simultaneously and, in multi-resident homes, may need to be
attributed to different residents. Nevertheless, the proposed layout-aligned trajectory
representation provides a natural basis for such settings, since concurrent activities
can produce spatially separated or partially overlapping traces on the floor plan. As a
preliminary feasibility analysis, Appendix~\ref{app:parallel_feasibility} reports a
clean-vs-parallel detection experiment on the multi-resident Kyoto7 dataset; full
multi-label, resident-aware parallel-activity recognition remains future work.

More broadly, our mixed-activity evaluation moves beyond conventional pure-window
protocols but remains a controlled approximation of real deployments. We adopt fixed
window lengths and coarsely discretized purity ranges to enable systematic stress
testing; pushing to finer-grained purity bands or denser windowing would generate a
much larger number of cross-activity windows and substantially increase computational
cost. We therefore view this work as a step toward, rather than a complete solution
for, realistic online ADL recognition. Our results suggest that trajectory-based
encoders, pure/cross gating, and boundary-aware inference can substantially improve
robustness under challenging mixed-activity conditions. Closing the remaining gap to
genuine online deployment---where labels may be delayed or incomplete and the
prevalence of cross-activity windows may differ substantially from our controlled
setting---will require adaptive windowing and gating policies, incremental boundary
detection, and tighter integration with online change-point detection.

% \end{revblock}

% We therefore view this work as a step towards, rather than a complete solution for, realistic online ADL recognition. Our results suggests that trajectory-based encoders, pure/cross gating, and boundary-aware inference can substantially improve robustness under challenging mixed-activity conditions. However, extending these ideas to genuine online activity recognition remains an open direction, since labels may be delayed or incomplete and the prevelance of cross-activity windows can be substantially higher than our simulations. Future work could investigate adaptive windowing and gating policies, incremental boundary detection, and tighter integration with online change-point detection to close the remaining gap between our simulated mixed-activity benchmarks and real-world online smart-home deployments.

\section{Acknowledgment}
Portions of the text in this manuscript were refined using OpenAI ChatGPT to improve clarity, phrasing, and grammar. The tool was not used to generate ideas, analyses, figures, tables, or any scientific content. The authors take full responsibility for the integrity and accuracy of all content in this manuscript.

\bibliography{reference}
\newpage
\appendix
\section{Appendix}
\subsection{Implementation Details}
\label{app:implementation}

Our proposed framework is implemented in PyTorch and trained in a structured two-stage procedure. In the first stage, we learn an activity-recognition backbone using pure windows, where each window contains a single ground-truth activity. In the second stage, the backbone parameters are frozen, and we train the boundary-aware modules responsible for pure/cross window gating and boundary localization.

For the activity recognition backbone, each window is converted into a $32\times 32$ layout-aligned trajectory image sequence with dot radius 2. We further augment the input by encoding cyclic time metadata, specifically hour-of-day and day-of-week, as learnable embeddings. The backbone architecture integrates spatial and temporal modeling: first, a 3-layer Convolutional Neural Network (CNN) processes the trajectory images with channel widths of $\{16,32,64\}$. Each layer utilizes ReLU activations and pooling, with a final adaptive average pooling layer producing a fixed-length spatial feature of dimension 1024 per time step. Simultaneously, a cyclic temporal encoder embeds the time metadata using lookup tables and projects the concatenated result to a feature with dimension 128.
These per-step features are then forwarded to a 2-layer unidirectional Long Short-Term Memory (LSTM) network with 256 hidden units and a dropout rate of 0.5. An attention pooling layer is applied to the LSTM outputs to derive a window-level representation, which is finally mapped to activity classes via a classification head consisting of $\mathrm{Linear}(256,512)\rightarrow \mathrm{ReLU}\rightarrow \mathrm{Dropout}(0.5)\rightarrow \mathrm{Linear}(512,|\mathcal{Y}|)$,
 where $|\mathcal{Y}|$ is the number of activities. We optimize cross-entropy with AdamW (learning rate $10^{-3}$, weight decay $10^{-4}$) using batch size 64.
We train for up to 100 epochs with early stopping (patience 20) based on validation performance.

The boundary-aware modules are trained on window-level supervision with all backbone parameters frozen. The Gating Module distinguishes between pure and cross windows using a 2049-dimensional summary vector derived from the spatial features. This vector is formed by concatenating (i) a 1024-d mean pooling, (ii) a 1024-d standard-deviation pooling, and (iii) a 1-d scalar temporal-change statistic representing the Top-1 difference. This summary is processed by a Multi-Layer Perceptron (MLP) with Layer Normalization and GELU activations: $LN\rightarrow Linear(2049,256)\rightarrow GELU \rightarrow Dropout(0.1) \rightarrow Linear(256,64) \rightarrow GELU\rightarrow Dropout(0.1) \rightarrow Linear(64,2)$. For cross windows, the Boundary Detection Module predicts a per-step boundary logit sequence. Spatial features are projected to dimension of 384 and passed through a 4-layer Transformer encoder with 4 attention heads and sinusoidal positional encodings. We train the boundary detector using the compound objective defined in Sec.~\ref{sec:boundary_detection}.
Unless otherwise noted, we use $\lambda_{\mathrm{card}}=0.1$, $\lambda_{\mathrm{late}}=0.1$, $\lambda_{\mathrm{last}}=0.2$, and $\lambda_{\mathrm{peak}}=0.03$ with $\sigma=1.0$ for the Gaussian target in $L_{\mathrm{last}}$.
We set the post-boundary penalty region in $L_{\mathrm{late}}$ to 4 steps. We train up to 100 epochs with batch size 64 and early stopping (patience 20), using AdamW with $lr=10^{-4}$ for the Transformer and $2\times 10^{-4}$ for the projection layer.

We compare our framework against several representative baselines. For change-point detection using CPC, we adhere to the hyperparameters and protocol established in TDOST. Other baselines are evaluated using their authors' default settings, ensuring consistency across data splits and evaluation protocols.

\subsection{Additional Results for the Controlled Mixed-Window Protocol}
Table~\ref{tab:full_counts} summarizes the total number of Pure and Cross-activity windows generated for each dataset under controlled window setting. We observe that while the number of episode-aligned (pure) windows remains constant regardless of $w$, the count of cross-activity windows typically declines as window length increases. As $w$ increases, many short-duration activities no longer meet this
minimum length and therefore cannot form valid cross-activity windows.

Across all window sizes and datasets, the purity generator maintained high consistency: the Low-purity bucket averaged a mean of $0.19$, Mid-purity averaged $0.49$, and High-purity averaged $0.79$, with a standard deviation of $\approx 0.08$ throughout.

\begin{table}[htbp]
\centering
\caption{Full window count distribution across all evaluated window sizes ($w$) and purity strata.}
\label{tab:full_counts}
\small
\resizebox{0.48\textwidth}{!}{%
\begin{tabular}{l l rrrrr}
\toprule
\textbf{Dataset} & \textbf{Purity} 
  & \multicolumn{5}{c}{\textbf{Window Size $w$}} \\
\cmidrule(lr){3-7}
& & 20 & 40 & 60 & 80 & 100 \\
\midrule
\multirow{4}{*}{Aruba} 
  & Pure (100\%)      & 3,968 & 3,968 & 3,968 & 3,968 & 3,968 \\
  & Low (10--30\%)    & 10,668 & 9,917 & 8,620 & 8,092 & 7,690 \\
  & Mid (40--60\%)    & 9,088 & 7,969 & 7,665 & 7,446 & 7,177 \\
  & High (70--90\%)   & 8,763 & 8,376 & 8,004 & 7,428 & 6,801 \\
\midrule
\multirow{4}{*}{Milan} 
  & Pure (100\%)      & 2,628 & 2,628 & 2,628 & 2,628 & 2,628 \\
  & Low (10--30\%)    & 6,684 & 6,514 & 6,326 & 6,047 & 5,625 \\
  & Mid (40--60\%)    & 6,399 & 5,140 & 4,382 & 3,819 & 3,437 \\
  & High (70--90\%)   & 5,571 & 4,323 & 3,635 & 3,179 & 2,814 \\
\midrule
\multirow{4}{*}{Kyoto} 
  & Pure (100\%)      &   591 &   591 &   591 &   591 &   591 \\
  & Low (10--30\%)    & 1,291 & 1,256 & 1,262 & 1,248 & 1,233 \\
  & Mid (40--60\%)    & 1,249 & 1,221 & 1,181 & 1,151 & 1,124 \\
  & High (70--90\%)   & 1,233 & 1,192 & 1,136 &   995 &   920 \\
\midrule
\multirow{4}{*}{Orange} 
  & Pure (100\%)      &   493 &   493 &   493 &   493 &   493 \\
  & Low (10--30\%)    & 1,218 & 1,218 & 1,212 & 1,212 & 1,212 \\
  & Mid (40--60\%)    & 1,218 & 1,218 & 1,218 & 1,215 & 1,212 \\
  & High (70--90\%)   & 1,218 & 1,218 & 1,218 & 1,218 & 1,218 \\
\bottomrule
\end{tabular}%
}
\end{table}

\begin{table}[htbp]
\centering
\caption{Number of windows per dataset and split under the day-based protocol under controlled window setting. 
``Pure'' denotes single-activity windows; ``10--30'', ``40--60'', and ``70--90'' denote cross windows grouped by purity ranges (in \%).}
\label{tab:day_based_counts}
\resizebox{0.40\textwidth}{!}{
\begin{tabular}{llrrrr}
\toprule
\textbf{Dataset} & \textbf{Split} & \textbf{Pure} & \textbf{10--30} & \textbf{40--60} & \textbf{70--90} \\
\midrule
Aruba  & Train & 2432 & 5313 & 4665 & 4029 \\
       & Val   &  755 & 1774 & 1641 & 1481 \\
       & Test  &  820 & 1751 & 1596 & 1420 \\
\midrule
Kyoto  & Train &  343 &  880 &  732 &  592 \\
       & Val   &  132 &  319 &  261 &  216 \\
       & Test  &  145 &  373 &  292 &  239 \\
\midrule
Milan  & Train & 1330 & 3037 & 2101 & 1659 \\
       & Val   &  623 & 1392 &  931 &  740 \\
       & Test  &  694 & 1591 & 1122 &  870 \\
\midrule
Orange & Train &  218 &  651 &  651 &  651 \\
       & Val   &   70 &  207 &  207 &  207 \\
       & Test  &   61 &  180 &  180 &  180 \\
\bottomrule
\end{tabular}}
\end{table}
Table~\ref{tab:day_based_counts} summarizes the number of windows per dataset and purity range. Across all datasets, cross windows (10--30\%, 40--60\%, 70--90\% purity) substantially outnumber pure windows; for example, on Aruba and Milan the training set contains roughly $2$--$3\times$ more cross windows than pure ones.

The label distributions further highlight how day-based splitting amplifies dataset-specific characteristics. On \textbf{Aruba}, both pure and cross windows are dominated by two high-frequency activities (\texttt{Relax} and \texttt{Meal\_Preparation}), while rare events such as \texttt{Respirate} or \texttt{Bed\_to\_Toilet} occupy less than $1\%$ of windows in some purity bands. \textbf{Milan} and \textbf{Kyoto} share the same basic sensor modalities (motion, door, temperature) but with different spatial scales: Milan has a rich set of room-level activities (e.g., \texttt{Kitchen\_Activity}, \texttt{Guest\_Bathroom}, \texttt{Master\_Bathroom}, \texttt{Read}), and these categories remain dominant as we move from pure to cross windows, whereas several fine-grained activities (e.g., \texttt{Morning\_Meds}, \texttt{Meditate}) stay in the extreme long tail. Kyoto also exhibits many activities, but the absolute number of windows per class is much smaller than in Aruba or Milan, making each minority class particularly hard to model under day-based splits.
The \textbf{Orange} dataset illustrates a different regime. It contains a relatively small number of days and is heavily imbalanced: a few high-level categories such as \texttt{Other} and \texttt{Work} account for about half of all windows, while activities like \texttt{Bathing} or \texttt{Sleep} are rare even in pure windows. 
The \textbf{Orange} dataset illustrates a different regime. \texttt{Work}, \texttt{Cook}, \texttt{Personal\_Hygiene}, and \texttt{Relax} together account for the vast majority of windows, while \texttt{Bathing} and \texttt{Sleep} are rare.

\begin{figure*}[htbp]
    \centering
    \includegraphics[width=0.70\textwidth]{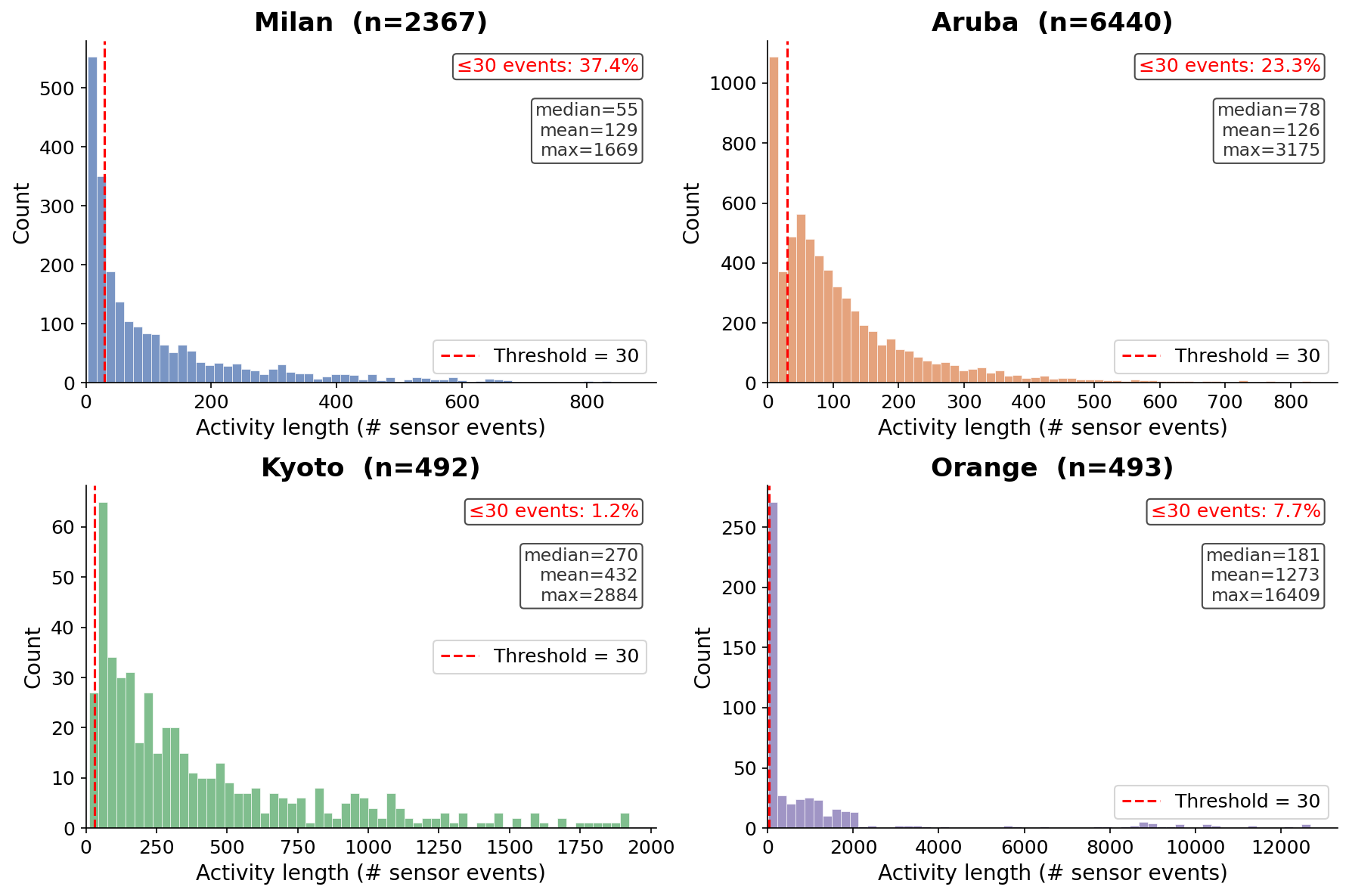}
    \caption{\textbf{Activity length distribution across four smart home datasets}. Each bar represents the number of sensor events fired during a single activity episode (begin–end pair). The red dashed line    
  marks the short-activity threshold of 30 sensor events. Milan and Aruba exhibit the most short episodes    
  (37.4\% and 23.3\% below the threshold, respectively), while Kyoto shows a broader, flatter distribution with a higher median (270 events). Orange has a heavy right tail driven by a small number of very
   long episodes (max = 16,409 events).}
    \label{fig:activity_length_distribution}
\end{figure*}

% \begin{revblock}
\subsection{Additional Raw-Stream Evaluation Details}
\label{app:raw_stream}

\subsubsection{Activity Episode Length Distribution}

Figure~\ref{fig:activity_length_distribution} reports the distribution of annotated activity episode lengths, measured by the number of sensor events per episode. The red dashed line marks 30 events as a descriptive short-episode reference point. The threshold is not used as a training or evaluation threshold.

The distributions show two properties relevant to the raw-stream protocol. First, Milan and Aruba contain many short episodes: 37.4\% and 23.3\% of episodes contain at most 30 events, respectively. Coarser strides can therefore introduce phase effects, where short or newly started activities may be skipped or under-sampled. Second, all datasets have heavy-tailed episode lengths, with maximum lengths of 1,669, 3,175, 2,884, and 16,409 events on Milan, Aruba, Kyoto, and Orange, respectively. Thus, dense raw-stream evaluation contains many windows located deep inside long-running activities rather than only near activity onsets. This motivates using stride~1 as the primary online evaluation setting and training the raw-stream activity backbone with sliding pure windows rather than only the first \(w\) events of each episode.

\subsubsection{Raw-Stream Test-Window Composition}
\label{app:raw_stream_purity_distribution}

Table~\ref{tab:raw_stream_stride_distribution} reports the empirical composition of raw-stream test windows under evaluation strides of 1, 50, and 100. These windows are extracted by applying fixed-stride sliding-window evaluation directly to the chronological test stream, without using activity-boundary annotations to determine window locations. The purity bins are computed only after window extraction by measuring the fraction of events belonging to the latest activity in each window; they are not used during window selection and are not provided to the model. This analysis therefore characterizes the natural purity distribution induced by raw-stream evaluation and shows how coarser strides reduce the number of evaluated windows while largely preserving the overall pure/cross composition and cross-window purity profile.
% Table~\ref{tab:raw_stream_purity_distribution_app} reports the empirical composition of stride-1 raw-stream test windows. These bins are computed only after window extraction; they are not used to choose window locations and are not provided to the model. Therefore, this analysis characterizes the natural purity distribution induced by dense sliding-window evaluation.

% \begin{table}[t]
% \centering
% \caption{\textcolor{blue}{Empirical distribution of stride-1 raw-stream test windows across purity bins under the day-based chronological split. \textit{Pure} denotes windows fully contained in a single annotated activity episode. Cross-window bins are computed post hoc by measuring the fraction of events belonging to the latest activity in the window.}}
% \label{tab:raw_stream_purity_distribution_app}
% \resizebox{0.95\textwidth}{!}{%
% \begin{tabular}{lrrrrrrr}
% \toprule
% Dataset & Pure & 0--20 & 21--40 & 41--60 & 61--80 & 81--99 & Total \\
% \midrule
% Aruba  & 64,711 (43.9\%)  & 15,403 (10.4\%) & 15,380 (10.4\%) & 17,445 (11.8\%) & 20,621 (14.0\%) & 13,892 (9.4\%) & 147,452 \\
% Milan  & 37,284 (44.4\%)  & 11,737 (14.0\%) & 9,113 (10.8\%)  & 8,751 (10.4\%)  & 8,807 (10.5\%)  & 8,352 (9.9\%)  & 84,044  \\
% Kyoto  & 18,644 (66.1\%)  & 2,029 (7.2\%)   & 1,941 (6.9\%)   & 2,235 (7.9\%)   & 1,807 (6.4\%)   & 1,559 (5.5\%)  & 28,215  \\
% Orange & 121,465 (93.8\%) & 1,694 (1.3\%)   & 1,493 (1.2\%)   & 1,331 (1.0\%)   & 1,116 (0.9\%)   & 2,393 (1.8\%)  & 129,492 \\
% \bottomrule
% \end{tabular}%
% }
% \end{table}

\begin{table*}[t]
\centering
\scriptsize
\setlength{\tabcolsep}{3pt}
\caption{Empirical distribution of raw-stream test windows under the day-based chronological split using evaluation strides of 1, 50, and 100. \textit{Pure} denotes windows fully contained in a single annotated activity episode. \textit{Cross} denotes windows spanning more than one annotated activity episode. Cross-window purity bins are computed post hoc by measuring the fraction of events belonging to the latest activity in each window.}
\label{tab:raw_stream_stride_distribution}
\resizebox{\textwidth}{!}{
\begin{tabular}{llrrrrrrrr}
\toprule
\multirow{2}{*}{Dataset} &
\multirow{2}{*}{Eval. stride} &
\multirow{2}{*}{Total} &
\multirow{2}{*}{Pure} &
\multirow{2}{*}{Cross} &
\multicolumn{5}{c}{Cross-window purity distribution} \\
\cmidrule(lr){6-10}
& & & & & 0--20\% & 21--40\% & 41--60\% & 61--80\% & 81--99\% \\
\midrule

\multirow{3}{*}{Milan}
& 1   & 84,044  & 37,284 (44.4\%)  & 46,760 (55.6\%) & 14,577 (31.2\%) & 10,541 (22.5\%) & 8,622 (18.4\%) & 7,248 (15.5\%) & 5,772 (12.3\%) \\
& 50  & 1,673   & 737 (44.1\%)     & 936 (55.9\%)    & 289 (30.9\%)    & 212 (22.6\%)    & 165 (17.6\%)   & 140 (15.0\%)   & 130 (13.9\%) \\
& 100 & 841     & 364 (43.3\%)     & 477 (56.7\%)    & 145 (30.4\%)    & 106 (22.2\%)    & 83 (17.4\%)    & 70 (14.7\%)    & 73 (15.3\%) \\
\midrule

\multirow{3}{*}{Aruba}
& 1   & 147,452 & 64,711 (43.9\%)  & 82,741 (56.1\%) & 21,576 (26.1\%) & 19,432 (23.5\%) & 16,933 (20.5\%) & 13,959 (16.9\%) & 10,841 (13.1\%) \\
& 50  & 2,970   & 1,297 (43.7\%)   & 1,673 (56.3\%)  & 437 (26.1\%)    & 405 (24.2\%)    & 316 (18.9\%)    & 301 (18.0\%)    & 214 (12.8\%) \\
& 100 & 1,480   & 641 (43.3\%)     & 839 (56.7\%)    & 223 (26.6\%)    & 186 (22.2\%)    & 162 (19.3\%)    & 151 (18.0\%)    & 117 (13.9\%) \\
\midrule

\multirow{3}{*}{Kyoto}
& 1   & 28,215  & 18,644 (66.1\%)  & 9,571 (33.9\%)  & 2,419 (25.3\%)  & 2,193 (22.9\%)  & 1,868 (19.5\%)  & 1,684 (17.6\%)  & 1,407 (14.7\%) \\
& 50  & 565     & 377 (66.7\%)     & 188 (33.3\%)    & 45 (23.9\%)     & 45 (23.9\%)     & 41 (21.8\%)     & 32 (17.0\%)     & 25 (13.3\%) \\
& 100 & 283     & 188 (66.4\%)     & 95 (33.6\%)     & 20 (21.1\%)     & 25 (26.3\%)     & 22 (23.2\%)     & 16 (16.8\%)     & 12 (12.6\%) \\
\midrule

\multirow{3}{*}{Orange}
& 1   & 129,492 & 121,465 (93.8\%) & 8,027 (6.2\%)   & 1,694 (21.1\%)  & 1,493 (18.6\%)  & 1,331 (16.6\%)  & 1,116 (13.9\%)  & 2,393 (29.8\%) \\
& 50  & 2,488   & 2,345 (94.3\%)   & 143 (5.7\%)     & 30 (21.0\%)     & 37 (25.9\%)     & 18 (12.6\%)     & 30 (21.0\%)     & 28 (19.6\%) \\
& 100 & 1,242   & 1,174 (94.5\%)   & 68 (5.5\%)      & 16 (23.5\%)     & 21 (30.9\%)     & 5 (7.4\%)       & 13 (19.1\%)     & 13 (19.1\%) \\
\bottomrule
\end{tabular}
}
\end{table*}

\subsubsection{Sensitivity to Backbone Training Protocol}
\label{app:raw_stream_backbone_training}

The raw-stream protocol in Section~\ref{sec:raw_stream_eval} uses a different activity-backbone training distribution from the controlled mixed-window protocol. In the controlled protocol, the activity backbone is trained using the first $w=100$ events of each annotated episode. In contrast, for raw-stream evaluation, we train the backbone using length-100 pure sliding windows sampled with step size~50 from the training days. This adjustment is intended to better match the raw-stream test distribution, where many windows occur inside long-running activities rather than only near activity onsets.

To verify that the raw-stream conclusions are not an artifact of this backbone-training choice, we compare two training protocols on the same stride-1 raw-stream test sets: (i) \textit{first-100}, where each annotated episode contributes one pure training window consisting of its first $w=100$ events; and (ii) \textit{sliding-50}, where pure training windows are extracted from each training episode using step size~50. In both cases, raw-stream test windows are extracted directly from the chronological event stream; ground-truth activity boundaries are not used to determine test-window positions.

\begin{table}[t]
\centering
\caption{Macro-F1 (\%, mean $\pm$ std) on the same stride-1 raw-stream sliding-window test sets under two backbone-training protocols. ``Stride-50 pure'' trains the backbone using length-100 pure windows sampled with stride 50 within activity episodes. ``First-100 pure'' trains the backbone using only the first 100 events of each activity episode. The $\Delta$ row reports the absolute Macro-F1 change: (Stride-50 pure) $-$ (First-100 pure). Positive changes are shown in green and negative changes in red.}
\label{tab:raw_stream_backbone_training}
\resizebox{0.8\textwidth}{!}{
\begin{tabular}{llrrrrrr}
\toprule
Dataset & Backbone Training & DeepCASAS & DCNN & TCN & CPC & TDOST & Ours \\
\midrule
\multirow{3}{*}{Milan}
& Stride-50 pure & 36.7$\pm$0.3 & 35.6$\pm$1.0 & 34.4$\pm$0.9 & 32.7$\pm$0.7 & 34.9$\pm$0.5 & \textbf{40.1$\pm$0.5} \\
& First-100 pure & 30.3$\pm$1.8 & 31.4$\pm$0.6 & 33.6$\pm$0.5 & 28.6$\pm$0.4 & 26.4$\pm$0.8 & \textbf{35.7$\pm$0.5} \\
& $\Delta$ & \textcolor{green!50!black}{+6.4} & \textcolor{green!50!black}{+4.2} & \textcolor{green!50!black}{+0.8} & \textcolor{green!50!black}{+4.1} & \textcolor{green!50!black}{+8.5} & \textcolor{green!50!black}{+4.4} \\
\midrule
\multirow{3}{*}{Aruba}
& Stride-50 pure & 38.6$\pm$0.5 & 36.4$\pm$0.7 & 37.0$\pm$0.9 & 38.4$\pm$0.5 & 36.5$\pm$0.1 & \textbf{44.6$\pm$1.9} \\
& First-100 pure & 36.1$\pm$1.9 & 33.2$\pm$0.8 & \textbf{38.8$\pm$0.8} & 33.2$\pm$0.7 & 28.9$\pm$2.1 & 38.3$\pm$1.0 \\
& $\Delta$ & \textcolor{green!50!black}{+2.5} & \textcolor{green!50!black}{+3.2} & \textcolor{red!70!black}{-1.8} & \textcolor{green!50!black}{+5.2} & \textcolor{green!50!black}{+7.6} & \textcolor{green!50!black}{+6.3} \\
\midrule
\multirow{3}{*}{Kyoto}
& Stride-50 pure & 41.0$\pm$1.8 & 34.2$\pm$0.6 & 36.6$\pm$2.0 & 35.5$\pm$1.5 & 42.5$\pm$1.2 & \textbf{48.6$\pm$8.5} \\
& First-100 pure & 42.4$\pm$2.4 & 36.6$\pm$0.7 & 38.4$\pm$0.6 & 34.2$\pm$0.8 & 40.6$\pm$0.8 & \textbf{46.9$\pm$0.5} \\
& $\Delta$ & \textcolor{red!70!black}{-1.4} & \textcolor{red!70!black}{-2.4} & \textcolor{red!70!black}{-1.8} & \textcolor{green!50!black}{+1.3} & \textcolor{green!50!black}{+1.9} & \textcolor{green!50!black}{+1.7} \\
\midrule
\multirow{3}{*}{Orange}
& Stride-50 pure & 40.9$\pm$0.9 & 44.7$\pm$1.9 & 26.3$\pm$1.4 & 53.1$\pm$2.5 & 63.1$\pm$0.6 & \textbf{72.9$\pm$7.8} \\
& First-100 pure & 19.1$\pm$2.3 & 9.6$\pm$6.0 & 11.7$\pm$1.3 & 18.1$\pm$1.5 & \textbf{33.7$\pm$1.3} & 25.1$\pm$1.3 \\
& $\Delta$ & \textcolor{green!50!black}{+21.8} & \textcolor{green!50!black}{+35.1} & \textcolor{green!50!black}{+14.6} & \textcolor{green!50!black}{+35.0} & \textcolor{green!50!black}{+29.4} & \textcolor{green!50!black}{+47.8} \\
\bottomrule
\end{tabular}
}
\end{table}

Table~\ref{tab:raw_stream_backbone_training} shows that sliding-window backbone training generally improves distribution matching under raw-stream evaluation, especially on Milan, Aruba, and Orange. The effect is largest on Orange, where the first-100 protocol severely under-represents the long within-activity windows that dominate raw-stream inference. Kyoto is less sensitive to the change, but LastAct remains the best-performing method under both training protocols. These results justify using the stride-50 pure-window backbone for the raw-stream experiments while showing that the conclusion is not solely driven by the modified training distribution.

\subsubsection{Sensitivity to Raw-Stream Evaluation Stride}
\label{app:raw_stream_stride}

We further evaluate whether the raw-stream conclusions depend on the choice of evaluation stride. Using the stride-50 pure-window backbone training protocol described above, we extract raw-stream test windows with strides $s\in\{1,50,100\}$, corresponding to dense online inference, 50\% overlap, and non-overlapping windows, respectively. In all cases, test windows are sampled directly from the chronological event stream rather than anchored to ground-truth episode boundaries.

\begin{table}[t]
\centering
\caption{Raw-stream sliding-window evaluation with window size $w=100$ and test strides $s\in\{1,50,100\}$. Windows are extracted directly from the chronological event stream without using ground-truth activity boundaries to determine test-window locations. Results are Macro-F1 (\%, mean $\pm$ std). Best result in each row is bolded. Classes with zero test support under a given stride are excluded from Macro-F1 computation.}
\label{tab:raw_stream_stride_macro}
\resizebox{0.7\linewidth}{!}{
\begin{tabular}{llcccccc}
\toprule
Dataset & Stride & DeepCASAS & DCNN & TCN & CPC & TDOST & Ours \\
\midrule
Milan  & 1   & 36.7$\pm$0.3 & 35.6$\pm$1.0 & 34.4$\pm$0.9 & 32.7$\pm$0.7 & 34.9$\pm$0.5 & \textbf{40.1$\pm$0.5} \\
Aruba  & 1   & 38.6$\pm$0.5 & 36.4$\pm$0.7 & 37.0$\pm$0.9 & 38.4$\pm$0.5 & 36.5$\pm$0.1 & \textbf{44.6$\pm$1.9} \\
Kyoto  & 1   & 41.0$\pm$1.8 & 34.2$\pm$0.6 & 36.6$\pm$2.0 & 35.5$\pm$1.5 & 42.5$\pm$1.2 & \textbf{48.6$\pm$8.5} \\
Orange & 1   & 40.9$\pm$0.9 & 44.7$\pm$1.9 & 26.3$\pm$1.4 & 53.1$\pm$2.5 & 63.1$\pm$0.6 & \textbf{72.9$\pm$7.8} \\
\midrule
Milan  & 50  & 37.0$\pm$1.4 & 36.4$\pm$1.1 & 36.0$\pm$1.3 & 33.9$\pm$0.9 & 36.0$\pm$0.1 & \textbf{40.3$\pm$1.8} \\
Aruba  & 50  & \textbf{38.8$\pm$1.9} & 36.3$\pm$0.6 & 36.8$\pm$0.9 & 36.7$\pm$2.9 & 36.2$\pm$0.1 & 37.9$\pm$0.8 \\
Kyoto  & 50  & 42.0$\pm$2.2 & 34.7$\pm$1.8 & 35.8$\pm$3.1 & 35.7$\pm$0.8 & 40.8$\pm$2.2 & \textbf{49.4$\pm$1.6} \\
Orange & 50  & 47.5$\pm$2.0 & 50.3$\pm$2.7 & 26.7$\pm$1.7 & 57.3$\pm$2.4 & 68.3$\pm$0.7 & \textbf{76.3$\pm$2.7} \\
\midrule
Milan  & 100 & 41.4$\pm$1.7 & 40.8$\pm$1.3 & 39.8$\pm$1.1 & 38.5$\pm$0.6 & 40.9$\pm$0.3 & \textbf{44.3$\pm$1.5} \\
Aruba  & 100 & \textbf{39.2$\pm$2.2} & 36.1$\pm$0.5 & 36.8$\pm$0.8 & 36.5$\pm$3.2 & 36.2$\pm$0.0 & 37.5$\pm$0.9 \\
Kyoto  & 100 & 41.8$\pm$2.4 & 34.6$\pm$1.6 & 34.5$\pm$2.8 & 35.7$\pm$1.3 & 38.8$\pm$1.4 & \textbf{47.3$\pm$2.1} \\
Orange & 100 & 49.2$\pm$2.1 & 52.7$\pm$2.8 & 26.6$\pm$1.3 & 55.6$\pm$3.2 & 68.6$\pm$1.6 & \textbf{74.1$\pm$3.6} \\
\bottomrule
\end{tabular}
}
\end{table}

Table~\ref{tab:raw_stream_stride_macro} shows that LastAct achieves the best Macro-F1 on all four datasets under the primary stride-1 setting. This setting is the most faithful raw-stream evaluation because it densely samples the continuous event stream and therefore most directly reflects online sliding-window inference. The coarser stride-50 and stride-100 settings show broadly consistent trends on Milan, Kyoto, and Orange, where LastAct remains the best-performing method. Aruba is the main exception: DeepCASAS slightly exceeds LastAct under the coarser strides. However, these coarser settings contain far fewer test windows, and several rare activities have single-digit support, making macro-level estimates more sensitive to individual prediction changes. We therefore treat stride~1 as the primary raw-stream result and use strides~50 and~100 as sensitivity analyses rather than as the main deployment protocol.

\subsubsection{Purity-Stratified Raw-Stream Performance}
\label{app:raw_stream_purity_performance}

Table~\ref{tab:raw_stream_per_purity_app} reports Macro-F1 by post-hoc purity bin for the primary stride-1 raw-stream setting. These bins are used only for analysis and do not affect window construction or model input. The results show that LastAct has the largest advantage in low- and mid-purity windows, where the latest activity occupies only a small fraction of the observation window and stale pre-boundary context is most misleading. In high-purity and pure windows, baselines become more competitive, which is expected because the task increasingly reduces to conventional single-activity recognition.

\begin{table}[htp]
\centering
\caption{Per-purity-bin Macro-F1 (\%) on the raw-stream test set. Bins denote the fraction of in-target (latest-activity) events within each test window; \textit{Pure} denotes windows entirely contained in a single annotated episode. Bins are used only for stratified post-hoc analysis and do not affect window construction or model input. Best per (dataset, bin) in \textbf{bold}; mean~$\pm$~std over three runs.}
\label{tab:raw_stream_per_purity_app}
\resizebox{0.7\textwidth}{!}{%
\begin{tabular}{lcccccc}
\toprule
Method & 0--20 & 21--40 & 41--60 & 61--80 & 81--99 & Pure \\
\midrule
\multicolumn{7}{l}{\textit{Aruba}} \\
\midrule
DeepCASAS & 8.6 $\pm$ 0.8 & 25.1 $\pm$ 3.1 & 54.3 $\pm$ 4.0 & 66.5 $\pm$ 1.7 & 68.4 $\pm$ 0.4 & 75.0 $\pm$ 1.1 \\
TCN       & 6.3 $\pm$ 1.5 & 22.9 $\pm$ 5.1 & 50.0 $\pm$ 0.5 & 66.2 $\pm$ 2.4 & 69.7 $\pm$ 1.3 & 76.0 $\pm$ 0.6 \\
CPC       & 4.3 $\pm$ 0.7 & 13.4 $\pm$ 1.1 & 47.6 $\pm$ 2.2 & \textbf{76.5 $\pm$ 2.0} & \textbf{82.8 $\pm$ 1.1} & 74.5 $\pm$ 0.7 \\
DCNN      & 6.3 $\pm$ 0.8 & 20.3 $\pm$ 2.0 & 52.4 $\pm$ 5.1 & 65.8 $\pm$ 1.6 & 69.4 $\pm$ 0.6 & 72.3 $\pm$ 0.7 \\
TDOST     & 8.0 $\pm$ 1.0 & 22.6 $\pm$ 1.1 & 47.5 $\pm$ 0.6 & 64.2 $\pm$ 0.4 & 68.0 $\pm$ 0.4 & \textbf{75.2 $\pm$ 0.6} \\
Ours      & \textbf{20.6 $\pm$ 2.2} & \textbf{37.3 $\pm$ 1.9} & \textbf{60.7 $\pm$ 2.4} & 68.7 $\pm$ 1.4 & 77.6 $\pm$ 1.1 & 70.3 $\pm$ 0.3 \\
\midrule
\multicolumn{7}{l}{\textit{Milan}} \\
\midrule
DeepCASAS & 8.5 $\pm$ 1.1 & 22.1 $\pm$ 0.7 & 37.3 $\pm$ 2.0 & 43.0 $\pm$ 0.4 & 55.0 $\pm$ 1.5 & \textbf{62.8 $\pm$ 0.2} \\
TCN       & 4.3 $\pm$ 0.7 & 13.1 $\pm$ 0.9 & 32.8 $\pm$ 1.4 & 43.3 $\pm$ 1.6 & \textbf{58.4 $\pm$ 1.9} & 62.4 $\pm$ 3.3 \\
CPC       & 5.2 $\pm$ 0.5 & 12.1 $\pm$ 1.5 & 29.4 $\pm$ 1.9 & 39.5 $\pm$ 2.9 & 55.3 $\pm$ 1.0 & 61.1 $\pm$ 4.0 \\
DCNN      & 6.4 $\pm$ 0.6 & 19.5 $\pm$ 2.2 & 33.7 $\pm$ 0.7 & 43.0 $\pm$ 1.2 & 56.2 $\pm$ 2.1 & 59.2 $\pm$ 2.9 \\
TDOST     & 7.0 $\pm$ 1.6 & 18.1 $\pm$ 2.7 & 34.6 $\pm$ 0.9 & 42.3 $\pm$ 1.2 & 54.7 $\pm$ 2.2 & 61.1 $\pm$ 0.8 \\
Ours      & \textbf{21.3 $\pm$ 1.9} & \textbf{36.4 $\pm$ 1.9} & \textbf{45.0 $\pm$ 0.5} & \textbf{44.8 $\pm$ 0.6} & 51.7 $\pm$ 3.1 & 51.2 $\pm$ 1.8 \\
\midrule
\multicolumn{7}{l}{\textit{Kyoto}} \\
\midrule
DeepCASAS & 14.3 $\pm$ 2.6 & 32.0 $\pm$ 1.3 & 40.3 $\pm$ 1.5 & 44.1 $\pm$ 1.6 & 50.3 $\pm$ 4.1 & 46.1 $\pm$ 3.1 \\
TCN       & 7.5 $\pm$ 2.1 & 11.8 $\pm$ 2.1 & 20.5 $\pm$ 2.2 & 32.5 $\pm$ 6.0 & 47.9 $\pm$ 6.1 & 44.5 $\pm$ 1.6 \\
CPC       & 7.1 $\pm$ 1.4 & 9.4 $\pm$ 1.9 & 14.9 $\pm$ 2.7 & 30.2 $\pm$ 4.3 & 47.0 $\pm$ 5.3 & 46.4 $\pm$ 0.5 \\
DCNN      & 5.3 $\pm$ 1.4 & 8.7 $\pm$ 2.4 & 24.0 $\pm$ 1.9 & 38.0 $\pm$ 2.3 & \textbf{52.3 $\pm$ 3.3} & 42.1 $\pm$ 0.7 \\
TDOST     & 14.2 $\pm$ 2.1 & 21.0 $\pm$ 3.2 & 18.3 $\pm$ 2.0 & 25.1 $\pm$ 0.3 & 42.8 $\pm$ 2.6 & \textbf{49.1 $\pm$ 2.1} \\
Ours      & \textbf{25.3 $\pm$ 2.4} & \textbf{47.7 $\pm$ 8.5} & \textbf{46.1 $\pm$ 5.3} & \textbf{50.7 $\pm$ 1.7} & 50.1 $\pm$ 0.6 & 47.5 $\pm$ 0.4 \\
\midrule
\multicolumn{7}{l}{\textit{Orange}} \\
\midrule
DeepCASAS & 17.3 $\pm$ 2.0 & 19.9 $\pm$ 1.4 & 27.0 $\pm$ 1.1 & 29.5 $\pm$ 3.0 & 37.4 $\pm$ 2.5 & 48.8 $\pm$ 2.0 \\
TCN       & 7.0 $\pm$ 1.4 & 10.3 $\pm$ 1.7 & 14.4 $\pm$ 1.6 & 15.8 $\pm$ 1.9 & 20.0 $\pm$ 2.9 & 28.7 $\pm$ 2.3 \\
CPC       & 11.5 $\pm$ 1.0 & 18.7 $\pm$ 2.4 & 34.4 $\pm$ 8.6 & 48.4 $\pm$ 7.0 & 56.7 $\pm$ 7.8 & 55.5 $\pm$ 5.2 \\
DCNN      & 11.4 $\pm$ 1.3 & 17.0 $\pm$ 3.3 & 26.2 $\pm$ 4.6 & 39.8 $\pm$ 5.3 & 60.3 $\pm$ 1.0 & 52.7 $\pm$ 2.3 \\
TDOST     & 19.4 $\pm$ 4.3 & 36.7 $\pm$ 5.2 & 53.1 $\pm$ 3.7 & 62.0 $\pm$ 1.1 & \textbf{73.4 $\pm$ 1.0} & 70.6 $\pm$ 0.4 \\
Ours      & \textbf{24.0 $\pm$ 7.5} & \textbf{45.7 $\pm$ 6.7} & \textbf{58.5 $\pm$ 4.1} & \textbf{66.3 $\pm$ 6.5} & 67.6 $\pm$ 2.9 & \textbf{74.9 $\pm$ 14.7} \\
\bottomrule
\end{tabular}%
}
\end{table}

\subsubsection{Per-Activity Raw-Stream Performance}
\label{app:raw_stream_per_activity}

To further examine whether the aggregate raw-stream gains are driven only by frequent activities, we report per-activity F1 scores under the primary stride-1 raw-stream protocol. Support denotes the number of raw-stream test windows whose latest-activity label corresponds to each activity. Macro-F1 is computed by averaging F1 equally across activity labels, so rare activities contribute equally to the final score despite having much smaller support.

Tables~\ref{tab:raw_stream_activity_milan}--\ref{tab:raw_stream_activity_orange} show that LastAct improves many transition-sensitive and minority activities, but the improvements are not uniform across all classes. On Milan, LastAct improves activities such as \textit{Bed\_to\_Toilet}, \textit{Guest\_Bathroom}, \textit{Master\_Bathroom}, \textit{Kitchen\_Activity}, and \textit{Read}. On Aruba, the gains are particularly visible for rare or short activities such as \textit{Bed\_to\_Toilet}, \textit{Wash\_Dishes}, \textit{Enter\_Home}, and \textit{Leave\_Home}, while the method remains competitive on high-support activities such as \textit{Relax} and \textit{Sleeping}. On Kyoto, LastAct improves several resident-specific activities, including \textit{R1\_Bed\_to\_Toilet}, \textit{R1\_Personal\_Hygiene}, \textit{R2\_Bed\_to\_Toilet}, and \textit{R2\_Work}. On Orange, the improvement is especially strong for difficult or ambiguous labels such as \textit{Other}, \textit{Sleep}, \textit{Enter\_Home}, and \textit{Personal\_Hygiene}. These patterns indicate that the aggregate improvement is not solely due to dominant classes, but also reflects better handling of activities whose latest evidence may occupy only a short suffix of the input window.

The per-activity results also reveal remaining failure cases under severe class sparsity. For example, \textit{Eve\_Meds} on Milan has only 113 raw-stream test windows and receives 0 F1 from all methods, including LastAct. Similarly, \textit{Morning\_Meds} on Milan and \textit{Study} on Kyoto remain poorly recognized. These cases suggest that boundary-aware inference can reduce stale-context interference, but it cannot by itself overcome extreme label sparsity or weakly represented activity signatures. This limitation is consistent with the long-tail class imbalance commonly observed in smart-home HAR and motivates future work on rare-activity augmentation, cost-sensitive training, or few-shot adaptation.

At the same time, no single method dominates every activity. Strong baselines remain competitive for some high-support or semantically informative activities, such as \textit{Meal\_Preparation} on Aruba and \textit{Bathing}, \textit{Cook}, and \textit{Work} on Orange. This behavior is expected because many raw-stream windows are high-purity or fully pure, in which case the task increasingly resembles conventional single-activity recognition. Therefore, LastAct's advantage should be interpreted as improved robustness across the full raw-stream activity distribution, especially for transition-sensitive and minority activities, rather than uniform improvement on every individual class.

These per-activity results are also consistent with the design motivation of LastAct. In raw-stream inference, some windows contain only a few events from the newly started activity while most of the window still reflects the previous activity. Architectures that aggregate information over the full window can be dominated by stale pre-boundary context in such cases. LastAct explicitly addresses this setting by estimating the last activity boundary and suppressing pre-boundary evidence during cross-window inference, which explains its stronger performance on many short, transitional, and minority activities.  

\begin{table*}[htbp]
\centering
\scriptsize
\caption{Per-activity F1 (\%) on Milan under the stride-1 raw-stream protocol. Support denotes the number of raw-stream test windows assigned to each latest-activity label. Best result per activity is shown in \textbf{bold}.}
\label{tab:raw_stream_activity_milan}
\resizebox{0.95\textwidth}{!}{%
\begin{tabular}{lrrrrrrr}
\toprule
Activity & Support & DeepCASAS & DCNN & TCN & CPC & TDOST & Ours \\
\midrule
Bed\_to\_Toilet & 500 & 0.0$\pm$0.0 & 0.0$\pm$0.0 & 0.0$\pm$0.0 & 0.0$\pm$0.0 & 0.0$\pm$0.0 & \textbf{19.4$\pm$3.5} \\
Sleep & 6662 & 57.6$\pm$2.4 & 56.9$\pm$1.8 & 58.7$\pm$3.7 & 56.2$\pm$2.4 & \textbf{61.0$\pm$1.8} & 58.0$\pm$5.5 \\
Morning\_Meds & 535 & 0.0$\pm$0.0 & 0.0$\pm$0.0 & 0.0$\pm$0.0 & 0.0$\pm$0.0 & 0.0$\pm$0.0 & \textbf{0.23$\pm$0.17} \\
Watch\_TV & 5411 & 57.1$\pm$0.9 & 55.8$\pm$1.9 & 52.6$\pm$1.4 & 51.7$\pm$1.3 & 39.2$\pm$1.0 & \textbf{59.4$\pm$2.9} \\
Kitchen\_Activity & 33311 & 77.3$\pm$0.4 & 74.4$\pm$0.5 & 73.1$\pm$0.4 & 72.3$\pm$1.8 & 73.2$\pm$1.1 & \textbf{80.3$\pm$1.0} \\
Leave\_Home & 584 & 22.1$\pm$4.7 & 16.6$\pm$1.3 & 18.8$\pm$5.3 & 18.6$\pm$4.9 & \textbf{27.2$\pm$1.3} & 20.7$\pm$1.9 \\
Read & 10529 & 70.5$\pm$0.5 & 68.5$\pm$0.8 & 66.2$\pm$1.1 & 63.1$\pm$1.8 & 56.3$\pm$2.2 & \textbf{77.0$\pm$1.0} \\
Guest\_Bathroom & 3559 & 23.3$\pm$1.1 & 18.9$\pm$2.2 & 14.7$\pm$0.3 & 13.1$\pm$1.4 & 17.2$\pm$3.6 & \textbf{39.3$\pm$2.5} \\
Master\_Bathroom & 5501 & 26.7$\pm$2.6 & 25.6$\pm$3.5 & 23.5$\pm$3.1 & 23.6$\pm$3.6 & 31.6$\pm$1.8 & \textbf{40.3$\pm$4.4} \\
Desk\_Activity & 3341 & 69.0$\pm$1.2 & 65.6$\pm$4.1 & 64.5$\pm$5.7 & 66.5$\pm$3.3 & \textbf{69.6$\pm$1.3} & 69.3$\pm$1.6 \\
Eve\_Meds & 113 & \textbf{0.0$\pm$0.0} & \textbf{0.0$\pm$0.0} & \textbf{0.0$\pm$0.0} & \textbf{0.0$\pm$0.0} & \textbf{0.0$\pm$0.0} & \textbf{0.0$\pm$0.0} \\
Meditate & 167 & \textbf{28.9$\pm$3.4} & 22.2$\pm$4.3 & 12.9$\pm$4.1 & 9.9$\pm$2.8 & 19.0$\pm$1.2 & 24.0$\pm$5.6 \\
Dining\_Rm\_Activity & 2171 & 55.8$\pm$3.3 & 60.1$\pm$1.9 & \textbf{60.6$\pm$2.2} & 54.8$\pm$2.2 & 58.8$\pm$4.0 & 53.3$\pm$1.1 \\
Master\_Bedroom\_Activity & 11660 & 25.7$\pm$2.3 & 34.3$\pm$6.8 & \textbf{36.4$\pm$9.5} & 28.6$\pm$7.5 & \textbf{36.4$\pm$2.8} & 19.8$\pm$9.7 \\
\midrule
Macro-F1 & 84044 & 36.7$\pm$0.3 & 35.6$\pm$1.0 & 34.4$\pm$0.9 & 32.7$\pm$0.7 & 34.9$\pm$0.5 & \textbf{40.1$\pm$0.5} \\
\bottomrule
\end{tabular}%
}
\end{table*}

\begin{table*}[htbp]
\centering
\scriptsize
\caption{Per-activity F1 (\%) on Aruba under the stride-1 raw-stream protocol. Support denotes the number of raw-stream test windows assigned to each latest-activity label. Best result per activity is shown in \textbf{bold}.}
\label{tab:raw_stream_activity_aruba}
\resizebox{0.95\textwidth}{!}{%
\begin{tabular}{lrrrrrrr}
\toprule
Activity & Support & DeepCASAS & DCNN & TCN & CPC & TDOST & Ours \\
\midrule
Meal\_Preparation & 62457 & 83.0$\pm$0.9 & 82.8$\pm$1.9 & 81.6$\pm$0.4 & 79.4$\pm$0.6 & \textbf{84.0$\pm$0.8} & 75.2$\pm$3.4 \\
Relax & 68761 & 85.9$\pm$2.5 & 84.5$\pm$0.6 & 84.8$\pm$0.7 & 82.0$\pm$0.8 & \textbf{88.1$\pm$0.7} & 88.0$\pm$0.3 \\
Eating & 2977 & \textbf{61.4$\pm$4.5} & 49.1$\pm$2.4 & 53.1$\pm$0.4 & 51.7$\pm$2.4 & 48.3$\pm$4.5 & 53.6$\pm$0.2 \\
Work & 3903 & 70.5$\pm$2.0 & 69.8$\pm$3.4 & 68.0$\pm$1.3 & 65.0$\pm$0.7 & 65.9$\pm$4.1 & \textbf{78.3$\pm$0.7} \\
Sleeping & 7466 & 81.9$\pm$0.8 & 75.9$\pm$1.3 & 77.0$\pm$1.5 & 75.0$\pm$0.7 & 78.5$\pm$1.0 & \textbf{82.0$\pm$0.7} \\
Wash\_Dishes & 570 & 0.0$\pm$0.0 & 0.0$\pm$0.0 & 2.6$\pm$0.3 & 1.6$\pm$0.2 & 0.0$\pm$0.0 & \textbf{4.2$\pm$0.6} \\
Bed\_to\_Toilet & 203 & 0.0$\pm$0.0 & 0.0$\pm$0.0 & 0.0$\pm$0.0 & 0.0$\pm$0.0 & 0.0$\pm$0.0 & \textbf{32.2$\pm$8.3} \\
Enter\_Home & 519 & 0.3$\pm$0.3 & 2.2$\pm$1.3 & 0.0$\pm$0.0 & 0.0$\pm$0.0 & 0.0$\pm$0.0 & \textbf{3.4$\pm$0.2} \\
Leave\_Home & 433 & 2.9$\pm$2.3 & 0.0$\pm$0.0 & 3.0$\pm$0.3 & 0.6$\pm$0.0 & 0.0$\pm$0.0 & \textbf{5.1$\pm$3.5} \\
Respirate & 163 & 0.0$\pm$0.0 & 0.0$\pm$0.0 & 0.0$\pm$0.0 & \textbf{29.1$\pm$2.6} & 0.0$\pm$0.0 & 23.8$\pm$2.7 \\
\midrule
Macro-F1 & 147452 & 38.6$\pm$0.5 & 36.4$\pm$0.7 & 37.0$\pm$0.9 & 38.4$\pm$0.5 & 36.5$\pm$0.1 & \textbf{44.6$\pm$1.9} \\
\bottomrule
\end{tabular}%
}
\end{table*}

\begin{table*}[htp]
\centering
\scriptsize
\caption{Per-activity F1 (\%) on Kyoto under the stride-1 raw-stream protocol. Support denotes the number of raw-stream test windows assigned to each latest-activity label. Best result per activity is shown in \textbf{bold}.}
\label{tab:raw_stream_activity_kyoto}
\resizebox{0.95\textwidth}{!}{%
\begin{tabular}{lrrrrrrr}
\toprule
Activity & Support & DeepCASAS & DCNN & TCN & CPC & TDOST & Ours \\
\midrule
R1\_Bed\_to\_Toilet & 569 & 1.6$\pm$0.2 & 3.9$\pm$0.6 & 16.1$\pm$9.3 & 8.9$\pm$5.3 & 8.8$\pm$3.7 & \textbf{37.2$\pm$0.8} \\
R1\_Sleep & 2596 & 32.4$\pm$13.2 & 40.2$\pm$2.9 & 38.4$\pm$0.5 & 33.7$\pm$4.9 & \textbf{54.5$\pm$3.4} & 38.7$\pm$1.3 \\
R1\_Personal\_Hygiene & 1564 & 14.9$\pm$9.3 & 10.5$\pm$9.6 & 15.8$\pm$9.0 & 27.8$\pm$3.5 & 27.9$\pm$10.7 & \textbf{49.3$\pm$0.9} \\
Watch\_TV & 2071 & 64.7$\pm$5.0 & 35.2$\pm$8.5 & 43.7$\pm$7.5 & 51.1$\pm$10.8 & 48.8$\pm$1.7 & \textbf{64.8$\pm$5.6} \\
Meal\_Preparation & 7463 & \textbf{82.1$\pm$2.5} & 71.8$\pm$1.6 & 69.8$\pm$1.9 & 66.7$\pm$2.8 & 74.6$\pm$2.3 & 77.8$\pm$3.5 \\
R1\_Work & 6267 & \textbf{82.5$\pm$0.5} & 76.7$\pm$2.1 & 76.4$\pm$1.0 & 69.4$\pm$1.6 & 79.8$\pm$1.6 & 71.4$\pm$1.0 \\
Study & 133 & 0.0$\pm$0.0 & 0.0$\pm$0.0 & 0.0$\pm$0.0 & 0.0$\pm$0.0 & \textbf{3.0$\pm$3.0} & 0.0$\pm$0.0 \\
R2\_Bed\_to\_Toilet & 506 & 6.3$\pm$0.9 & 0.0$\pm$0.0 & 9.2$\pm$6.8 & 6.9$\pm$2.7 & 41.6$\pm$1.7 & \textbf{50.8$\pm$1.8} \\
R2\_Personal\_Hygiene & 1867 & \textbf{52.3$\pm$1.8} & 45.4$\pm$1.7 & 42.9$\pm$0.6 & 41.0$\pm$5.0 & 41.2$\pm$8.2 & 41.9$\pm$1.1 \\
R2\_Sleep & 3957 & \textbf{76.6$\pm$0.7} & 67.7$\pm$3.8 & 64.5$\pm$1.2 & 56.2$\pm$3.0 & 61.1$\pm$0.6 & 63.9$\pm$0.2 \\
R2\_Work & 1222 & 37.8$\pm$3.9 & 25.2$\pm$2.7 & 25.2$\pm$2.3 & 28.6$\pm$5.4 & 26.6$\pm$6.7 & \textbf{38.6$\pm$3.2} \\
\midrule
Macro-F1 & 28215 & 41.0$\pm$1.8 & 34.2$\pm$0.6 & 36.6$\pm$2.0 & 35.5$\pm$1.5 & 42.5$\pm$1.2 & \textbf{48.6$\pm$8.5} \\
\bottomrule
\end{tabular}%
}
\end{table*}

\begin{table*}[htp]
\centering
\scriptsize
\caption{Per-activity F1 (\%) on Orange under the stride-1 raw-stream protocol. Support denotes the number of raw-stream test windows assigned to each latest-activity label. Best result per activity is shown in \textbf{bold}.}
\label{tab:raw_stream_activity_orange}
\resizebox{0.95\textwidth}{!}{%
\begin{tabular}{lrrrrrrr}
\toprule
Activity & Support & DeepCASAS & DCNN & TCN & CPC & TDOST & Ours \\
\midrule
Bathing & 4203 & 43.2$\pm$8.9 & 39.2$\pm$0.5 & 27.3$\pm$8.5 & 61.7$\pm$6.0 & \textbf{92.2$\pm$1.3} & 90.3$\pm$0.8 \\
Cook & 5107 & 61.4$\pm$1.2 & 56.0$\pm$2.5 & 34.2$\pm$2.9 & 79.2$\pm$3.2 & \textbf{80.3$\pm$1.6} & 71.3$\pm$10.8 \\
Enter\_Home & 546 & 13.9$\pm$2.3 & 31.0$\pm$4.5 & 0.0$\pm$0.0 & 32.3$\pm$9.4 & 44.8$\pm$1.3 & \textbf{55.5$\pm$8.0} \\
Leave\_Home & 567 & 20.2$\pm$3.5 & 18.6$\pm$4.4 & 0.0$\pm$0.0 & 30.4$\pm$3.2 & \textbf{73.4$\pm$0.1} & 71.8$\pm$3.1 \\
Other & 6481 & 3.0$\pm$0.6 & 16.0$\pm$2.1 & 0.0$\pm$0.0 & 7.9$\pm$1.1 & 11.5$\pm$1.1 & \textbf{50.1$\pm$24.0} \\
Personal\_Hygiene & 2127 & 26.4$\pm$2.6 & 50.1$\pm$9.6 & 0.0$\pm$0.0 & 44.9$\pm$4.9 & 64.8$\pm$1.4 & \textbf{71.9$\pm$5.6} \\
Relax & 10480 & 72.2$\pm$1.3 & 73.3$\pm$1.0 & 61.9$\pm$1.6 & 78.7$\pm$3.5 & 78.4$\pm$0.5 & \textbf{84.5$\pm$4.6} \\
Sleep & 5383 & 31.1$\pm$2.1 & 24.8$\pm$5.0 & 16.7$\pm$2.0 & 44.5$\pm$3.0 & 24.9$\pm$3.8 & \textbf{63.6$\pm$12.7} \\
Work & 94598 & 96.7$\pm$0.3 & 93.3$\pm$0.1 & 96.5$\pm$0.4 & \textbf{97.9$\pm$0.1} & 97.7$\pm$0.4 & 97.5$\pm$1.1 \\
\midrule
Macro-F1 & 129492 & 40.9$\pm$0.9 & 44.7$\pm$1.9 & 26.3$\pm$1.4 & 53.1$\pm$2.5 & 63.1$\pm$0.6 & \textbf{72.9$\pm$7.8} \\
\bottomrule
\end{tabular}%
}
\end{table*}

\begin{figure}[t]
    \centering
    \includegraphics[width=0.95\linewidth]{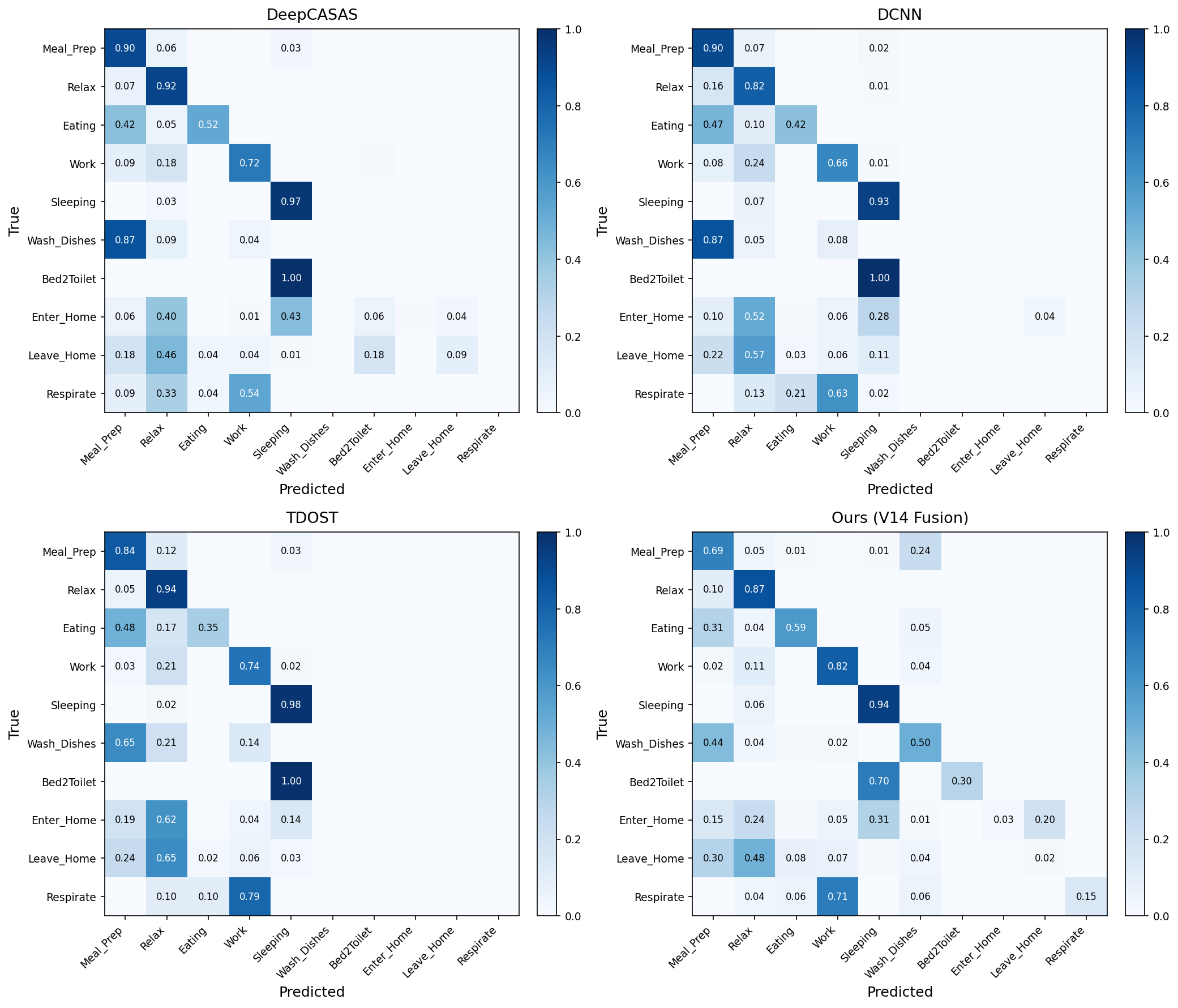}
    \caption{%
    Confusion matrices of DeepCASAS, TDOST, DCNN, and our method on Aruba.
    }
    \label{fig:aruba_cm_raw_str}
\end{figure}
To further analyze these behaviors, we examine the confusion matrices of representative models, including DeepCASAS, DCNN, TDOST, and our method in Fig~\ref{fig:aruba_cm_raw_str}. DeepCASAS, DCNN, and TDOST achieve strong recall on activities with stable and distinctive sensor patterns, such as Sleeping (0.93–0.98) and Bed2Toilet (1.00). However, they consistently fail on Wash\_Dishes, with 65–87\% of its instances misclassified as Meal\_Prep. This systematic error indicates a shared limitation: these models struggle to distinguish between co-located kitchen activities that activate highly overlapping sensor regions.
In contrast, our method breaks this ambiguity by correctly identifying Wash\_Dishes with a recall of 0.50—the only method to achieve non-trivial recognition for this class—while also attaining the highest recall on Work (0.82) and Eating (0.59). This improved discrimination within the kitchen domain introduces a measurable trade-off: 24\% of Meal\_Prep instances are misclassified as Wash\_Dishes, reducing its recall from 0.84–0.90 (baseline range) to 0.69. This reflects a deliberate refinement of the decision boundary between two previously conflated activities.

Overall, the confusion matrix analysis shows that our method improves discrimination among semantically similar and previously conflated activities, particularly in the kitchen domain, while introducing limited trade-offs on spatially or temporally adjacent activities. These results highlight both the benefits and inherent difficulty of fine-grained activity disambiguation in smart home environments.
\subsection{Sensitivity to Continuous-Value Bin Size}
\label{app:bin_sensitivity}
For sensors with continuous readings, our trajectory construction discretizes each value 
into one of \(B\) bins and uses the corresponding sensor-bin template during trajectory 
rendering. This discretization is necessary for the reusable template-cache design, but 
it may introduce information loss if \(B\) is too small or unnecessary sparsity if \(B\) 
is too large. We therefore evaluate the sensitivity of the model to the number of 
continuous-value bins using \(B \in \{4,8,16,32\}\).

We conduct this ablation on Orange because it contains the richest set of continuous and 
heterogeneous sensors among the evaluated datasets. In contrast, Milan, Aruba, and Kyoto7 
are dominated by binary motion/door events, with only sparse continuous temperature 
readings; therefore, varying \(B\) on those datasets affects only a small subset of 
events and is less informative for evaluating the discretization mechanism.

Table~\ref{tab:orange_bin_pure_cross} reports pure-window and overall cross-window 
Macro-F1 under different bin counts. Overall, the model remains stable across bin 
choices. Pure-window Macro-F1 ranges from \(58.54\%\) to \(62.28\%\), while overall 
cross-window Macro-F1 ranges from \(59.93\%\) to \(64.77\%\). The differences are modest 
and do not show a monotonic relationship with \(B\), suggesting that the method is not 
highly sensitive to the exact bin count.

Table~\ref{tab:orange_bin_purity} further reports cross-window performance by activity 
purity. The same conclusion holds across low-, mid-, and high-purity windows. In the 
10--30\% purity range, Macro-F1 varies from \(49.89\%\) to \(56.57\%\); in the 40--60\% 
range, it varies from \(62.89\%\) to \(67.79\%\); and in the 70--90\% range, it varies 
from \(64.18\%\) to \(69.84\%\). Although \(B=32\) obtains the highest overall 
cross-window Macro-F1 on Orange, the gains are small relative to the fold-level 
variation and do not justify increasing the template-cache size for all datasets. We 
therefore retain \(B=8\) as the default setting because it provides a compact and 
computationally efficient representation while achieving comparable performance.

\begin{table}[t]
\centering
\caption{Sensitivity of Orange performance to the number of continuous-value bins. 
Results are Macro-F1 (\%) under 3-fold cross-validation.}
\label{tab:orange_bin_pure_cross}
\resizebox{0.58\textwidth}{!}{%
\begin{tabular}{c|cc}
\toprule
Number of bins \(B\) & Pure-window Macro-F1 & Cross-window Macro-F1 \\
\midrule
4  & \(62.01 \pm 4.13\) & \(61.42 \pm 5.31\) \\
8  & \(58.54 \pm 3.67\) & \(63.26 \pm 4.22\) \\
16 & \(62.28 \pm 3.50\) & \(59.93 \pm 2.65\) \\
32 & \(60.85 \pm 6.44\) & \(64.77 \pm 2.12\) \\
\bottomrule
\end{tabular}
}
\end{table}

\begin{table}[t]
\centering
\caption{Sensitivity of Orange cross-window performance to the number of continuous-value 
bins across activity-purity ranges. Results are Macro-F1 (\%) under 3-fold cross-validation.}
\label{tab:orange_bin_purity}
\resizebox{0.58\textwidth}{!}{%
\begin{tabular}{c|ccc}
\toprule
Number of bins \(B\) & 10--30\% purity & 40--60\% purity & 70--90\% purity \\
\midrule
4  & \(53.18 \pm 4.50\) & \(65.54 \pm 5.55\) & \(64.18 \pm 5.64\) \\
8  & \(55.22 \pm 3.53\) & \(66.54 \pm 2.59\) & \(67.00 \pm 7.13\) \\
16 & \(49.89 \pm 3.47\) & \(62.89 \pm 3.82\) & \(65.20 \pm 2.87\) \\
32 & \(56.57 \pm 4.23\) & \(67.79 \pm 2.01\) & \(69.84 \pm 5.34\) \\
\bottomrule
\end{tabular}
}
\end{table}

% \end{revblock}

\subsection{Ablation of Cyclic Temporal Encoding}
\label{app:cyclic_temporal}
Table~\ref{tab:temporal_ablation_macro_f1} analyzes the contribution of different temporal components by enabling weekday, hour, and minute cues in various combinations. For ablation studies, when no temporal features are enabled, the time embedding is explicitly set to zeros, guaranteeing a fair comparison by keeping the model architecture unchanged while removing temporal cues.

\begin{table}[htbp]
\centering
\small
\setlength{\tabcolsep}{6pt}
\renewcommand{\arraystretch}{1.15}
\caption{\textbf{Ablation study on temporal components.} Macro-F1 (mean$\pm$std) across four datasets under different temporal cue configurations (Weekday, Hour, Minute). Each cell reports \emph{Pure / Cross} Macro-F1 (mean$\pm$std).}
\label{tab:temporal_ablation_macro_f1}
\resizebox{0.95\textwidth}{!}{%
\begin{tabular}{ccc|cccc}
\toprule
\multicolumn{3}{c|}{\textbf{Temporal Components}} & \multicolumn{4}{c}{\textbf{Datasets (Pure / Cross)}} \\
\cmidrule(lr){1-3} \cmidrule(lr){4-7}
\textbf{Weekday} & \textbf{Hour} & \textbf{Minute} &
\textbf{Milan} & \textbf{Aruba} & \textbf{Kyoto} & \textbf{Orange} \\
\midrule
\textcolor{red}{X} & \textcolor{red}{X} & \textcolor{red}{X} &
68.45 $\pm$ 0.73 / 52.31 $\pm$ 0.79 &
74.03 $\pm$ 2.15 / 56.40 $\pm$ 1.28 &
\underline{63.06 $\pm$ 4.68} / \underline{58.69 $\pm$ 1.49} &
64.18 $\pm$ 2.69 / 56.35 $\pm$ 5.26 \\
\textcolor{green}{Y} & \textcolor{red}{X} & \textcolor{red}{X} &
69.48 $\pm$ 2.28 / 53.96 $\pm$ 0.88 &
\underline{74.17 $\pm$ 1.50} / 56.27 $\pm$ 1.80 &
58.42 $\pm$ 4.03 / \textbf{59.85 $\pm$ 2.71} &
61.00 $\pm$ 3.63 / 62.49 $\pm$ 2.02 \\
\textcolor{red}{X} & \textcolor{green}{Y} & \textcolor{red}{X} &
68.78 $\pm$ 3.54 / 52.81 $\pm$ 1.74 &
72.82 $\pm$ 0.61 / 55.80 $\pm$ 1.35 &
60.95 $\pm$ 2.60 / 57.61 $\pm$ 1.75 &
\textbf{67.30 $\pm$ 2.97} / 62.77 $\pm$ 1.54 \\
\textcolor{red}{X} & \textcolor{red}{X} & \textcolor{green}{Y} &
\underline{72.91 $\pm$ 1.13} / \textbf{56.11 $\pm$ 1.62} &
72.65 $\pm$ 1.40 / 56.12 $\pm$ 1.98 &
59.44 $\pm$ 5.37 / 57.22 $\pm$ 4.88 &
58.00 $\pm$ 3.31 / \underline{63.43 $\pm$ 1.22} \\
\midrule
\textcolor{green}{Y} & \textcolor{green}{Y} & \textcolor{red}{X} &
68.66 $\pm$ 2.32 / 52.06 $\pm$ 2.64 &
74.05 $\pm$ 1.20 / 54.87 $\pm$ 0.43 &
62.28 $\pm$ 3.22 / 56.81 $\pm$ 2.66 &
\underline{67.04 $\pm$ 5.12} / 60.90 $\pm$ 2.53 \\
\textcolor{red}{X} & \textcolor{green}{Y} & \textcolor{green}{Y} &
71.04 $\pm$ 1.51 / 55.48 $\pm$ 0.60 &
73.07 $\pm$ 1.82 / \underline{56.43 $\pm$ 1.20} &
\textbf{65.56 $\pm$ 1.71} / 57.75 $\pm$ 2.99 &
64.06 $\pm$ 3.73 / 62.93 $\pm$ 1.77 \\
\textcolor{green}{Y} & \textcolor{red}{X} & \textcolor{green}{Y} &
72.04 $\pm$ 0.78 / \underline{55.74 $\pm$ 2.70} &
73.58 $\pm$ 1.23 / 56.14 $\pm$ 1.64 &
63.02 $\pm$ 3.08 / 57.73 $\pm$ 4.37 &
63.68 $\pm$ 4.81 / 61.82 $\pm$ 3.78 \\
\midrule
\textcolor{green}{Y} & \textcolor{green}{Y} & \textcolor{green}{Y} &
\textbf{73.40 $\pm$ 1.50} / 54.54 $\pm$ 1.68 &
\textbf{76.02 $\pm$ 1.53} / \textbf{58.73 $\pm$ 1.40} &
60.04 $\pm$ 6.86 / 54.03 $\pm$ 2.73 &
63.85 $\pm$ 3.04 / \textbf{64.39 $\pm$ 3.33} \\
\bottomrule
\end{tabular}}
\end{table}

A first observation from Table~\ref{tab:temporal_ablation_macro_f1} is that none of the temporal configurations dramatically changes performance relative to the trajectory-only baseline (``none''). On pure windows, all time variants stay within roughly $\pm 3$ F1 points of the baseline across homes, and several configurations are essentially tied within variance. This indicates that the floorplan-aligned trajectory representation already captures most of the discriminative structure when episodes are cleanly segmented.

Second, there is no single temporal configuration that dominates across all homes. For instance, using all time components (hour + weekday + minute) yields the best pure and cross performance on Aruba and the best cross performance on Orange, while Milan benefits most from minute-only encodings on cross windows, and Kyoto prefers weekday-only encodings for cross windows and hour+minute for pure windows. In several cases, over-parameterized temporal channels (e.g., ``all'' on Kyoto cross windows) even underperform the time-free baseline.

Given this heterogeneity, we adopt the {hour, weekday} configuration as a conservative, low-capacity temporal side-channel in our main experiments. As shown in Table~\ref{tab:temporal_ablation_macro_f1}, hour+weekday is never dramatically worse than the best-performing configuration on any dataset, remains competitive with the time-free backbone on pure windows, and provides consistent though modest gains on Orange, where daily and weekly routines are more structured. At the same time, its negative impact is limited on datasets where temporal cues are less informative, which makes it a reasonable default for cross-home comparisons.
Minute-level encodings behave like a high-frequency time channel: they slightly improve Milan cross-window F1 (from 0.5231 to 0.5611) but can hurt pure-window performance on Kyoto and Orange. This pattern supports our design principle of treating minute as an optional ablation factor rather than a core component of the main model.

The Orange home---a two-resident, multi-storey environment with annotated daily routines---exhibits the largest gains from temporal cues on cross windows: cross F1 increases from 0.5635 (none) to 0.6439 (all). Although residents do not follow the scripted schedule perfectly, most days roughly adhere to the prescribed routine, so hour- and weekday-level patterns carry strong predictive signal that cyclic encodings can exploit. In contrast, the other homes correspond to more naturalistic deployments where residents follow less rigid schedules; in these settings, temporal side-channels can still provide small improvements on certain datasets, but human behavior is not strictly periodic and simple calendar-based encodings cannot reliably capture all the variability, which explains the more modest and sometimes even negative effects.

% \begin{revblock}
\subsection{Sensitivity to Rendering Choices}
\label{app:color-disk_sensitivity}

We assess how sensitive the trajectory representation is to the two rendering
hyperparameters introduced in Section~\ref{sec:trajectory_backbone}:the color
assignment used to encode sensor modality, and the geometric primitive used
to draw each sensor. All other components (encoder architecture, optimizer,
window construction, splits, and seeds) are held fixed; only the template
bank $\mathcal{B}$ is re-rendered.

\paragraph{Sensor-type color encoding.}
We first compare the default modality-aware \emph{Color} assignment against
a \emph{White} baseline in which every sensor is rendered with an identical
(white) disk, so that any sensor-type information must instead be inferred
from spatial location alone. Table~\ref{tab:color-ablation} reports macro-F1
on pure and cross-activity windows.

\begin{table}[t]
\centering
\caption{Effect of sensor-type color encoding on macro-F1 (\%, mean $\pm$ std
over $3$ seeds). \emph{Color} assigns each sensor modality a distinct color;
\emph{White} renders all sensors identically. Best per column in \textbf{bold}.}
\label{tab:color-ablation}
\small
\resizebox{0.95\textwidth}{!}{%
\begin{tabular}{l cc cc cc cc}
\toprule
& \multicolumn{2}{c}{Aruba} & \multicolumn{2}{c}{Milan}
& \multicolumn{2}{c}{Kyoto} & \multicolumn{2}{c}{Orange} \\
\cmidrule(lr){2-3}\cmidrule(lr){4-5}\cmidrule(lr){6-7}\cmidrule(lr){8-9}
Setting & Pure & Cross & Pure & Cross & Pure & Cross & Pure & Cross \\
\midrule
Color & \textbf{72.37\,$\pm$\,0.29} & \textbf{56.87\,$\pm$\,0.60}
      & \textbf{69.25\,$\pm$\,0.39} & 53.06\,$\pm$\,0.85
      & 59.95\,$\pm$\,2.65 & \textbf{57.64\,$\pm$\,5.94}
      & \textbf{62.28\,$\pm$\,2.85} & 59.93\,$\pm$\,2.17 \\
White & 67.07\,$\pm$\,5.00 & 52.90\,$\pm$\,1.66
      & 67.50\,$\pm$\,0.85 & \textbf{53.15\,$\pm$\,1.07}
      & \textbf{62.38\,$\pm$\,1.70} & 54.16\,$\pm$\,1.10
      & 57.72\,$\pm$\,5.18 & \textbf{59.98\,$\pm$\,0.96} \\
\bottomrule
\end{tabular}
}
\end{table}

Across all four datasets, the modality-aware color encoding consistently
outperforms the uniform-white variant in both pure and cross-window settings,
indicating the importance of preserving sensor-type heterogeneity in the
spatial encoding. The effect is most pronounced on Aruba, where Pure-F1 drops
from $72.37 \pm 0.29$ to $67.07 \pm 5.00$ when color is removed---a 5.3-point
drop alongside a roughly 17-fold increase in variance---and a similar
degradation is observed in Cross-F1, suggesting that sensor-type ambiguity
negatively impacts both steady-state recognition and boundary-sensitive
inference.

The Kyoto dataset is a partial exception: the White variant achieves a
marginal Pure-F1 gain ($62.38\%$ vs.\ $59.95\%$), but this advantage does
not transfer to the cross-window setting, where performance drops to $54.16\%$
(vs.\ $57.64\%$ with color). This pattern suggests that reduced feature
diversity may occasionally simplify intra-activity discrimination but
undermines the model's ability to capture transitional dynamics, which are
critical for boundary detection and mixed-activity windows.

\paragraph{Disk shape.}
We next vary the geometric primitive used to render each sensor---\emph{Circle}
(default), \emph{Triangle}, and \emph{Square}---while keeping the modality-aware
color assignment fixed. Table~\ref{tab:shape-ablation} shows that mean
performance is largely insensitive to this choice, with differences typically
within one F1 point in both settings.

\begin{table}[t]
\centering
\caption{Effect of disk shape on macro-F1 (\%, mean $\pm$ std over $3$ seeds).
The modality-aware color encoding is held fixed; only the geometric primitive
varies. Best per column in \textbf{bold}.}
\label{tab:shape-ablation}
\small
\resizebox{0.95\textwidth}{!}{%
\begin{tabular}{l cc cc cc cc}
\toprule
& \multicolumn{2}{c}{Aruba} & \multicolumn{2}{c}{Milan}
& \multicolumn{2}{c}{Kyoto} & \multicolumn{2}{c}{Orange} \\
\cmidrule(lr){2-3}\cmidrule(lr){4-5}\cmidrule(lr){6-7}\cmidrule(lr){8-9}
Shape & Pure & Cross & Pure & Cross & Pure & Cross & Pure & Cross \\
\midrule
Circle   & 72.37\,$\pm$\,0.29 & \textbf{56.87\,$\pm$\,0.60}
         & \textbf{69.25\,$\pm$\,0.39} & 53.06\,$\pm$\,0.85
         & 59.95\,$\pm$\,2.65 & 57.64\,$\pm$\,5.94
         & \textbf{62.28\,$\pm$\,2.85} & 59.93\,$\pm$\,2.17 \\
Triangle & \textbf{73.70\,$\pm$\,1.78} & 55.31\,$\pm$\,0.82
         & 67.68\,$\pm$\,1.79 & 54.49\,$\pm$\,0.47
         & 58.05\,$\pm$\,3.60 & 57.30\,$\pm$\,4.43
         & 60.99\,$\pm$\,0.34 & 59.08\,$\pm$\,5.95 \\
Square   & 73.25\,$\pm$\,1.64 & \textbf{56.87\,$\pm$\,0.60}
         & 68.83\,$\pm$\,2.41 & \textbf{54.56\,$\pm$\,0.63}
         & \textbf{62.79\,$\pm$\,2.28} & \textbf{59.58\,$\pm$\,3.38}
         & 57.94\,$\pm$\,1.21 & \textbf{60.32\,$\pm$\,0.34} \\
\bottomrule
\end{tabular}
}
\end{table}

No single shape consistently dominates across datasets and settings: in the
pure-window setting, Triangle achieves the best mean on Aruba while Square
does so on Kyoto, and the cross-window winner varies similarly. However,
the Triangle configuration tends to exhibit higher variance and less consistent
performance, particularly in more complex environments such as Kyoto and
Orange. While its mean performance is not drastically lower, the increased
variability suggests reduced robustness under challenging conditions,
especially in cross-window scenarios. Circles and squares yield comparable
and more stable representations across seeds. We adopt circles as the default
because they are rotation-invariant and avoid the orientation ambiguity of
triangles, while matching the stability of squares.

The rendering choice that materially affects recognition is the modality-aware coloring, which is essential whenever the deployment contains heterogeneous sensor types. The choice of geometric primitive, by contrast, behaves as a low-sensitivity hyperparameter. This supports the broader claim that the framework's gains derive from the layout-aligned trajectory \emph{structure} itself rather than from a hand-tuned visual encoding.
% \end{revblock}

\subsection{Consistent-layout perturbation.}
\label{app:consistent_layout_perturb}
We first revisit the consistent-layout setting, where perturbed coordinates are used for both model training and evaluation. Specifically, after adding i.i.d. zero-mean Gaussian noise with variance $\sigma^2 \in \{5,10,15\}$ to sensor locations, we regenerate the trajectory images and retrain the full model from scratch using the same training scheme. Table~\ref{tab:reposition_sigma} reports Macro-F1 on pure and cross-activity windows for the original layout and perturbed layouts.

\begin{table}[htbp]
\centering
\caption{Consistent-layout perturbation results. The same perturbed sensor coordinates are used during both training and testing.
We add zero-mean Gaussian noise with variance $\sigma^2 \in \{5,10,15\}$ 
to all sensor locations in the floor plan and report macro-F1 (\%, mean~$\pm$~std) 
on pure and cross-activity windows for each dataset.}
\label{tab:reposition_sigma}
\resizebox{0.95\textwidth}{!}{%
\begin{tabular}{l|cc|cc|cc|cc}
\toprule
\multirow{2}{*}{$\sigma^2$} &
\multicolumn{2}{c|}{\textbf{Milan}} &
\multicolumn{2}{c|}{\textbf{Aruba}} &
\multicolumn{2}{c|}{\textbf{Kyoto}} &
\multicolumn{2}{c}{\textbf{Orange}} \\
\cmidrule(lr){2-3}\cmidrule(lr){4-5}\cmidrule(lr){6-7}\cmidrule(lr){8-9}
 & Pure F1 & Cross F1 & Pure F1 & Cross F1 & Pure F1 & Cross F1 & Pure F1 & Cross F1 \\
\midrule
None  & $68.88 \pm 1.01$ & $ 52.11 \pm 0.78 $
   & $72.64 \pm 0.30$ & $ 55.75 \pm 1.36 $
   & $59.20 \pm 3.23$ & $ 55.78 \pm 4.40 $
   & $63.83 \pm 0.64$ & $ 61.19 \pm 3.22 $ \\
\midrule   
5  & $69.38 \pm 1.98$ & $53.37 \pm 2.22$
   & $73.39 \pm 2.43$ & $55.84 \pm 2.97$
   & $60.25 \pm 3.21$ & $58.62 \pm 0.81$
   & $59.99 \pm 4.69$ & $60.49 \pm 1.87$ \\
10 & $68.69 \pm 2.77$ & $53.66 \pm 1.62$
   & $72.94 \pm 0.39$ & $56.19 \pm 0.59$
   & $60.25 \pm 3.21$ & $58.62 \pm 0.81$
   & $58.56 \pm 1.02$ & $62.09 \pm 1.68$ \\
15 & $71.49 \pm 2.78$ & $54.24 \pm 0.30$
   & $73.21 \pm 1.58$ & $55.35 \pm 2.11$
   & $60.39 \pm 3.02$ & $58.60 \pm 0.80$
   & $59.82 \pm 2.70$ & $65.42 \pm 1.14$ \\
\bottomrule
\end{tabular}}
\end{table}

Across all four datasets, recognition performance remains stable when the same approximate layout is used consistently during training and testing. On Milan and Aruba, pure-window F1 fluctuates within only a few points of the unperturbed baseline, while cross-activity window F1 remains essentially unchanged as $\sigma^2$ increases from 5 to 15. Kyoto shows a similarly stable profile despite its denser deployment of motion and door sensors. Orange also remains robust, even though its sensor coordinates are only pseudo-locations sampled from room-level assignments.

These results should not be interpreted as evidence that spatial information is irrelevant. Rather, they show that LastAct does not depend on finely calibrated absolute pixel coordinates. The model appears to rely more on coarse spatial structure, including room-level topology, relative trajectory shape, event ordering, and room-to-room transitions. Thus, the consistent-layout experiment demonstrates that LastAct can operate with approximate but stable spatial metadata, which is important for real smart-home deployments where exact sensor coordinates are rarely available.

\subsection{Full Results Across Activity-Purity Buckets}
\label{app:purity_full_results}
\begin{figure}[t]
    \centering
    \includegraphics[width=0.95\linewidth]{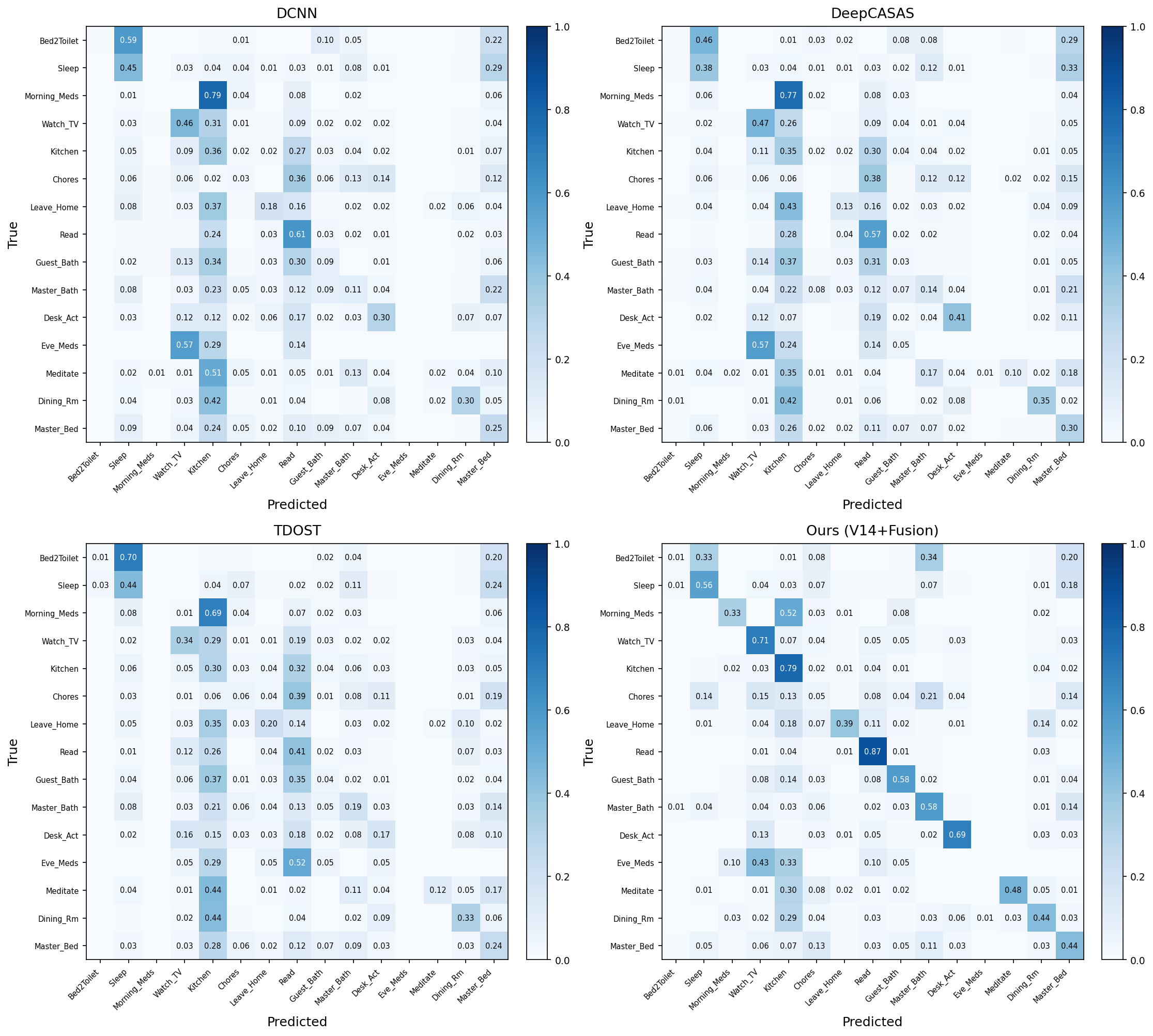}
    \caption{%
    Confusion matrices of DeepCASAS, DCNN, TDOST, and our method on Milan.
    }
    \label{fig:milan_cm}
\end{figure}

For completeness, Table~\ref{tab:purity_results_f1} lists the full purity-stratified Macro-F1 scores for all methods discussed in Section~\ref{sec:purity_ana}, from heavily contaminated (10--30\%) to fully pure (100\%) windows. We examine the confusion matrices of representative models, including DeepCASAS, DCNN, TDOST, and our method in Fig~\ref{fig:milan_cm} on milan dataset. Our method achieves substantially higher recall on Kitchen (0.79 vs. 0.30–0.36), Read (0.87 vs. 0.41–0.61), Master\_Bath and Guest\_Bath (0.58 vs. 0.11–0.19), and Desk\_Act (0.69 vs. 0.17–0.41). These activities tend to occur in well-defined spatial zones with relatively clean entry and exit transitions; once the boundary detector correctly segments these windows, classification within pure segments becomes considerably more reliable. In contrast, baseline methods confuse these activities with temporally adjacent ones — for instance, Kitchen is frequently misclassified as Watch\_TV or Morning\_Meds, reflecting the inability to resolve cross-window ambiguity near morning routine transitions.

\begin{table*}[htbp]
\centering
\caption{Macro-F1 across activity-purity buckets for each dataset and method.
Results are reported as Macro-F1 (mean $\pm$ std).}
\label{tab:purity_results_f1}
\resizebox{0.9\textwidth}{!}{
\begin{tabular}{cc|cccccc}
\hline
\textbf{Dataset} & \textbf{Purity Range}
& DeepCASAS
& DCNN
& TCN
& CPC
& TDOST
& \textbf{Ours} \\
\hline

\multirow{4}{*}{\textbf{Milan}}
& 10--30  & 5.92 $\pm$ 1.17 & 2.64 $\pm$ 0.25 & 3.25 $\pm$ 0.48 & 2.69 $\pm$ 0.16 & 2.41 $\pm$ 0.63 & \textbf{49.77 $\pm$ 0.94} \\
& 40--60  & 19.74 $\pm$ 0.70 & 15.74 $\pm$ 3.54 & 21.30 $\pm$ 2.79 & 11.68 $\pm$ 1.41 & 15.46 $\pm$ 1.52 & \textbf{54.36 $\pm$ 2.23} \\
& 70--90  & 41.78 $\pm$ 3.40 & 43.41 $\pm$ 4.49 & 46.53 $\pm$ 1.81 & 36.61 $\pm$ 1.80 & 31.74 $\pm$ 3.74 & \textbf{55.83 $\pm$ 3.50} \\
& 100     & 70.08 $\pm$ 3.17
           & 67.51 $\pm$ 3.69
           & 73.20 $\pm$ 4.34
           & \underline{73.25 $\pm$ 1.02}
           & \textbf{73.65 $\pm$ 3.80}
           & 68.88 $\pm$ 1.01 \\
\hline

\multirow{4}{*}{\textbf{Aruba}}
& 10--30  & 3.08 $\pm$ 1.15 & 1.13 $\pm$ 0.20 & 2.58 $\pm$ 1.27 & 1.13 $\pm$ 0.17 & 1.38 $\pm$ 0.50 & \textbf{58.72 $\pm$ 1.56} \\
& 40--60  & 8.51 $\pm$ 0.60 & 10.91 $\pm$ 0.69 & 20.74 $\pm$ 4.34 & 8.86 $\pm$ 1.98 & 7.26 $\pm$ 4.49 & \textbf{49.85 $\pm$ 0.65} \\
& 70--90  & 40.76 $\pm$ 0.78 & \underline{50.75 $\pm$ 2.88} & \textbf{52.84 $\pm$ 0.34} & 44.64 $\pm$ 4.60 & 16.41 $\pm$ 9.40 & 48.24 $\pm$ 3.26 \\
& 100     & 73.60 $\pm$ 7.55
           & 68.37 $\pm$ 4.27
           & \textbf{76.55 $\pm$ 2.37}
           & 74.31 $\pm$ 1.04
           & \underline{75.27 $\pm$ 1.25}
           & 72.64 $\pm$ 0.30 \\
\hline

\multirow{4}{*}{\textbf{Kyoto}}
& 10--30  & 20.09 $\pm$ 1.71 & 5.44 $\pm$ 0.25 & 3.71 $\pm$ 0.75 & 3.28 $\pm$ 0.60 & 2.98 $\pm$ 0.19 & \textbf{49.47 $\pm$ 5.11} \\
& 40--60  & 43.48 $\pm$ 2.99 & 17.54 $\pm$ 0.62 & 19.33 $\pm$ 2.67 & 14.11 $\pm$ 1.91 & 23.07 $\pm$ 3.78 & \textbf{57.63 $\pm$ 4.48} \\
& 70--90  & 50.57 $\pm$ 3.93 & 46.98 $\pm$ 1.47 & 47.95 $\pm$ 2.67 & 42.69 $\pm$ 7.40 & 41.02 $\pm$ 1.92 & \textbf{59.09 $\pm$ 5.49} \\
& 100     & 49.91 $\pm$ 5.65
           & 57.57 $\pm$ 3.56 
           & 57.60 $\pm$ 3.62
           & 52.52 $\pm$ 4.50
           & 55.62 $\pm$ 8.14
           & \textbf{59.20 $\pm$ 3.23} \\
\hline

\multirow{4}{*}{\textbf{Orange}}
& 10--30  & 5.87 $\pm$ 0.08 & 10.81 $\pm$ 0.08 & 5.87 $\pm$ 0.08 & 7.89 $\pm$ 0.18 & 6.21 $\pm$ 0.85 & \textbf{53.14 $\pm$ 5.97} \\
& 40--60  & 19.99 $\pm$ 2.63 & 15.30 $\pm$ 2.12 & 15.78 $\pm$ 4.08 & 15.71 $\pm$ 2.19 & 45.53 $\pm$ 2.24 & \textbf{64.14 $\pm$ 4.23} \\
& 70--90  & 24.86 $\pm$ 2.33 & 45.18 $\pm$ 4.07 & 22.82 $\pm$ 7.04 & 34.86 $\pm$ 6.97 & \textbf{67.54 $\pm$ 3.95} &  \underline{64.24 $\pm$ 1.02} \\
& 100     & 25.13 $\pm$ 0.30
           & \underline{69.37 $\pm$ 6.56}
           & 28.38 $\pm$ 2.90
           & 50.50 $\pm$ 14.81
           & \textbf{79.00 $\pm$ 3.01}
           & 63.83 $\pm$ 0.64 \\
\hline
\end{tabular}
}
\end{table*}

% \begin{revblock}

\subsection{Feasibility Study on Parallel-Activity Detection}
\label{app:parallel_feasibility}

Although the main LastAct framework is designed for single-label latest-activity recognition, the proposed layout-aligned trajectory representation may also be useful for parallel-activity scenarios. In multi-resident smart homes, concurrent activities can activate spatially distinct sensor regions, producing multiple trajectory components on the floor plan. To examine whether our representation captures such signals, we conduct a preliminary isolated-vs-parallel activity detection experiment on Kyoto7, the multi-resident dataset in our evaluation.

\paragraph{Dataset construction.}
We construct a binary classification task from the original Kyoto7 episode annotations. Each annotated episode is represented by its time interval and activity label. We define a \emph{parallel episode} as an episode whose time interval overlaps with at least one episode from a different activity category. Because Kyoto7 contains resident-specific prefixes, activity categories are determined after removing the \texttt{R1\_}/\texttt{R2\_} prefix. For example, \texttt{R1\_Work} and \texttt{R2\_Work} are treated as the same activity category, whereas \texttt{R1\_Work} and \texttt{R2\_Personal\_Hygiene} are treated as different categories. We define an \emph{isolated episode} as an episode during which no other activity is active at any point; that is, the episode is the only active activity label throughout its temporal interval.

We exclude same-category overlaps across residents, such as \texttt{R1\_Sleep}$\times$\texttt{R2\_Sleep} and \texttt{R1\_Work}$\times$\texttt{R2\_Work}. These cases are inherently ambiguous under anonymous ambient sensing because the two residents may trigger similar or identical sensors in the same functional regions. Distinguishing such cases would require explicit resident-identity tracking, which is not available in the sensor stream. We also remove severely underrepresented categories, including \texttt{Clean} and \texttt{Wash\_Bathtub}, to avoid degenerate strata during data splitting.

For each retained episode, we extract the first 100 valid sensor events, convert them into the same layout-aligned trajectory representation used in LastAct, and apply zero-padding if the episode contains fewer than 100 valid events. Longer episodes are truncated to the first 100 valid events. Each sample is assigned a binary label: isolated = 0 and parallel = 1. We split the data into train, validation, and test sets using stratification over a composite key of binary label and activity bucket whenever possible, falling back to label-only stratification when sparse strata prevent composite stratification. This yields an overall 64\%/16\%/20\% train/validation/test split.

\begin{table}[h]
\centering
\caption{Dataset statistics for the Kyoto7 isolated-vs-parallel feasibility experiment.}
\label{tab:parallel_dataset}
\begin{tabular}{lcccc}
\toprule
Split & Isolated & Parallel & Total & Parallel ratio \\
\midrule
Train & 135 & 177 & 312 & 56.7\% \\
Validation & 34 & 45 & 79 & 57.0\% \\
Test & 42 & 56 & 98 & 57.1\% \\
\bottomrule
\end{tabular}
\end{table}

\paragraph{Model and training protocol.}
We reuse the trajectory CNN encoder pretrained in the main LastAct framework and train a lightweight binary classifier for isolated-vs-parallel discrimination. The classifier follows the same representation pipeline from sensor events to trajectory frames and temporal sequence modeling, but replaces the original multi-class activity head with a binary output head. To stabilize learning on this small feasibility dataset, we use balanced sampling and inverse-frequency weighted binary cross-entropy. The CNN backbone is frozen during the initial training stage so that the temporal classifier and binary head can first learn a stable mapping from pretrained spatial features before the encoder is updated. We do not introduce a parallel-activity-specific architecture, resident-disentanglement module, multi-label objective, or trajectory-decomposition loss.

\paragraph{Evaluation metrics.}
Since the task is binary classification, we report two groups of metrics. The first group consists of task-level metrics that evaluate both classes jointly: accuracy, balanced accuracy, and Macro-F1. Accuracy measures the total fraction of correctly classified samples, balanced accuracy averages recall over isolated and parallel episodes, and Macro-F1 averages the F1 scores of the two classes. The second group consists of parallel-class metrics, where parallel episodes are treated as the positive class. Parallel precision measures how many episodes predicted as parallel are truly parallel, parallel recall measures how many true parallel episodes are detected, and parallel F1 summarizes the precision--recall trade-off for the parallel class.

\begin{table}[h]
\centering
\caption{
Performance of the Kyoto7 isolated-vs-parallel feasibility experiment over three independent runs.
Task-level metrics evaluate both classes jointly, whereas parallel-class metrics evaluate detection of the positive class only.
}
\label{tab:parallel_feasibility}
\begin{tabular}{llc}
\toprule
Metric group & Metric & Score \\
\midrule
\multirow{3}{*}{Task-level}
& Accuracy & $0.7347 \pm 0.0083$ \\
& Balanced accuracy & $0.7153 \pm 0.0134$ \\
& Macro F1 & $0.7182 \pm 0.0135$ \\
\midrule
\multirow{3}{*}{Parallel-class only}
& Precision for parallel & $0.7305 \pm 0.0180$ \\
& Recall for parallel & $0.8512 \pm 0.0223$ \\
& F1 for parallel & $0.7857 \pm 0.0012$ \\
\bottomrule
\end{tabular}
\end{table}

\paragraph{Results and interpretation.}
As shown in Table~\ref{tab:parallel_feasibility}, the model achieves stable overall discrimination between isolated and parallel episodes, with $0.7153 \pm 0.0134$ balanced accuracy and $0.7182 \pm 0.0135$ Macro-F1. These task-level metrics indicate that the classifier does not simply collapse to one label. When focusing specifically on the parallel class, the model obtains $0.8512 \pm 0.0223$ recall and $0.7857 \pm 0.0012$ F1, suggesting that the trajectory representation is sensitive to spatial signatures associated with concurrent activity. The lower parallel precision ($0.7305 \pm 0.0180$) indicates that some isolated episodes are falsely flagged as parallel, which is expected given the limited data size and the absence of a task-specific parallel-activity architecture.

Overall, this experiment supports the feasibility of applying layout-aligned trajectory encoding to parallel-activity detection. However, it should not be interpreted as solving full parallel-activity recognition. The current formulation only detects whether an episode contains overlap; it does not predict multiple simultaneous activity labels or assign activities to specific residents. A full extension would require multi-label prediction heads, resident-aware modeling, and potentially trajectory-decomposition objectives that separate simultaneous spatial activity components.

% \end{revblock}
\end{document}